\documentclass[12pt]{article} 

\usepackage[sectionbib]{natbib}
\usepackage{bbm}
\usepackage{array,epsfig,fancyhdr,rotating}
\usepackage[hidelinks]{hyperref}
\usepackage{secdot}
\usepackage{flafter}
\renewcommand{\theequation}{\thesection\arabic{equation}}
\usepackage{placeins}

\usepackage{subcaption}
\usepackage[linesnumbered,ruled,vlined]{algorithm2e}
\usepackage[table]{xcolor}

\usepackage{amsmath, amssymb, amsfonts}
\usepackage{amsthm}
\usepackage{enumitem}
\usepackage{graphicx}
\usepackage{makecell}
\usepackage{multirow}
\usepackage{float}
\usepackage{booktabs}
\usepackage{tabularx}
\newcolumntype{Y}{>{\raggedright\arraybackslash}X}

\newtheorem{theorem}{Theorem}
\newtheorem{lemma}{Lemma}
\newtheorem{corollary}{Corollary}

\theoremstyle{definition}

\newtheorem{assumption}[theorem]{Assumption}

\newtheorem{remark}{Remark}

\DeclareMathOperator*{\argmin}{arg\,min}

\usepackage{xr}
\begin{document}



\fontsize{12}{14pt plus.8pt minus .6pt}\selectfont \vspace{0.8pc}
\centerline{\large\bf Exact Recovery by Neighborhood Smoothing}
\vspace{2pt}
\centerline{\large\bf in Directed Stochastic Block Models}
\vspace{.4cm}

\centerline{\bf Behzad Aalipur\textsuperscript{*}}
\vspace{1pt}
\centerline{\it Department of Statistics, University of Cincinnati}
\vspace{2pt}
\centerline{\textsuperscript{*}Corresponding author email: \href{mailto:aalipubd@mail.uc.edu}{aalipubd@mail.uc.edu}}
\vspace{.3cm}

\centerline{\bf Yichen Qin}
\vspace{1pt}
\centerline{\it Lindner College of Business, University of Cincinnati}
\vspace{2pt}
\centerline{\href{mailto:qinyn@ucmail.uc.edu}{qinyn@ucmail.uc.edu}}
\vspace{.55cm}

\fontsize{9}{11.5pt plus.8pt minus.6pt}\selectfont

\vspace{.55cm} \fontsize{9}{11.5pt plus.8pt minus.6pt}\selectfont


\begin{abstract}
We study exact community recovery in sparse directed stochastic block
models using neighborhood smoothing of connection-probability profiles.
The proposed method clusters vertices according to their estimated
outgoing connection-probability profiles. For each vertex, its complete
outgoing profile is estimated by averaging the adjacency rows of
empirically similar vertices, after which \(K\)-means is applied to the
estimated profiles. An analogous procedure based on incoming
connection-probability profiles is obtained by applying the same
construction to the transposed adjacency matrix.

We establish a finite-sample uniform row-wise error bound for the
asymmetric smoothed estimator and derive consistency in the normalized
two-to-infinity norm. We then show that exact recovery follows when the
minimum separation between distinct population profiles dominates the
row-wise estimation error. The result permits a vanishing sparsity
factor, a general asymmetric block-probability matrix, and a number of
communities that may diverge with the network size. We are unaware of a previous exact-recovery theorem that simultaneously
covers these features for a directed stochastic block model. Numerical studies and an application to a directed neuronal
connectome illustrate the practical behavior and limitations of the
profile-based clustering approach.
\end{abstract}

\vspace{9pt}
\noindent {\it Key words and phrases:}
stochastic block model; directed networks; exact recovery; community detection; neighborhood smoothing; sparse graphs
\par
\def\thefigure{\arabic{figure}}
\def\thetable{\arabic{table}}

\renewcommand{\theequation}{\thesection.\arabic{equation}}

\fontsize{12}{14pt plus.8pt minus .6pt}\selectfont

\section{Introduction}

Stochastic block models (SBMs) are foundational models for community
detection in network data \citep{holland1983stochastic}. The central
inferential task is to recover latent community labels from a single
observed network. Common approaches include modularity- and likelihood-based criteria
\citep{newman2004detecting,bickel2009nonparametric,
zhao2012consistency}, pseudo-likelihood methods
\citep{amini2013pseudo}, and method-of-moments procedures
\citep{bickel2011method}. Other
widely used approaches are spectral clustering
\citep{rohe2011spectral,lei2015consistency,qin2013regularized,
joseph2016impact,su2019strong} and spectral embeddings
\citep{sussman2012consistent,lyzinski2014perfect}.

Exact recovery requires all node labels to be recovered correctly with
probability tending to one, up to a permutation of the community
labels. This objective is particularly delicate for directed networks,
because the adjacency and conditional mean matrices are generally
nonsymmetric, and a vertex may have different outgoing and incoming
connectivity profiles. Symmetrizing a directed graph can therefore
discard important directional information, especially in citation,
hyperlink, communication, and flow networks
\citep{malliaros2013clustering}. This motivates procedures that preserve
edge orientation throughout both estimation and clustering.

The closest literature has two main components. The first concerns
directed community detection through singular-vector embeddings,
co-clustering, score normalization, random walks, and likelihood-based
criteria. Examples include spectral co-clustering for stochastic
co-block models \citep{rohe2016co}, D-SCORE for directed
degree-corrected block models \citep{wang2020spectral}, and adjacency
eigenvector methods for partial recovery in sparse directed networks
\citep{coste2021simpler}. In particular,
\citet{wang2020spectral} establish a high-probability upper bound on the
number of vertices misclustered by D-SCORE under a directed
degree-corrected block model. Their result provides a consistency or
misclustering guarantee, but it does not establish exact recovery of all
vertices.

Exact recovery by adjacency spectral embedding has been established for
undirected stochastic block models and related random dot product graph
models under conditions requiring sufficient eigenvalue separation,
population-position separation, and graph density
\citep{lyzinski2014perfect}. These results do not address a general
asymmetric directed block matrix.

Exact recovery is also available in several more specialized asymmetric
settings. For example, \citet{cohenaddad2022community} study a
two-community directed degree-heterogeneous model in which each sender
has separate within- and between-community connection probabilities.
Other results concern bipartite or rectangular block models with
separate row and column community assignments, often using a spectral
initialization followed by an iterative refinement step
\citep{ndaoud2022improved,braun2023minimax,braun2024strong}.
These settings differ from the present sparse directed SBM, which has a
single community assignment for both edge endpoints, a general
asymmetric block matrix, and a number of communities that may diverge.

The second component concerns direct estimation of the edge-probability
matrix. The neighborhood-smoothing method of
\citet{zhang2017estimating} estimates an undirected probability matrix
by averaging empirically similar adjacency rows, with emphasis on matrix
estimation and link prediction.

We are unaware of a previous exact-recovery result for a sparse
directed stochastic block model that simultaneously permits a single
community assignment, a general asymmetric block matrix, a vanishing
sparsity factor, and a diverging number of communities. We obtain such
a guarantee by constructing an asymmetric estimator
\(\widetilde{\mathbf P}\) of the sparsity-scaled block-profile target
\(\mathbf P\), establishing the uniform row-wise bound
\[
\|\widetilde{\mathbf P}-\mathbf P\|_{2\to\infty},
\]
and combining it with separation between distinct outgoing block
profiles. The resulting inferential route is
\[
\mathbf A
\longrightarrow
\widetilde{\mathbf P}
\longrightarrow
\widehat\pi,
\]
so estimation and recovery are both formulated on the
outgoing-profile scale.

Our contribution bridges these two strands of literature: directed
community recovery and neighborhood-based probability-matrix
estimation. The smoothing rule is adapted from
\citet{zhang2017estimating}; the new contribution is its directed SBM
recovery analysis. The main contributions are as follows.

(i) We extend neighborhood smoothing to an asymmetric directed
block-profile target and cluster the estimated outgoing profiles rather
than a singular-vector, random-walk, or score-normalized embedding.

(ii) We prove a uniform two-to-infinity error bound for the asymmetric
smoothed estimator under sparsity and a potentially increasing number of
communities.

(iii) We combine this row-wise bound with a minimum outgoing-profile
separation condition to prove exact recovery in a regime that permits
both \(\gamma_n\to0\) and \(K_n\to\infty\) under a general asymmetric
block matrix. The recovery condition is expressed on the
outgoing-profile scale rather than through a spectral embedding.

(iv) We compare \(K\)-means on estimated block profiles (KMP)
numerically with \(K\)-means on adjacency rows (KMA), left-right
spectral clustering, and D-SCORE using \(MC=100\) paired replicates,
including sparsity, nested-growth, and controlled growing-\(K_n\) stress
tests, and we analyze the directed chemical-synapse connectome of
\textit{C. elegans}.

The remainder of the paper is organized as follows.
Section~\ref{sec:framework} introduces the sparse directed SBM, and
Section~\ref{sec:methodology} presents the smoothing estimator and
clustering procedure. Section~\ref{sec:theory} establishes the uniform
row-wise error bound and exact-recovery result;
Section~\ref{subsec:threshold-comparison} compares the resulting
sufficient condition with classical exact-recovery thresholds.
Section~\ref{sec:simulations} reports the simulation studies, and
Section~\ref{sec:real-data} analyzes the directed chemical-synapse
connectome of \textit{C. elegans}. Technical proofs, auxiliary lemmas,
and additional numerical results are provided in the Supplementary
Material.
\section{Framework}\label{sec:framework}
Consider a directed binary network with \(n\) nodes, represented by a graph \(G=(V,E)\), where \(V=[n]=\{1,\ldots,n\}\) and \(E\subseteq V\times V\). Its adjacency matrix is \(\mathbf A=[A_{ij}]\in\{0,1\}^{n\times n}\), where \(A_{ij}=1\) if there is a directed edge from node \(i\) to node \(j\), and \(A_{ij}=0\) otherwise. Self-loops are excluded, so \(A_{ii}=0\) for all \(i\).

We assume that \(G\) follows a sparse directed stochastic block model,
which we denote by
\begin{equation}
\label{eq:DSBM}
G\sim
\mathrm{DSBM}
\bigl(n,K_n,\boldsymbol{\rho},\mathbf B,\gamma_n\bigr).
\end{equation}
Each node \(i\in V\) is independently assigned to one of the \(K_n\)
communities according to the community proportions
\(\boldsymbol{\rho}\).
Let \(Z_i\in[K_n]\) denote the latent label of node \(i\), with
\begin{equation}
\label{model:z_i}
Z_i
\stackrel{\mathrm{i.i.d.}}{\sim}
\mathrm{Categorical}(\boldsymbol\rho),
\qquad
\boldsymbol\rho=(\rho_1,\ldots,\rho_{K_n}),
\qquad
i=1,\ldots,n.
\end{equation}
Write
\[
\mathbf Z=(Z_1,\ldots,Z_n)^\top
\]
for the vector of latent labels. Let \(n_j\) denote the number of nodes
assigned to community \(j\), so that
\(n=\sum_{j=1}^{K_n}n_j\).
Define
\[
\rho_{\min}:=\min_{j\in[K_n]}\rho_j.
\]
This quantity may depend on \(n\) and may tend to zero, subject to the
lower-bound conditions imposed in Section~\ref{sec:theory}. Let \(z_i\)
denote the realized value of \(Z_i\), and define the community-assignment
map by
\[
\pi(i):=z_i.
\]
Let \(\mathbf B\in[0,1]^{K_n\times K_n}\) be the directed
block-probability matrix, where \(B_{jk}\) is the base probability of an
edge from a node in community \(j\) to a node in community \(k\).

We scale the block probabilities by a sequence
\(\{\gamma_n\}_{n\geq1}\), where \(0<\gamma_n\leq1\). The sparse regime
corresponds to \(\gamma_n\to0\), whereas \(\gamma_n\equiv1\) gives a
dense directed SBM. Conditional on the community assignments, edges are
generated independently as
\begin{align}
\label{adjacency_mat}
A_{ij} \stackrel{\text{ind.}}{\sim} \mathrm{Bernoulli}\bigl(\gamma_n B_{\pi(i) \pi(j)}\bigr), \quad i \ne j, 
\end{align}
and \(A_{ii}=0\) for \(i\in[n]\). To separate the block geometry from
the sparsity scaling, define the unscaled block-profile matrix
\(\mathbf P^0=[P^0_{ij}]\) by
\begin{equation}
\label{Prob_n*n_matrix}
P^0_{ij}
:=
B_{\pi(i)\pi(j)},
\qquad
i,j\in[n],
\qquad
\mathbf P
:=
\gamma_n\mathbf P^0.
\end{equation}
Thus, \(\mathbf P^0\) records the unscaled outgoing block profiles and
\(\mathbf P\) is the corresponding sparsity-scaled block-profile
target. For \(i\neq j\),
\[
P_{ij}
=
\mathbb P(A_{ij}=1\mid\mathbf Z),
\]
whereas the diagonal entries \(P_{ii}\) are retained only to preserve
the complete block-row geometry. Since self-loops are excluded, the zero-diagonal conditional mean
adjacency matrix is
\[
\mathbf P^{\mathrm{off}}
:=
\mathbb E(\mathbf A\mid\mathbf Z)
=
\mathbf P-\operatorname{diag}(\mathbf P).
\]
Thus, \(\mathbf P^{\mathrm{off}}\) is the conditional mean of the
observed adjacency matrix, whereas \(\mathbf P\) is the full
block-profile target used in the estimation and recovery analysis. Their
difference is confined to the diagonal, and
\[
\frac{1}{\sqrt n}
\|\mathbf P-\mathbf P^{\mathrm{off}}\|_{2\to\infty}
=
\frac{1}{\sqrt n}\max_{i\in[n]}P_{ii}
\leq
\frac{\gamma_n}{\sqrt n}
\leq
\frac{1}{\sqrt n}.
\]
Hence, the two matrices are asymptotically equivalent on the normalized
row-wise scale, while retaining \(\mathbf P\) preserves exactly
identical population rows within each community. After observing
\(\mathbf A\), the clustering task is to estimate the community
assignment \(\pi\), or equivalently, the latent labels
\(Z_1,\ldots,Z_n\). This is a transductive recovery problem for the
vertices in the single observed network, up to label permutation; we do
not study out-of-sample classification of newly arriving vertices in the
theoretical analysis.

Assume that the rows of \(\mathbf B\) are distinct. For communities
\(r,s\in[K_n]\), define
\[
d_{\mathbf B}(r,s)
:=
\|\mathbf B_{r\cdot}-\mathbf B_{s\cdot}\|_2,
\qquad
d_{\mathbf B}^*
:=
\min_{r\neq s}d_{\mathbf B}(r,s)>0.
\]
For nodes \(i\) and \(j\) from different communities,
\begin{equation}
\label{P_p_nul_B relation}
\begin{aligned}
d_{\mathbf P}(i,j)
&:=
\|\mathbf P_{i\cdot}-\mathbf P_{j\cdot}\|_2\\
&=
\gamma_n
\left\{
\sum_{k=1}^{K_n}
n_k\bigl(B_{\pi(i)k}-B_{\pi(j)k}\bigr)^2
\right\}^{1/2}\\
&\geq
\gamma_n\sqrt{n_{\min}}\,d_{\mathbf B}^*,
\end{aligned}
\end{equation}
where
\[
n_{\min}:=\min_{k\in[K_n]}n_k.
\]
Accordingly, define the minimum between-community row separation by
\[
d_{\mathbf P}^*
:=
\min_{\pi(i)\neq\pi(j)}
d_{\mathbf P}(i,j).
\]
Equation~\eqref{P_p_nul_B relation} gives
\[
d_{\mathbf P}^*
\geq
\gamma_n\sqrt{n_{\min}}\,d_{\mathbf B}^*.
\]

Let \(\widehat\pi:[n]\to[K_n]\) be an estimator of the true community
assignment \(\pi\). Because community labels are identifiable only up
to permutation, define the label-matched clustering accuracy by
\begin{equation}
\label{eq:A_pi}
A(\pi,\widehat\pi)
=
\max_{\sigma\in\mathcal S_{K_n}}
\frac1n
\sum_{i=1}^n
\mathbbm 1
\left\{
\pi(i)=\sigma\bigl(\widehat\pi(i)\bigr)
\right\},
\end{equation}
where \(\mathcal S_{K_n}\) is the set of permutations of \([K_n]\).
The estimator achieves exact recovery if
\[
\mathbb P\{A(\pi,\widehat\pi)=1\}
=
1-o(1).
\]

\begin{table}[t]
\centering
\caption{Main notation used in the model, estimator, and recovery theory.}
\label{tab:main-notation}
\small
\begin{tabularx}{\textwidth}
{@{}>{\raggedright\arraybackslash}p{0.26\textwidth}
>{\raggedright\arraybackslash}X@{}}
\toprule
Symbol & Meaning\\
\midrule
\(K_n\) & number of communities, allowed to depend on \(n\)\\
\(\pi(i)\) & latent community label of vertex \(i\)\\
\(\boldsymbol\rho\), \(\rho_{\min}\) & community proportions and their minimum\\
\(n_{\min}\) & minimum realized community size\\
\(\mathbf B\) & base block-probability matrix; row
\(\mathbf B_{a\cdot}\) is the unscaled outgoing
connection-probability profile of community \(a\)\\
\(\gamma_n\) & sparsity multiplier, with \(0<\gamma_n\le 1\)\\
\(\mathbf P^0\) & node-level expansion of \(\mathbf B\);
\(P^0_{ij}=B_{\pi(i)\pi(j)}\), so row
\(\mathbf P^0_{i\cdot}\) is the unscaled outgoing profile of vertex \(i\)\\
\(\mathbf P=\gamma_n\mathbf P^0\) & sparsity-scaled block-profile target\\
\(\mathbf P^{\mathrm{off}}=\mathbb E(\mathbf A\mid\mathbf Z)\)
& zero-diagonal conditional mean adjacency matrix\\
\(d_{\mathbf B}^*\) & minimum Euclidean separation between distinct rows of \(\mathbf B\)\\
\(d_{\mathbf P}^*\) & minimum Euclidean separation between distinct
population rows of \(\mathbf P\)\\
\(N_i\) & empirical neighborhood used to smooth row \(i\)\\
\(\widetilde{\mathbf P}\) & neighborhood-smoothed estimator of \(\mathbf P\)\\
\(A(\pi,\widehat\pi)\) & label-matched clustering accuracy\\
\bottomrule
\end{tabularx}
\end{table}

Having established the statistical model and notation, we next present
the proposed methodology for estimating the directed block-profile
target \(\mathbf P\) and recovering the community assignments.

\section{Methodology}
\label{sec:methodology}
KMP has two steps. First, it estimates the directed block-profile target
\(\mathbf P\) by neighborhood smoothing. Second, it applies
\(K\)-means to the estimated rows to recover the community labels.

\subsection{Outgoing Block-Profile Estimation via Neighborhood Smoothing}

Following \citet{zhang2017estimating}, for each node \(i\in[n]\) we
construct a neighborhood \(N_i\) of vertices having similar outgoing
adjacency profiles. For distinct nodes \(i\) and \(j\), define
\begin{equation}
\label{eq:distance}
d(i,j)
=
\max_{k\neq i,j}
\left|
\frac{1}{n}
\left\langle
\mathbf A_{i\cdot}-\mathbf A_{j\cdot},
\mathbf A_{k\cdot}
\right\rangle
\right|,
\end{equation}
where \(\mathbf A_{i\cdot}\) denotes the \(i\)-th row of
\(\mathbf A\), and \(\langle\cdot,\cdot\rangle\) is the standard inner
product. The factor \(1/n\) is a normalization convention and does not
affect the empirical-quantile neighborhoods, because multiplying all
dissimilarities by the same positive constant leaves their ordering
unchanged.

For a chosen quantile parameter \(h\in(0,1)\), define the lower empirical
\(h\)-quantile by
\[
q_i(h)
:=
\inf\left\{
t\in\mathbb R:
\frac{1}{n-1}
\sum_{j\neq i}
\mathbbm 1\{d(i,j)\leq t\}
\geq h
\right\}.
\]
Equivalently, \(q_i(h)\) is the
\(\lceil h(n-1)\rceil\)-th order statistic of
\(\{d(i,j):j\neq i\}\). Define
\begin{equation}
\label{eq:neighborhood_i}
N_i
=
\left\{
j\in[n]\setminus\{i\}:d(i,j)\leq q_i(h)
\right\}.
\end{equation}

The empirical quantile rule targets approximately an \(h\)-fraction of
the candidate nodes. Let
\[
s_n:=\left\lceil h(n-1)\right\rceil
\]
denote the nominal neighborhood size. Because ties at the empirical
quantile may inflate \(|N_i|\), the realized neighborhood size can exceed \(s_n\); this behavior is
preserved in both the estimator and the simulations. The
neighborhood-smoothed block-profile estimator
\(\widetilde{\mathbf P}=[\widetilde P_{ij}]\) is defined by

\begin{equation}
\label{P_tilde_Estimate}
 \widetilde P_{ij} = \frac{1}{|N_i|} \sum_{i' \in N_i} A_{i'j}.   
\end{equation}
The estimator is asymmetric and is not symmetrized: it estimates the
outgoing connection-probability profile of node \(i\) by averaging
adjacency rows from empirically similar vertices. An incoming-profile
version is obtained by applying the same construction to
\(\mathbf A^\top\), thereby averaging adjacency columns rather than
rows.

\begin{remark}[Incoming-profile analogue]
Conditional on the labels, \(\mathbf A^\top\) follows the same directed
SBM with block matrix \(\mathbf B^\top\). Consequently, all estimation
and recovery results apply to the incoming-profile procedure after
replacing the minimum row separation \(d_{\mathbf B}^*\) by
\[
d_{\mathbf B,\mathrm{in}}^*
:=
\min_{a\neq b}
\|\mathbf B_{\cdot a}-\mathbf B_{\cdot b}\|_2.
\]
In particular, exact recovery from incoming profiles requires distinct
columns of \(\mathbf B\) and the corresponding signal condition
\[
\gamma_n d_{\mathbf B,\mathrm{in}}^*\rho_{\min}
\geq
C_{\mathrm{sep}}
\left(\frac{\log n}{n}\right)^{1/4}.
\]
\end{remark}

The parameter \(h\) controls the bias-variance tradeoff. Larger
neighborhoods reduce variance but can average across different
population profiles, while smaller neighborhoods reduce this bias at
the cost of higher variance. In the stochastic block model, the theory
below uses
\[
h=C_h\sqrt{\frac{\log n}{n}}.
\]
This choice is not claimed to be optimal; it is a sufficient bandwidth
that balances neighborhood growth with contamination control.

Our theoretical results establish uniform row-wise accuracy of
\(\widetilde{\mathbf P}\). We measure this error using the
two-to-infinity norm,
\[
\|\widetilde{\mathbf P}-\mathbf P\|_{2\to\infty}
:=
\max_{i\in[n]}
\|\widetilde{\mathbf P}_{i\cdot}-\mathbf P_{i\cdot}\|_2.
\]
Thus, KMP estimates the complete outgoing block profile of every vertex
before clustering is performed.

\subsection{Community Detection via \texorpdfstring{$K$}{K}-means Clustering}

After obtaining \(\widetilde{\mathbf P}\), we cluster its rows to recover
the community assignments. Corollary~\ref{cor:two_to_infinity}
establishes uniform consistency on the normalized row scale,
\[
\frac{1}{\sqrt n}
\|\widetilde{\mathbf P}-\mathbf P\|_{2\to\infty}
=
\mathcal O_P\!\left\{
\left(\frac{\log n}{n}\right)^{1/4}
\right\}
=
o_P(1).
\]
We use this row-wise accuracy as the input to a deterministic
exact-recovery argument, extending the probability-matrix estimation
framework of \citet{zhang2017estimating} to community recovery.
Specifically, define
\begin{equation}
\begin{aligned}
\widehat{\mathbf C}
&\in
\argmin_{\mathbf C\in\mathcal C_{K_n}}
\|\mathbf C-\widetilde{\mathbf P}\|_F,\\
\mathcal C_{K_n}
&:=
\left\{
\mathbf C\in\mathbb R^{n\times n}:
\mathbf C\text{ has at most }K_n\text{ distinct rows}
\right\}.
\end{aligned}
\label{eq:cluster_projection}
\end{equation}
Each distinct row of \(\widehat{\mathbf C}\) represents a fitted
centroid, and vertices assigned the same row form an estimated cluster.
This is the global \(K\)-means objective with at most \(K_n\) fitted
centroids. Under the separation condition in
Lemma~\ref{lem:deterministic_clustering}, every global minimizer uses
exactly \(K_n\) distinct centroids, one for each true community.

The recovery analysis combines uniform row-wise accuracy of
\(\widetilde{\mathbf P}\) with separation between distinct rows of
\(\mathbf P\). The number of communities is treated as known in the
main theorem; the next subsection records the corresponding plug-in
result for a consistent estimator of \(K_n\).

\subsection{Known and estimated numbers of communities}
\label{subsec:estimating-K}

The main theory treats \(K_n\) as known and does not analyze a specific
model-selection procedure. A singular-gap rule need not identify
\(K_n\), because \(\operatorname{rank}(\mathbf B)\) may be smaller than
\(K_n\), and some community-relevant singular directions may be weak.
Corollary~\ref{cor:estimated-K} shows that exact recovery is nevertheless
preserved when \(K_n\) is replaced by any consistent estimator
\(\widehat K_n\).

If a working value \(K<K_n\) is used, exact recovery is impossible
because at most \(K\) fitted centroids cannot represent the \(K_n\)
nonempty true communities. If \(K>K_n\), the enlarged clustering
parameter space permits additional fitted centroids and may split one
or more true communities, although a global minimizer need not use all
\(K\) available centroids. Thus, graded criteria such as ARI may remain high under mild
misspecification of \(K_n\). Consistency of \(\widehat K_n\) is a
simple sufficient condition under which the oracle exact-recovery
theorem transfers to the plug-in procedure. Underestimation rules out
exact recovery, whereas overestimation enlarges the feasible clustering
space and may lead to splitting; we do not analyze recovery under such
overestimation.

\subsection{Computational cost}
\label{subsec:comp_cost}

Computing the exact dissimilarities in \eqref{eq:distance} has dense
worst-case cost \(O(n^3)\), because \(O(n^2)\) node pairs are compared
over \(O(n)\) witness nodes. For comparison, a truncated rank-\(K_n\)
singular-value calculation has approximate dense cost \(O(n^2K_n)\).

The simulations use a landmark set
\(\mathcal A\subset[n]\), \(|\mathcal A|=m\ll n\), and replace
\(d(i,j)\) by
\[
d_m(i,j)
=
\max_{k\in\mathcal A\setminus\{i,j\}}
\left|
\frac1n
\left\langle
\mathbf A_{i\cdot}-\mathbf A_{j\cdot},
\mathbf A_{k\cdot}
\right\rangle
\right|.
\]
Forming \(\mathbf A\mathbf A_{\mathcal A}^{\top}\) and evaluating the
landmark dissimilarities cost \(O(n^2m)\) dense operations and require
\(O(nm)\) memory, apart from optional storage of all pairwise
dissimilarities. Row averaging and \(K\)-means incur additional costs
depending on the realized neighborhood sizes and the number of
iterations. Algorithm~\ref{alg:landmark-kmp} and complete memory
tradeoffs are given in the Supplementary Material.

The theory concerns the exact all-witness dissimilarity. The landmark
implementation preserves the same endpoint-exclusion rule but maximizes
over only \(m\) randomly selected admissible witnesses. A conditional
stability result for this approximation is given in
Remark~\ref{rem:landmark-stability} of the Supplementary Material;
obtaining an explicit landmark-sampling rate remains open.

\section{Theoretical Properties}
\label{sec:theory}

This section develops the recovery theory in three steps. First, we give
a deterministic lemma showing that uniform row-wise accuracy and
population-profile separation imply exact \(K\)-means recovery. Second,
we prove a uniform two-to-infinity error bound for the smoothed
block-profile estimator \(\widetilde{\mathbf P}\). Third, we combine
the two results with concentration of the realized community sizes to
obtain exact recovery.

The analysis differs from the original neighborhood-smoothing theory of
\citet{zhang2017estimating} in three respects. Edge probabilities may
vanish through the sparsity factor \(\gamma_n\); the directed
neighborhood and averaging operations are asymmetric; and the
estimation result must hold uniformly over rows to support exact
recovery.

We begin with the deterministic recovery condition used in the
subsequent probabilistic analysis.

\begin{lemma}[Row-wise separation implies exact \(K\)-means recovery]
\label{lem:deterministic_clustering}
Let \(\widehat{\mathbf P}\) be an estimator of \(\mathbf P\), and suppose
that \(\mathbf P\) has exactly \(K_n\) distinct rows corresponding to
the true communities. Let
\[
\widehat{\mathbf C}
\in
\argmin_{\mathbf C\in\mathcal C_{K_n}}
\|\mathbf C-\widehat{\mathbf P}\|_F
\]
be a global \(K\)-means solution, and let \(\widehat\pi\) be its induced
community assignment. Define
\[
d_{\mathbf P}^*
=
\min_{\pi(i)\neq\pi(j)}
\|\mathbf P_{i\cdot}-\mathbf P_{j\cdot}\|_2,
\qquad
\beta
=
\|\widehat{\mathbf P}-\mathbf P\|_{2\to\infty}.
\]
If
\[
\beta
<
\frac{d_{\mathbf P}^*}{8}
\sqrt{\frac{n_{\min}}{n}},
\]
then
\[
A(\pi,\widehat\pi)=1.
\]
\end{lemma}

The proof is given in the Supplementary Material. The factor
\(\sqrt{n_{\min}/n}\) prevents a global \(K\)-means optimum from
sacrificing a small community in order to reduce the objective within
larger communities. Under the displayed condition, one fitted centroid
is forced near each population row, and every estimated row is assigned
to the centroid associated with its true community.

We next state the assumptions required to verify this deterministic condition with high probability under the directed SBM.

\begin{assumption}
\label{aasump:condition_epsilon_pi}
There exist fixed constants \(C_\rho>C_\pi>0\) and an integer
\(n_\rho\) such that, for all \(n\geq n_\rho\),
\[
\rho_{\min}
\geq
C_\rho\sqrt{\frac{\log n}{n}},
\qquad
\epsilon_\pi
=
C_\pi\sqrt{\frac{\log n}{n}},
\]
and
\[
\frac{n\epsilon_\pi^2}{2}
-
\log(2K_n)
\longrightarrow\infty.
\]
\end{assumption}
This assumption ensures that every community contains enough vertices
for the empirical quantile neighborhood to include sufficiently many
same-community candidates. In particular, a multiplicative Chernoff
bound gives
\[
\mathbb P\left\{
n_{\min}<\frac{n\rho_{\min}}{2}
\right\}
\leq
K_n\exp\left(-\frac{n\rho_{\min}}{8}\right)
=o(1).
\]
Thus, no separate community-size assumption is required.

\begin{assumption}
\label{assump:gammaBrho}
Let \(C_1\) be the fixed constant in
Corollary~\ref{cor:two_to_infinity}. There exists a fixed constant
\(C_{\mathrm{sep}}>16C_1^{1/2}\) and an integer \(n_{\mathrm{sep}}\)
such that, for all \(n\geq n_{\mathrm{sep}}\),
\[
\gamma_n d_{\mathbf B}^*\rho_{\min}
\geq
C_{\mathrm{sep}}
\left(\frac{\log n}{n}\right)^{1/4}.
\]
\end{assumption}

The relevant signal quantity is
\(\gamma_n\rho_{\min}d_{\mathbf B}^*\). When \(K_n\) grows, both
\(\rho_{\min}\) and \(d_{\mathbf B}^*\) may decrease, making recovery
more demanding.

\subsection{Comparison with classical exact-recovery thresholds}
\label{subsec:threshold-comparison}

To compare Assumption~\ref{assump:gammaBrho} with classical
exact-recovery thresholds, suppose that \(K_n=K\) is fixed, the
community proportions are bounded away from zero, and
\(d_{\mathbf B}^*>0\) is fixed. Then the assumption reduces, up to
constants, to
\[
\gamma_n
\gtrsim
\left(\frac{\log n}{n}\right)^{1/4}.
\]
Define the outgoing degree by
\[
d_i^{\mathrm{out}}
:=
\sum_{j\neq i}A_{ij}.
\]
If the weighted outgoing block intensities
\[
\sum_{b=1}^{K}\rho_b B_{ab},
\qquad
a\in[K],
\]
are bounded above and away from zero, then concentration of the
realized community proportions implies, with probability tending to
one,
\[
\mathbb E\!\left(
d_i^{\mathrm{out}}
\mid\mathbf Z
\right)
\asymp
n\gamma_n
\]
uniformly over \(i\). The recovery condition therefore corresponds to
\[
\min_{i\in[n]}
\mathbb E\!\left[
d_i^{\mathrm{out}}
\mid
\mathbf Z
\right]
\gtrsim
n^{3/4}(\log n)^{1/4}.
\]

This is denser than the classical fixed-\(K\) exact-recovery scale. In
the balanced two-community model with
\[
p=\alpha\frac{\log n}{n},
\qquad
q=\beta\frac{\log n}{n},
\]
the sharp boundary is
\(
(\sqrt{\alpha}-\sqrt{\beta})^2=2
\)
\citep{abbe2015exact}, while the general fixed-\(K\) boundary is
characterized by the Chernoff-Hellinger divergence
\citep{abbe2015recovering,abbe2017community}. Thus,
Theorem~\ref{thm:exact_recovery} provides a sufficient guarantee for
asymmetric directed SBMs with potentially growing \(K_n\), rather than
a characterization of the information-theoretic boundary.

Unless stated otherwise, all probabilities below are joint over the
latent labels \(\mathbf Z\) and the directed edges. Edge-concentration
bounds are first established conditionally on \(\mathbf Z\). Because
the resulting bounds are uniform over label realizations, integrating
over \(\mathbf Z\) yields the corresponding unconditional statements.

\begin{theorem}\label{thm:rowwise_bound}
Consider a directed network
\[
G
\sim
\mathrm{DSBM}
(n,K_n,\boldsymbol\rho,\mathbf B,\gamma_n),
\]
where
\(\boldsymbol\rho=(\rho_1,\ldots,\rho_{K_n})\),
\(\rho_{\min}>0\), and
\(\mathbf B\in[0,1]^{K_n\times K_n}\) may vary with \(n\).
Suppose that Assumption~\ref{aasump:condition_epsilon_pi} holds.
Let \(\widetilde{\mathbf P}\) be constructed according to
\eqref{P_tilde_Estimate}, using the empirical-quantile neighborhoods in
\eqref{eq:neighborhood_i}. Then, for all sufficiently large \(n\), for
any \(\epsilon_m>0\) and \(\epsilon_{AP}>2/n\), provided that
\(h<\rho_{\min}-\epsilon_\pi\), the following holds:
\begin{align}
\label{final 2inf bound}
\frac{1}{n}
\|\widetilde{\mathbf P}_{i\cdot}-\mathbf P_{i\cdot}\|_2^2
\leq
\frac{2}{|N_i|}
+
2\epsilon_m
+
\frac{4}{n}
+
16\epsilon_{AP},
\qquad
\text{for all }i\in[n],
\end{align}
with probability at least
\begin{equation}
\label{tota_bound}
\begin{aligned}
1
&-2K_n
\exp\left\{
-\frac{n\epsilon_\pi^2}{2}
\right\}\\
&-2n^2
\exp\left\{
-\frac{n(\epsilon_{AP}-2/n)^2}
{4(1+\epsilon_{AP})}
\right\}\\
&-n(n-1)
\exp\left\{
-\frac{n\epsilon_m^2}{4(1+\epsilon_m)}
\right\}.
\end{aligned}
\end{equation}
\end{theorem}

Assumption~\ref{aasump:condition_epsilon_pi}, together with the
following conditions on \(\epsilon_m\) and \(\epsilon_{AP}\), ensures
that the probability lower bound in \eqref{tota_bound} converges to one.
\begin{enumerate}[label=(\alph*)]
\item
\begin{equation}\label{assump:condition_epsilon_AP}
\frac{\frac{1}{4} n (\epsilon_{AP} - \frac{2}{n})^2}{1 + \epsilon_{AP}} - 2 \log n = \omega(1).
\end{equation}
\item
\begin{equation} \label{assump:condition_epsilon_m}
\frac{n\epsilon_m^2}{4(1+\epsilon_m)}
-
\log\!\bigl(n(n-1)\bigr)
=
\omega(1).
\end{equation}
\end{enumerate}

\begin{corollary}
\label{cor:two_to_infinity}
Under the assumptions of Theorem~\ref{thm:rowwise_bound}, let
\[
h
=
C_h\sqrt{\frac{\log n}{n}},
\qquad
0<C_h<C_\rho-C_\pi.
\]
Then there exists a fixed constant \(C_1>0\), independent of \(n\),
such that
\begin{equation}
\label{eq:2infty_bound}
\mathbb P\left\{
\|\mathbf P-\widetilde{\mathbf P}\|_{2\to\infty}
\geq
C_1^{1/2}(n\log n)^{1/4}
\right\}
\longrightarrow0.
\end{equation}
\end{corollary}

Equivalently,
\[
\frac{1}{\sqrt n}
\|\widetilde{\mathbf P}-\mathbf P\|_{2\to\infty}
=
\mathcal O_P\!\left\{
\left(\frac{\log n}{n}\right)^{1/4}
\right\}
=
o_P(1).
\]
Thus, every outgoing block profile is estimated consistently in
normalized Euclidean distance, or equivalently in per-coordinate
root-mean-square error. The same row-wise result immediately yields the
following Frobenius-norm bound.

\begin{corollary}[Frobenius-norm bound]
\label{cor:frobenius_bound}
Under the assumptions of Theorem~\ref{thm:rowwise_bound} and
Corollary~\ref{cor:two_to_infinity},
\begin{equation}
\label{eq:frobenius_bound}
\mathbb P\!\left\{
\|\mathbf P-\widetilde{\mathbf P}\|_F
\geq
C_1^{1/2}n^{3/4}(\log n)^{1/4}
\right\}
\longrightarrow0,
\end{equation}
where \(C_1\) is the constant in
Corollary~\ref{cor:two_to_infinity}.
\end{corollary}
The preceding results control both the maximum normalized row-wise error
and the total Frobenius error. Exact recovery additionally requires the
distinct population rows to remain sufficiently separated relative to
the row-wise estimation error. We next verify this separation condition
under the directed SBM.

A single minimum-community-size event controls all population
separations simultaneously. Specifically, the multiplicative Chernoff
bound in Assumption~\ref{aasump:condition_epsilon_pi} implies
\[
\mathbb P\left\{
n_{\min}\geq\frac{n\rho_{\min}}{2}
\right\}
\longrightarrow1.
\]
On this event, equation~\eqref{P_p_nul_B relation} gives
\[
d_{\mathbf P}^*
\geq
\gamma_n
\sqrt{\frac{n\rho_{\min}}{2}}\,
d_{\mathbf B}^*.
\]
No additional union bound over pairs of communities is needed, because
the same event \(n_{\min}\geq n\rho_{\min}/2\) yields this inequality
simultaneously for every pair of distinct block rows.

The raw population separation alone is not the complete global
\(K\)-means requirement. Lemma~\ref{lem:deterministic_clustering}
requires
\[
\|\widetilde{\mathbf P}-\mathbf P\|_{2\to\infty}
<
\frac{d_{\mathbf P}^*}{8}
\sqrt{\frac{n_{\min}}{n}}.
\]
Combining the two lower bounds shows that the right-hand side is at
least
\[
\frac{\gamma_n d_{\mathbf B}^*\rho_{\min}\sqrt n}{16}
\]
with probability tending to one. Assumption~\ref{assump:gammaBrho}
makes this quantity dominate the row-wise estimation error in
Corollary~\ref{cor:two_to_infinity}. Thus, for growing \(K_n\), the
relevant sufficient signal is
\(\gamma_n\rho_{\min}d_{\mathbf B}^*\), and shrinking block-row
separation remains a genuine limitation. The following theorem
formalizes the argument.

\begin{theorem}
\label{thm:exact_recovery}
Consider a directed graph \( G \sim \mathrm{DSBM}(n, K_n, \boldsymbol{\rho}, \mathbf{B}, \gamma_n) \) with adjacency matrix \( \mathbf{A} \), where \( \boldsymbol{\rho} = (\rho_1, \ldots, \rho_{K_n}) \) and \( \mathbf{B} \in [0,1]^{K_n \times K_n} \) may vary with \( n \). Assume that Assumptions~\ref{aasump:condition_epsilon_pi}
and~\ref{assump:gammaBrho} hold. Let
\(\widetilde{\mathbf P}\) denote the estimator of the
sparsity-scaled block-profile target \(\mathbf P\), constructed as in
\eqref{P_tilde_Estimate} using
\[
h=C_h\sqrt{\frac{\log n}{n}},
\qquad
0<C_h<C_\rho-C_\pi.
\]
Let \(\widehat{\mathbf C}\in\mathbb R^{n\times n}\) and
\(\widehat\pi:[n]\to[K_n]\) denote the fitted centroid matrix and
community assignment obtained by applying \(K\)-means to the rows of
\(\widetilde{\mathbf P}\), as defined in
\eqref{eq:cluster_projection}. Then
\[
\mathbb P\!\left\{
A(\pi,\widehat\pi)=1
\right\}
\longrightarrow1,
\]
where \(A(\pi,\widehat\pi)\) is defined in \eqref{eq:A_pi}.
\end{theorem}

The proof is provided in the Supplementary Material.

\begin{corollary}[Exact recovery with an estimated number of communities]
\label{cor:estimated-K}
Let \(\widehat K_n\) be an estimator of \(K_n\), and let
\(\widehat\pi_{\widehat K_n}\) denote the clustering obtained by solving
the optimization problem in \eqref{eq:cluster_projection} over
\(\mathcal C_{\widehat K_n}\), that is, with at most
\(\widehat K_n\) fitted centroids. Suppose that the assumptions of
Theorem~\ref{thm:exact_recovery} hold and
\[
\mathbb P(\widehat K_n=K_n)\longrightarrow1.
\]
Then
\[
\mathbb P\left\{
\substack{
\widehat K_n=K_n
\ \text{and there exists }\sigma\in\mathcal S_{K_n}\\
\pi(i)=\sigma\bigl(\widehat\pi_{\widehat K_n}(i)\bigr)
\text{ for every }i\in[n]
}
\right\}
\longrightarrow1.
\]
\end{corollary}

The proof is provided in the Supplementary Material.


\section{Simulation Studies}
\label{sec:simulations}

We examine the finite-sample behavior of the proposed KMP procedure.
The numerical study addresses four questions. First, we compare KMP
with three benchmark procedures across nine independently generated
directed SBM designs. Second, we distinguish partial recovery from exact recovery by reporting
empirical exact-recovery frequencies. Third, we track a fixed initial cohort along nested growing
networks. Fourth, we stress the method by reducing expected degree and
by allowing the number of communities to increase with network size.
All reported SBM simulation experiments use \(MC=100\) paired
replicates.

All KMP simulations use the landmark implementation described in
Section~\ref{subsec:comp_cost}, with
\[
m_n=\min\{128,n\}.
\]
Thus, \(128\) landmarks are used except at the \(n=100\) checkpoint of
the nested experiment, where all \(100\) vertices serve as landmarks.
The numerical results evaluate this scalable landmark procedure,
whereas the exact-recovery theorem concerns the all-witness
dissimilarity in \eqref{eq:distance}.

\subsection{Simulation setup and evaluation}

In the theoretical sections, the number of communities is denoted by
\(K_n\) because it may depend on \(n\). In the simulations, we write
\(K\) when the number of communities is fixed within a design. Except
for the eight-community hub experiment and the controlled growing-\(K_n\)
experiment, all designs use \(K=5\). For every pair \((n,K)\), the
community labels are generated as evenly as possible and then randomly
permuted across vertices. Community sizes therefore differ by at most
one; when \(K\) divides \(n\), every community contains exactly \(n/K\)
vertices. This balanced-label design removes Monte Carlo variation due
solely to random community counts and isolates the clustering behavior
of the four procedures.

Conditional on the labels, all ordered off-diagonal edges are generated
independently according to
\[
A_{ij}\sim
\operatorname{Bernoulli}
\!\left(\gamma_nB_{\pi(i)\pi(j)}\right),
\qquad i\neq j,
\qquad A_{ii}=0.
\]
The primary nine-design comparison uses \(\gamma_n=1\) and
\[
n\in\{200,400,600,800,1000,1200,1500,2000\}.
\]
A new network is generated at every value of \(n\) within each
replicate, so these are independent-size rather than nested-network
comparisons.

Three additional studies examine nested growth, sparsity, and increasing
numbers of communities. The nested study analyzes leading principal
subgraphs at
\[
n\in\{100,200,400,600,800,1200,1500\}
\]
while tracking the initial \(100\) vertices. The sparsity study uses
weak-layered \(D\) over
\[
n\in\{400,600,800,1000,1200,1500,2000\}
\]
under
\[
\gamma_n\in
\left\{
1,\,
\left(\frac{\log n}{n}\right)^{1/4},\,
\min\left(1,\frac4{\sqrt n}\right),\,
\min\left(1,\frac{10\log n}{n}\right)
\right\}.
\]
The growing-\(K_n\) study compares fixed \(K=5\) with
\(K_n=\lceil\log n\rceil\) through \(n=3000\), while approximately
matching density and minimum row separation. The corresponding
subsections give the complete constructions and interpretations.

All experiments use the oracle number of communities. KMA applies
\(K\)-means directly to the outgoing adjacency rows. Spectral-LR
clusters the concatenated coordinates
\[
[\widehat{\mathbf U}_{K},\widehat{\mathbf V}_{K}],
\]
where \(\widehat{\mathbf U}_{K}\) and
\(\widehat{\mathbf V}_{K}\) are the leading \(K\) left and right
singular vectors of \(\mathbf A\). The implemented D-SCORE benchmark uses the corresponding truncated
left-right singular-vector ratio construction, with truncation at
\(\log n\). We include D-SCORE because it is a prominent theoretically
analyzed procedure for directed degree-corrected block models. The
theorem established by \citet{wang2020spectral} controls the number of
misclustered vertices rather than proving exact recovery of all labels.
The two spectral benchmarks use both incoming and outgoing
singular-vector information, whereas KMA and KMP use outgoing rows.

All \(K\)-means fits use \(20\) \(K\)-means++ initializations, and the
fit with the smallest within-cluster sum of squares is retained. The
recovery theorem is stated for a global \(K\)-means minimizer, whereas
the numerical implementation uses the best solution returned by these
\(20\) Lloyd-algorithm runs and is therefore a practical approximation
to the optimization step assumed in the theory. The base random seed is
\(20260708\).

For KMP,
\[
h=\sqrt{\frac{\log n}{n}},
\qquad
m_n=\min\{128,n\}.
\]
The landmarks are sampled uniformly without replacement. For each
candidate pair \((i,j)\), landmarks equal to \(i\) or \(j\) are
excluded from the witness maximum, matching the endpoint-exclusion
convention in \eqref{eq:distance}. The corresponding nominal
neighborhood sizes
\(s_n=\lceil h(n-1)\rceil\) are 33, 49, 62, 74, 84, 93, 105, and
\(124\) for \(n=200,400,600,800,1000,1200,1500,\) and \(2000\),
respectively; at \(n=3000\), the nominal size is \(155\). All ties at the empirical quantile are retained,
and no upper cap is imposed. Consequently, the realized neighborhoods
may be substantially larger than their nominal sizes when the landmark
dissimilarities are discrete or highly tied.

The landmark procedure otherwise uses the same asymmetric row-averaging
estimator,
\[
\widetilde P_{ij}
=
\frac1{|N_i|}
\sum_{u\in N_i}A_{uj},
\]
and no symmetrization is applied before clustering.

We report mean adjusted Rand index (ARI), label-matched
misclassification, empirical exact-recovery frequency, paired ARI wins,
neighborhood purity, normalized row-wise profile error, and runtime.
For two methods, the paired win count is the number of Monte Carlo
replicates in which the first method has strictly larger ARI. For node
\(i\), neighborhood purity is
\[
\operatorname{purity}(N_i)
=
\frac{1}{|N_i|}
\sum_{u\in N_i}
\mathbbm{1}\{Z_u=Z_i\},
\]
and we average this quantity over vertices and replicates. The
normalized profile error is
\[
e_{2\to\infty}
=
\frac1{\sqrt n}
\|\widetilde{\mathbf P}-\mathbf P\|_{2\to\infty},
\]
which is the finite-sample counterpart of the normalized error in
Corollary~\ref{cor:two_to_infinity}.

For independently generated networks, the empirical global
exact-recovery rate is
\[
\widehat p_{\mathrm{exact}}(n)
=
\frac1{MC}
\sum_{b=1}^{MC}
\mathbbm{1}
\left\{
A\bigl(\pi_n^{(b)},\widehat\pi_n^{(b)}\bigr)=1
\right\}.
\]
The same \(100\) paired network realizations are used for all methods at
each setting, so empirical exact-recovery rates have resolution
\(0.01\). ARI figures display one Monte Carlo standard deviation; the
standard error of a displayed mean is the corresponding standard
deviation divided by \(10\).

\subsection{Block-matrix designs and diagnostics}
\label{subsec:block-designs}

The nine primary designs form three empirical regimes. The first is a
\emph{weak-spectrum, separated-profile regime}, represented by
weak-layered \(D\), weak-profile \(C\), and weak-cross \(B\). These
matrices have similar mean edge probabilities, distinct outgoing block
rows, and only a few comparatively strong singular directions. They are
used to examine whether smoothing complete outgoing profiles can be
advantageous when some sample singular directions are difficult to
estimate.

The second is a \emph{heterogeneous or weak-separation regime},
represented by weak-irregular \(A\), the strong-asymmetric matrix, and
the two hub settings. These examples introduce distinct obstacles to
smoothing, including informative outgoing-degree variation, smaller
minimum row separation, persistent neighborhood contamination, and an
increased number of communities. Weak-irregular \(A\) is particularly
useful as a bridge design: it has weak trailing singular values but also
stronger degree information than the three separated-profile examples.

The third is a \emph{well-conditioned, strongly separated regime},
represented by the cyclic-banded and matched-separation matrices. These
designs serve as controls with large row separation and well-conditioned
spectra.

The regimes are not matched in every structural characteristic.
Accordingly, performance comparisons are made within a block matrix,
while cross-regime conclusions concern qualitative patterns associated
with outgoing-row separation, spectral conditioning, degree
information, neighborhood quality, and the number of communities.

For a block matrix \(\mathbf B\), define
\[
\overline B
=
K^{-2}\sum_{r,s=1}^K B_{rs},
\qquad
a(\mathbf B)
=
\frac{\|\mathbf B-\mathbf B^\top\|_F}{\|\mathbf B\|_F},
\qquad
\delta_{\mathrm{row}}(\mathbf B)
=
\min_{r\neq s}
\|\mathbf B_{r\cdot}-\mathbf B_{s\cdot}\|_2.
\]
Table~\ref{tab:design-diagnostics} reports these quantities.

\begin{table}[t]
\centering
\caption{Structural diagnostics for the primary block matrices.}
\label{tab:design-diagnostics}
\small
\resizebox{\textwidth}{!}{%
\begin{tabular}{@{}lrrrrl@{}}
\toprule
Design
& \(K\)
& \(\overline B\)
& \(a(\mathbf B)\)
& \(\delta_{\mathrm{row}}(\mathbf B)\)
& Trailing singular values\\
\midrule
Weak-layered \(D\)
&5&0.110&0.641&0.119&\((0.033,0.007,0.002)\)\\
Weak-profile \(C\)
&5&0.111&0.595&0.106&\((0.025,0.007,0.001)\)\\
Weak-cross \(B\)
&5&0.110&0.719&0.109&\((0.031,0.005,<0.001)\)\\
Weak-irregular \(A\)
&5&0.110&0.696&0.114&\((0.021,0.005,0.004)\)\\
Strong asymmetric
&5&0.097&0.703&0.088&\((0.066,0.024,0.010)\)\\
Five-community hub
&5&0.109&0.992&0.128&\((0.101,0.086,0.008)\)\\
Eight-community hub
&8&0.111&0.939&0.120&
\((0.158,0.100,0.083,0.072,0.062,0.044)\)\\
Cyclic-banded
&5&0.389&0.515&0.397&\((0.417,0.212,0.204)\)\\
Matched-separation asymmetric
&5&0.389&0.641&0.397&\((0.417,0.212,0.204)\)\\
\bottomrule
\end{tabular}%
}
\end{table}

The neighborhood-purity and normalized profile-error diagnostics defined
above are interpreted jointly. Weak singular directions alone do not
imply that smoothing will dominate degree- or spectral-based procedures:
high purity can coexist with insufficient between-row separation, while
a small absolute profile error can be uninformative when the probability
scale and population row distances also shrink.

The matched-separation control is especially informative. Writing
\[\mathbf B_{\mathrm{match}}=\mathbf B_{\mathrm{cyc}}\mathbf Q\] 
for a permutation matrix \(\mathbf Q\), right multiplication by \(\mathbf Q\)
preserves all Euclidean distances between rows and all singular values.
The matched and cyclic-banded designs therefore differ in asymmetry but
not in the two geometric quantities most directly tied to the present
recovery argument. Their comparison helps separate the effect of
asymmetry from the effects of weak row separation and poor spectral
conditioning.

\subsection{Nested growing-network design}
\label{subsec:nested-growth-design}

The nested experiment is designed to answer a different question from
the independent-size comparison. Instead of regenerating a complete
network at every value of \(n\), each replicate begins with a graph of
size \(1500\) and reveals its leading principal subgraphs. The four
settings are weak-layered \(D\), weak-profile \(C\), weak-irregular
\(A\), and cyclic-banded, spanning the favorable weak-spectrum regime,
a degree-informative contrast, and a well-conditioned control.

Let \(\mathcal I_0=\{1,\ldots,100\}\) denote the initial cohort. At
each checkpoint, every method is refitted to the entire observed graph,
and labels are aligned using the permutation maximizing full-graph
agreement. For replicate \(b\), define the number of errors in the fixed
cohort by
\[
M_{\mathcal I_0}^{(b)}(n)
=
\sum_{i\in\mathcal I_0}
\mathbbm{1}
\left\{
\pi^{(b)}(i)
\neq
\widehat\sigma_n^{(b)}
\bigl(\widehat\pi_n^{(b)}(i)\bigr)
\right\},
\]
and define the fixed-cohort exact-recovery rate by
\[
\widehat p_{\mathrm{cohort}}(n)
=
\frac1{MC}
\sum_{b=1}^{MC}
\mathbbm{1}
\left\{
M_{\mathcal I_0}^{(b)}(n)=0
\right\}.
\]
We report both the mean cohort error and the fixed-cohort exact-recovery
rate. Because the cohort contains only \(100\) vertices while the full
network grows, fixed-cohort exact recovery is weaker than global exact
recovery. It is used to measure whether additional network information
improves the classification of previously observed vertices and is not
a substitute for the global criterion in
Theorem~\ref{thm:exact_recovery}.

\subsection{Independent-size recovery}
\label{subsec:independent-size-results}

Table~\ref{tab:primary-weak-ari} summarizes mean ARI at \(n=1500\).
The complete trajectories, including \(n=2000\), are given in the
Supplementary Material.

\begin{table}[t]
\centering
\caption{Independent-size mean ARI at \(n=1500\), based on \(MC=100\)
paired replicates.}
\label{tab:primary-weak-ari}
\small
\begin{tabular}{@{}lrrl@{}}
\toprule
Design & KMP & Best competitor & Competitor\\
\midrule
Weak-layered \(D\) & \textbf{0.842} & 0.746 & KMA\\
Weak-profile \(C\) & \textbf{0.884} & 0.770 & KMA\\
Weak-cross \(B\) & \textbf{0.837} & 0.739 & KMA\\
Weak-irregular \(A\) & \textbf{0.812} & 0.775 & KMA\\
Strong asymmetric & 0.802 & \textbf{0.939} & D-SCORE\\
Five-community hub & 0.419 & \textbf{0.999} & D-SCORE\\
Eight-community hub & 0.146 & \textbf{0.922} & D-SCORE\\
Cyclic-banded & \textbf{1.000} & \textbf{1.000} & all competitors\\
Matched-separation asymmetric & \textbf{1.000} & \textbf{1.000} & all competitors\\
\bottomrule
\end{tabular}
\end{table}

\begin{figure}[p]
\centering
\begin{subfigure}[t]{0.49\textwidth}
\centering
\vspace{0pt}
\includegraphics[width=\textwidth]{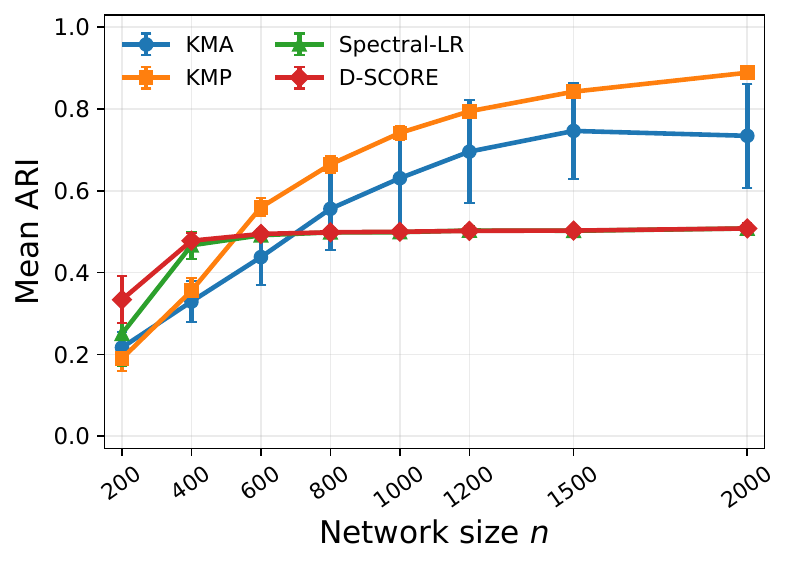}
\caption{Weak-layered \(D\).}
\end{subfigure}
\hfill
\begin{subfigure}[t]{0.49\textwidth}
\centering
\vspace{0pt}
\includegraphics[width=\textwidth]{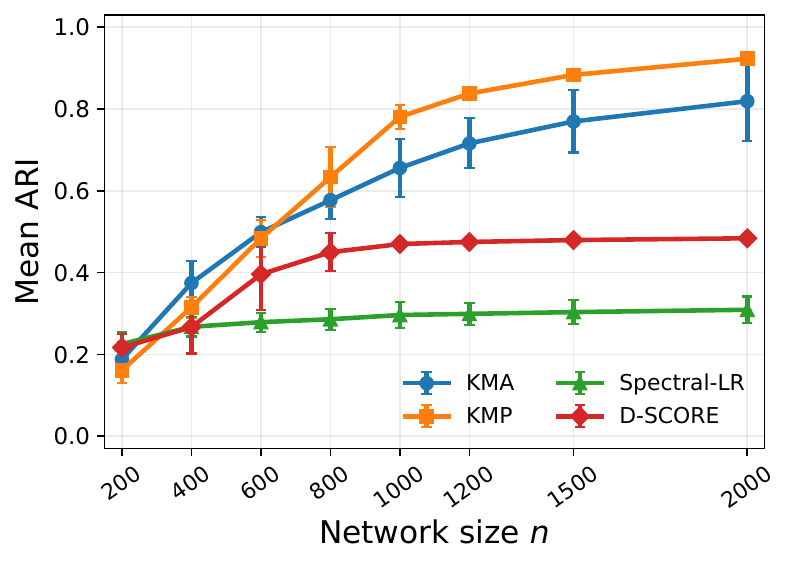}
\caption{Weak-profile \(C\).}
\end{subfigure}

\begin{subfigure}[t]{0.49\textwidth}
\centering
\vspace{0pt}
\includegraphics[width=\textwidth]{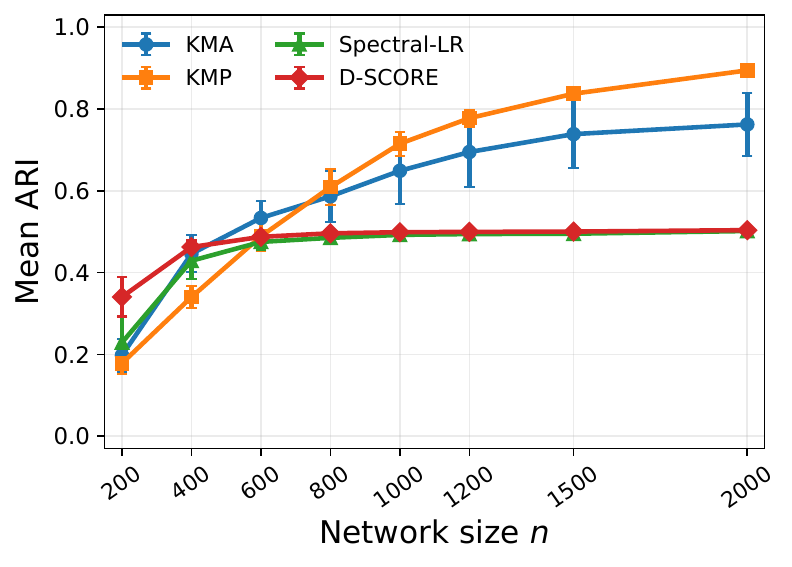}
\caption{Weak-cross \(B\).}
\end{subfigure}
\hfill
\begin{subfigure}[t]{0.49\textwidth}
\centering
\vspace{0pt}
\includegraphics[width=\textwidth]{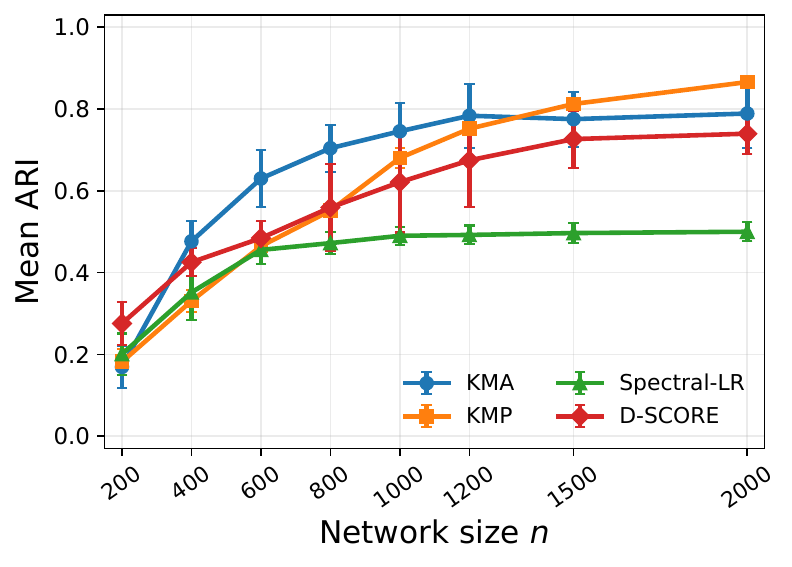}
\caption{Weak-irregular \(A\).}
\end{subfigure}
\caption{Mean ARI with one Monte Carlo standard-deviation error bar in
the three weak-spectrum designs and the weak-irregular contrast, based
on \(MC=100\) paired replicates.}
\label{fig:ari-favorable-designs}
\end{figure}

\paragraph{Weak-spectrum, separated-profile regime.}
KMP becomes the strongest method at moderate or large \(n\) in the
three weak-spectrum designs and, at larger \(n\), in the weak-irregular
contrast. At \(n=1500\), it beats KMA in \(91\), \(88\), \(96\), and
\(93\) of the \(100\) paired replicates for weak-layered, weak-profile,
weak-cross, and weak-irregular, respectively. It also beats both
spectral procedures in all \(100\) replicates for each of these four
designs. Thus, the favorable finding is not a small-\(MC\) artifact.

The common feature of weak-layered \(D\), weak-profile \(C\), and
weak-cross \(B\) is not a particular block pattern. Rather, the designs
combine distinct outgoing profiles with several weak singular
directions. In these designs, smoothing provides a distinct practical advantage
when the complete outgoing profiles can be estimated more reliably than
the community-relevant singular directions extracted from a single
noisy adjacency matrix. Weak-irregular \(A\) qualifies this conclusion:
its KMP advantage emerges later and is smaller, while its raw adjacency
rows retain more informative degree variation.

\paragraph{Heterogeneous or weak-separation regime.}
D-SCORE is strongest in the strong-asymmetric and hub settings.
Informative outgoing degree can favor KMA, while small row separation
or persistent neighborhood contamination can cause smoothing to mix
distinct population profiles. The eight-community hub additionally
shows that a larger number of groups can make the quantile
neighborhoods difficult to purify even when the minimum row distance is
not exceptionally small.

\paragraph{Well-conditioned, strongly separated regime.}
All methods recover rapidly in the cyclic-banded and matched-separation
controls. When outgoing rows are widely separated and the relevant singular
directions are well conditioned, smoothing offers little advantage, and
KMA or the spectral procedures often reach high empirical
exact-recovery rates earlier. The close behavior of the two controls also shows
that pronounced asymmetry is not itself an obstacle when separation and
conditioning are favorable.

\begin{figure}[p]
\centering
\begin{subfigure}[t]{0.49\textwidth}
\centering
\vspace{0pt}
\includegraphics[width=\textwidth]{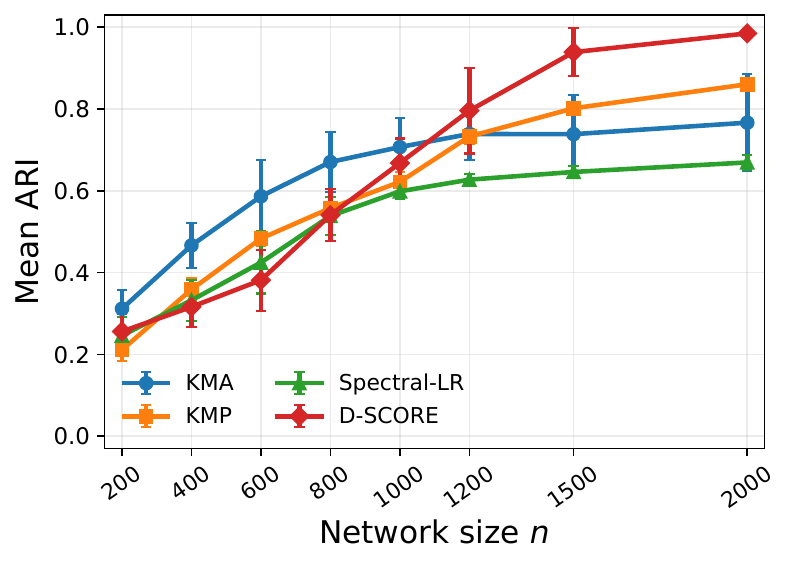}
\caption{Strong asymmetric.}
\end{subfigure}
\hfill
\begin{subfigure}[t]{0.49\textwidth}
\centering
\vspace{0pt}
\includegraphics[width=\textwidth]{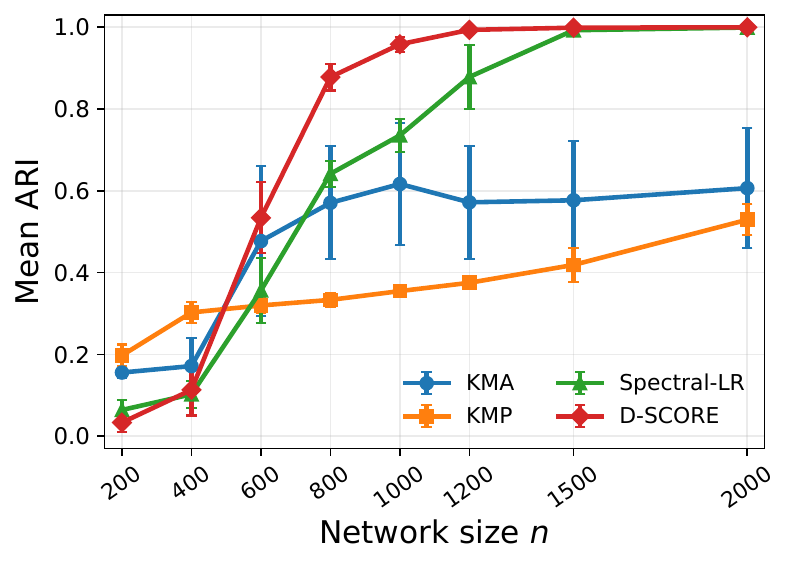}
\caption{Five-community hub.}
\end{subfigure}

\begin{subfigure}[t]{0.49\textwidth}
\centering
\vspace{0pt}
\includegraphics[width=\textwidth]{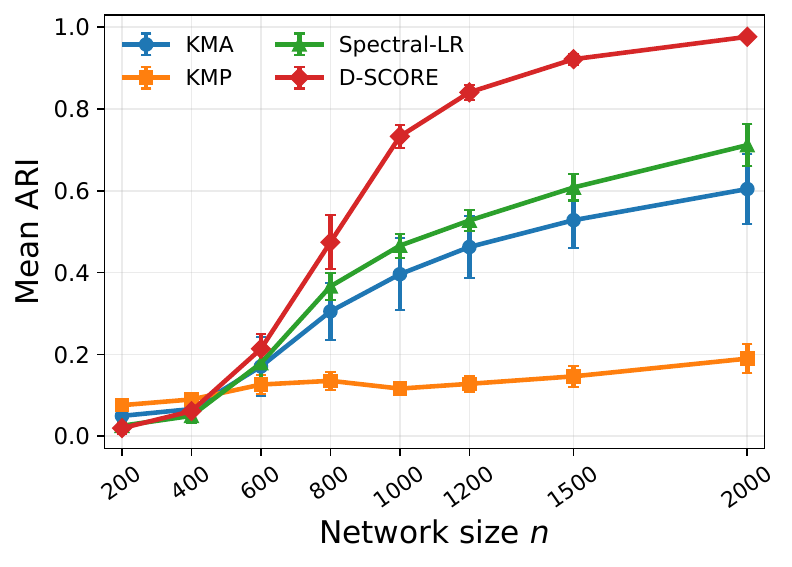}
\caption{Eight-community hub.}
\end{subfigure}
\hfill
\begin{subfigure}[t]{0.49\textwidth}
\centering
\vspace{0pt}
\includegraphics[width=\textwidth]{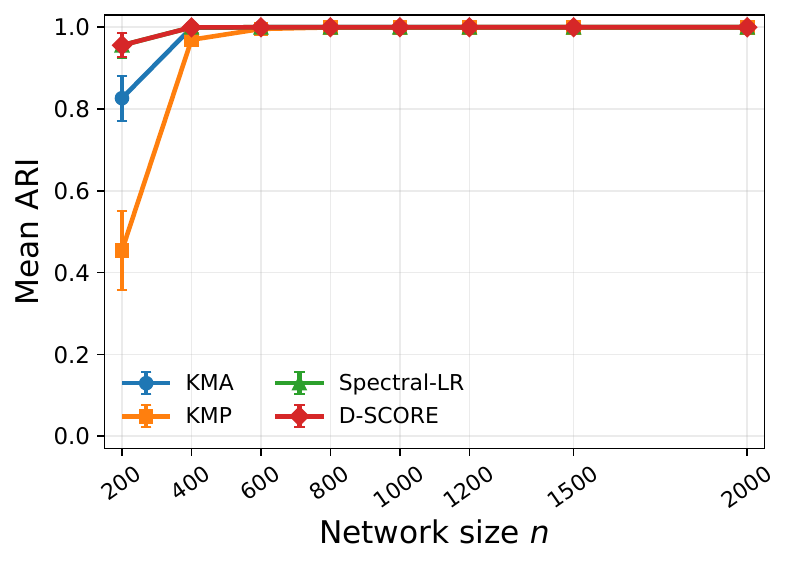}
\caption{Cyclic-banded control.}
\end{subfigure}
\caption{Representative contrast designs, based on \(MC=100\) paired
replicates.}
\label{fig:ari-weak-controls}
\end{figure}

In the matched-separation control, the KMP exact-recovery rate is
\(0.35\) at \(n=600\), \(0.85\) at \(n=800\), \(0.98\) at
\(n=1000\), \(0.99\) at \(n=1200\), and \(1.00\) from \(n=1500\)
onward. By contrast, the strong-asymmetric design, whose minimum row
separation is \(0.088\) rather than \(0.397\), remains substantially
more difficult. Thus, strong asymmetry is compatible with recovery when
the outgoing profiles remain well separated.

\subsection{Empirical exact recovery}

The same \(MC=100\) primary experiment is used to examine empirical
exact recovery. Table~\ref{tab:exact-recovery-extended} summarizes the main
transitions; the complete table is in the Supplementary Material.

\begin{table}[t]
\centering
\caption{Summary of empirical global exact-recovery transitions, based
on \(MC=100\) paired replicates.}
\label{tab:exact-recovery-extended}
\small
\begin{tabularx}{\textwidth}{@{}p{0.25\textwidth}p{0.34\textwidth}Y@{}}
\toprule
Design group & Representative rates & Interpretation\\
\midrule
Cyclic-banded
& At \(n=400\), KMA, KMP, Spectral-LR, and D-SCORE have rates
\(0.84,0.01,0.96,0.96\); KMP reaches \(1.00\) at \(n=1200\).
& Well-conditioned, with a later KMP transition.\\
Matched separation
& KMP rates are \(0.35,0.85,0.98,0.99,1.00\) at
\(n=600,800,1000,1200,1500\).
& Large row separation offsets strong asymmetry.\\
Five-community hub
& At \(n=2000\), Spectral-LR and D-SCORE have rates \(0.61\) and
\(0.92\); KMP remains at \(0\).
& Spectral procedures dominate.\\
Weak-spectrum \(B,C,D\)
& No method recovers globally through \(n=1500\); at \(n=2000\), only
KMA in weak-cross has rate \(0.01\).
& KMP gains are partial-recovery gains.\\
Weak-irregular \(A\)
& KMA has rate \(0.08\) at \(n=2000\); all other rates are zero.
& Isolated exact-recovery events do not imply small average error.\\
\bottomrule
\end{tabularx}
\end{table}

ARI and exact recovery describe different finite-sample properties:
near-perfect average clustering may still leave one or more vertices
misclassified. KMP improves average partial recovery in the difficult weak-spectrum
designs but does not reach an earlier empirical exact-recovery
transition over the displayed range. Conversely, isolated perfect KMA
replicates in weak-cross and weak-irregular occur alongside larger mean
errors, while the well-conditioned and hub designs often favor the
spectral procedures in exact recovery. These patterns also show that strong average clustering performance and
early empirical exact recovery need not occur in the same regime.

\subsection{Sparsity stress test}
\label{subsec:sparsity-stress-test}

The theoretical model permits \(\gamma_n\to0\), but the recovery
condition requires the scaled row separation to remain large relative
to the normalized estimation error. The sparsity experiment keeps the
unscaled weak-layered \(D\) geometry fixed and lowers \(\gamma_n\), so
both edge probabilities and population row distances shrink. It is
therefore a direct finite-sample check of whether the empirical landmark
neighborhoods remain informative at lower expected degrees.

\begin{figure}[p]
\centering
\begin{subfigure}[t]{0.49\textwidth}
\centering
\vspace{0pt}
\includegraphics[width=\textwidth]{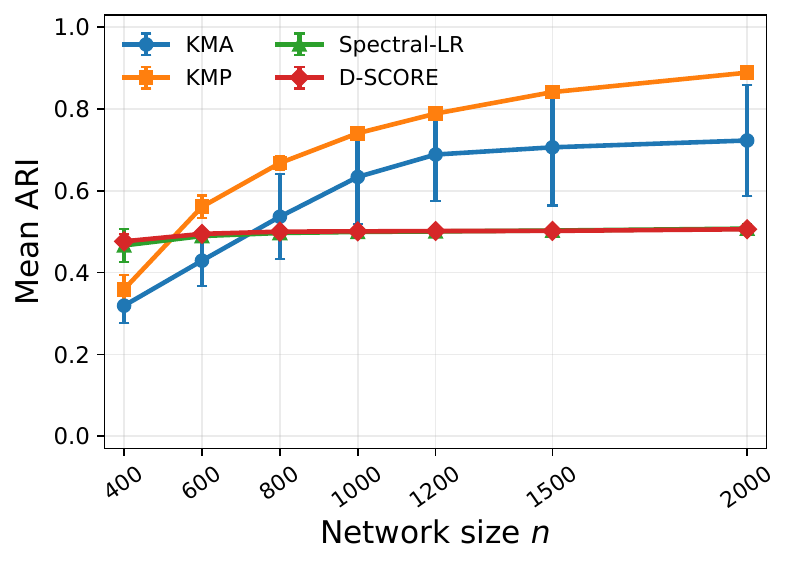}
\caption{Dense.}
\end{subfigure}
\hfill
\begin{subfigure}[t]{0.49\textwidth}
\centering
\vspace{0pt}
\includegraphics[width=\textwidth]{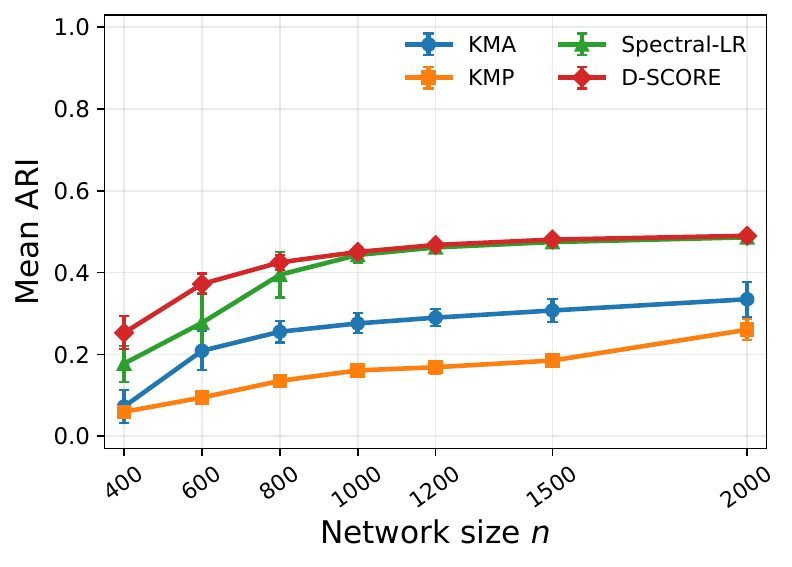}
\caption{Quarter-power.}
\end{subfigure}

\begin{subfigure}[t]{0.49\textwidth}
\centering
\vspace{0pt}
\includegraphics[width=\textwidth]{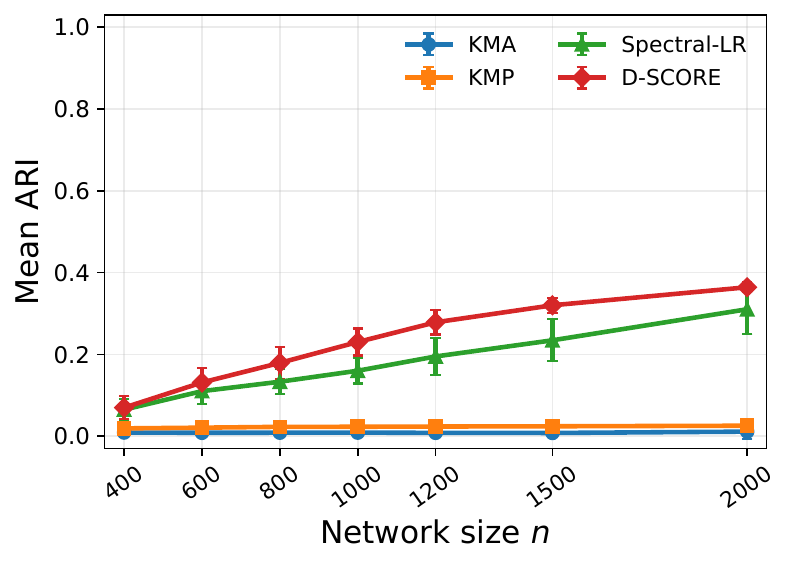}
\caption{Square-root degree.}
\end{subfigure}
\hfill
\begin{subfigure}[t]{0.49\textwidth}
\centering
\vspace{0pt}
\includegraphics[width=\textwidth]{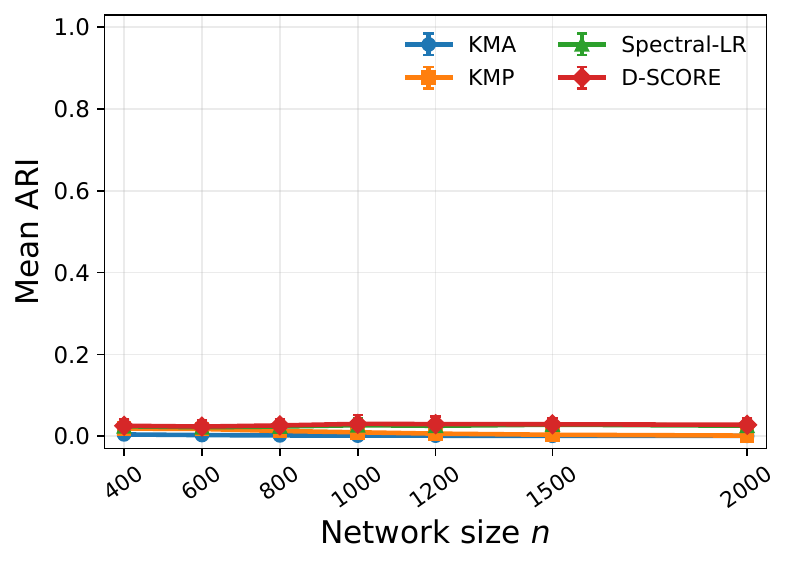}
\caption{Logarithmic degree.}
\end{subfigure}
\caption{Sparsity stress test for the weak-layered \(D\) design,
based on \(MC=100\) paired replicates.}
\label{fig:sparse-ari-mc100}
\end{figure}

Figure~\ref{fig:sparse-ari-mc100} and
Table~\ref{tab:sparse-n2000} show that KMP's dense advantage does not
persist as the expected degree decreases.

The dense regime in this subsection was generated as part of the
separate sparsity-study batch. Although it uses the same block matrix
and \(\gamma_n=1\) as the primary weak-layered experiment, it consists
of a different set of \(MC=100\) network realizations. Its Monte Carlo
summaries therefore need not agree exactly with those from the primary
simulation batch.
\begin{table}[t]
\centering
\caption{Sparse-design performance at \(n=2000\), based on \(MC=100\)
paired replicates. Mean error is the label-matched misclassification
rate.}
\label{tab:sparse-n2000}
\small
\resizebox{\textwidth}{!}{%
\begin{tabular}{@{}llrrrr@{}}
\toprule
Regime & Method & Mean ARI & Mean error & Exact rate & Median runtime\\
\midrule
Dense & KMA & 0.723 & 0.238 & 0.01 & 2.31\\
 & KMP & \textbf{0.889} & \textbf{0.046} & 0.00 & 22.59\\
 & Spectral-LR & 0.508 & 0.361 & 0.00 & 0.33\\
 & D-SCORE & 0.506 & 0.363 & 0.00 & 0.32\\
\addlinespace[2pt]
Quarter-power & KMA & 0.335 & 0.549 & 0.00 & 0.32\\
 & KMP & 0.260 & 0.464 & 0.00 & 13.50\\
 & Spectral-LR & 0.486 & 0.390 & 0.00 & 0.19\\
 & D-SCORE & \textbf{0.490} & \textbf{0.384} & 0.00 & 0.17\\
\addlinespace[2pt]
Square-root degree & KMA & 0.011 & 0.751 & 0.00 & 0.14\\
 & KMP & 0.025 & 0.710 & 0.00 & 10.93\\
 & Spectral-LR & 0.310 & 0.497 & 0.00 & 0.18\\
 & D-SCORE & \textbf{0.364} & \textbf{0.445} & 0.00 & 0.18\\
\addlinespace[2pt]
Log-degree & KMA & 0.000 & 0.785 & 0.00 & 0.10\\
 & KMP & 0.001 & 0.784 & 0.00 & 10.64\\
 & Spectral-LR & 0.025 & 0.722 & 0.00 & 0.17\\
 & D-SCORE & \textbf{0.028} & \textbf{0.722} & 0.00 & 0.18\\
\bottomrule
\end{tabular}%
}
\end{table}

The runtime values in Table~\ref{tab:sparse-n2000} were collected within
the sparsity-study batch and should therefore be compared across methods
within that experiment, rather than directly with the timings from the
separate primary simulation batch reported below.

The corresponding KMP diagnostics are
\[
\begin{array}{c|rrrr}
\text{Regime}
&\text{Expected degree}
&\text{Purity}
&\text{Tie inflation}
&e_{2\to\infty}\\
\hline
\text{Dense}&220.3&0.920&1.23&0.080\\
\text{Quarter-power}&54.7&0.350&3.48&0.025\\
\text{Square-root degree}&19.7&0.228&9.24&0.013\\
\text{Log-degree}&8.4&0.200&13.99&0.009
\end{array}
\]
As sparsity increases, many landmark dissimilarities tie, realized
neighborhoods become much larger than their nominal size, and their
purity approaches the \(1/5\) level associated with nearly random
mixing across five balanced communities. In the dense regime, KMP has
the largest ARI and smallest mean error, although it still has no global
exact-recovery replicate at \(n=2000\). Under quarter-power sparsity,
Spectral-LR and D-SCORE overtake KMP; under square-root-degree scaling,
D-SCORE is clearly strongest; and under logarithmic-degree scaling, all
methods have very low ARI.

The decreasing value of \(e_{2\to\infty}\) must be interpreted relative
to the shrinking probability scale. A small absolute profile error does
not imply recoverability if the separation between distinct probability
rows shrinks at the same or a faster rate. The purity and tie-inflation
diagnostics provide the complementary explanation: in sparse finite
samples, the landmark maximum becomes too coarse to distinguish many
pairs, and the empirical quantile rule averages across different
communities. Thus, the sparse experiment is consistent with the asymptotic theory:
scaled row separation and the quality of the practical landmark
neighborhoods are both consequential in finite samples.

\subsection{Fixed-cohort recovery in nested growing networks}
\label{subsec:fixed-cohort-results}

\begin{figure}[t]
\centering
\includegraphics[width=0.88\textwidth]{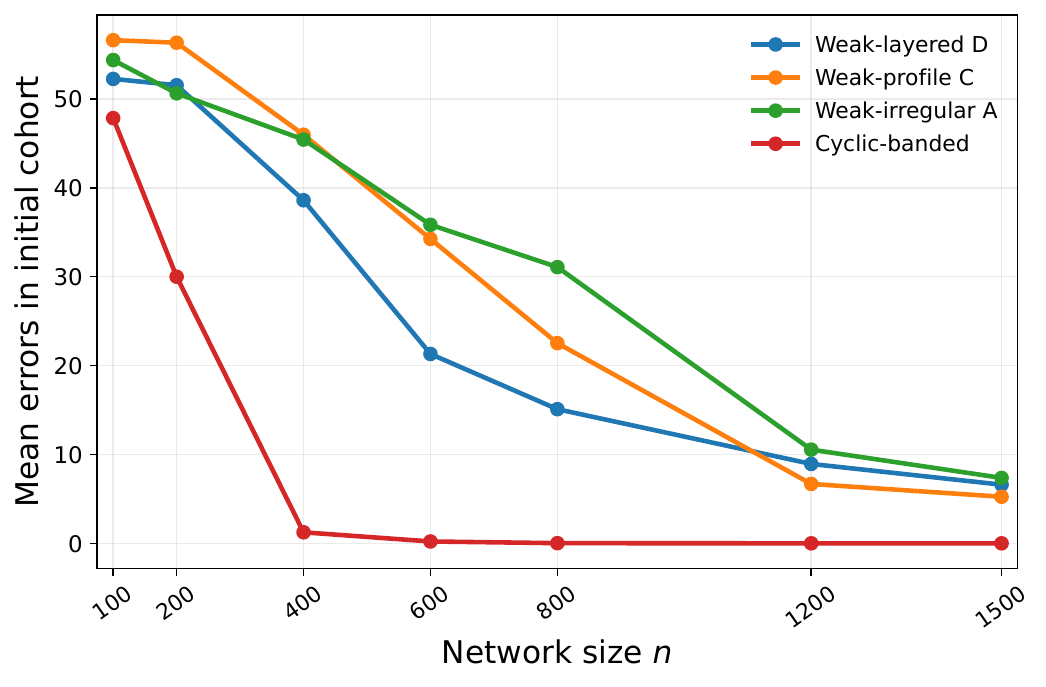}
\caption{Mean number of KMP errors among the initial \(100\) vertices
in the four nested designs, based on \(MC=100\) growing-network paths.}
\label{fig:nested-kmp-cohort}
\end{figure}

Figure~\ref{fig:nested-kmp-cohort} reports the KMP mean error among the
initial \(100\) vertices along the nested growing-network paths.

At \(n=1500\), KMP has the smallest mean cohort error in each difficult
design.

\begin{table}[t]
\centering
\caption{Fixed-cohort performance at \(n=1500\), based on \(MC=100\)
paired growing-network paths. Entries are the mean number of errors
among the initial \(100\), with fixed-cohort exact-recovery rates in
parentheses.}
\label{tab:cohort-n1500-comparison}
\small
\resizebox{\textwidth}{!}{%
\begin{tabular}{@{}lcccc@{}}
\toprule
Design & KMA & KMP & Spectral-LR & D-SCORE\\
\midrule
Weak-layered \(D\)
&24.08 \((0.04)\)
&\textbf{6.60} \((0.00)\)
&36.92 \((0.00)\)
&36.93 \((0.00)\)\\
Weak-profile \(C\)
&19.17 \((0.02)\)
&\textbf{5.24} \((0.00)\)
&50.31 \((0.00)\)
&39.00 \((0.00)\)\\
Weak-irregular \(A\)
&20.35 \((0.06)\)
&\textbf{7.36} \((0.00)\)
&37.25 \((0.00)\)
&20.34 \((0.00)\)\\
Cyclic-banded
&\textbf{0.00} \((1.00)\)
&\textbf{0.00} \((1.00)\)
&\textbf{0.00} \((1.00)\)
&\textbf{0.00} \((1.00)\)\\
\bottomrule
\end{tabular}%
}
\end{table}

The isolated exact-recovery events for KMA occur alongside substantially
larger mean errors. KMP's nested advantage is therefore an improvement
in average partial recovery over this range.

The full KMP trajectories in the Supplementary Material show steady
information accumulation. Its mean cohort errors decrease from
\(52.24\), \(56.60\), and \(54.36\) at \(n=100\) to \(6.60\),
\(5.24\), and \(7.36\) at \(n=1500\) in weak-layered, weak-profile,
and weak-irregular, respectively. In the cyclic-banded control, all
methods eventually recover both the fixed cohort and the complete
observed graph. The nested and independent-size experiments need not
have identical transition probabilities: the former use dependent
principal subgraphs from a common growing path, while the latter
generate a new graph at every network size.

\subsection{Controlled growth in the number of communities}
\label{subsec:growing-k-results}

The controlled experiment keeps \(\gamma_n=1\), mean block probability
\(0.11\), and approximately constant minimum row separation while
comparing fixed \(K=5\) with \(K_n=\lceil\log n\rceil\).

\begin{figure}[t]
\centering
\begin{subfigure}[t]{0.49\textwidth}
\centering
\vspace{0pt}
\includegraphics[width=\textwidth]{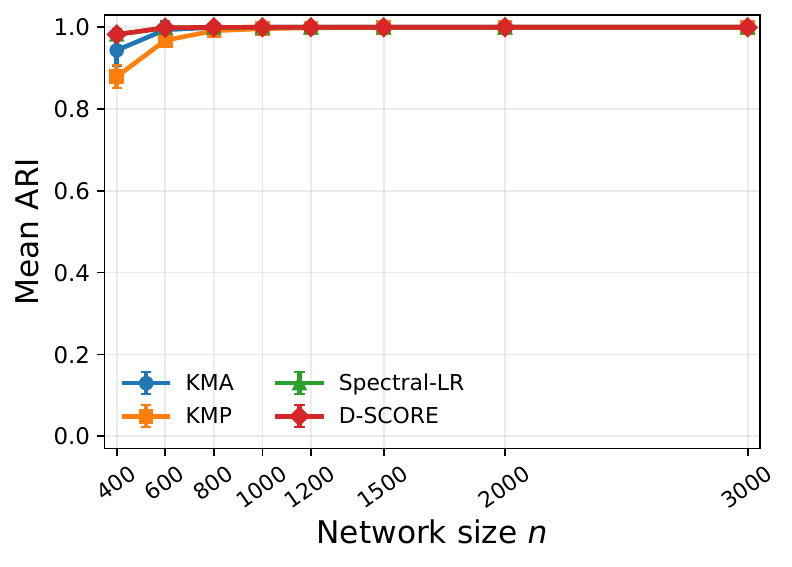}
\caption{Fixed \(K=5\).}
\end{subfigure}
\hfill
\begin{subfigure}[t]{0.49\textwidth}
\centering
\vspace{0pt}
\includegraphics[width=\textwidth]{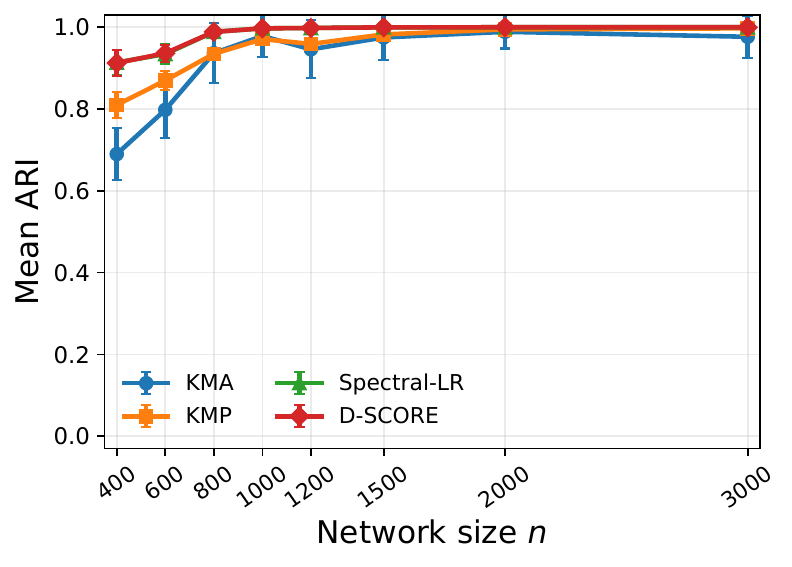}
\caption{\(K_n=\lceil\log n\rceil\).}
\end{subfigure}
\caption{Controlled growing-\(K_n\) experiment, based on \(MC=100\)
paired replicates.}
\label{fig:growing-k-mc100}
\end{figure}

With fixed \(K=5\), all four procedures have exact-recovery rate
\(1.00\) at \(n=3000\). With \(K_n=\lceil\log n\rceil\), the endpoint
has \(K_n=9\). KMP then has mean ARI \(0.997\) and mean
misclassification rate \(0.0012\), but its exact-recovery rate is only
\(0.01\). Spectral-LR and D-SCORE both have ARI and exact-recovery rate
\(1.00\), while KMA has ARI \(0.977\) and exact-recovery rate \(0.82\).

Because density and minimum row separation are approximately matched,
the slower KMP exact-recovery transition cannot be attributed solely to
those quantities. The result highlights the additional roles of
shrinking community sizes, simultaneous classification over more
groups, and the finite-sample purity of the quantile neighborhoods.

\subsection{Computational comparison}

For the dense implementation considered here, the \(m=128\) landmark
approximation reduces the dissimilarity-stage cost from \(O(n^3)\) to
\(O(n^2m)\), although KMP remains slower than the comparison procedures. The following values come from the primary
weak-layered simulation batch and are intended for within-batch
comparison among the four methods. At \(n=2000\), the median runtimes
are \(6.15\) seconds for KMA, \(8.09\) for KMP, \(0.54\) for
Spectral-LR, and \(0.53\) for D-SCORE.

\begin{figure}[t]
\centering
\includegraphics[width=0.88\textwidth]{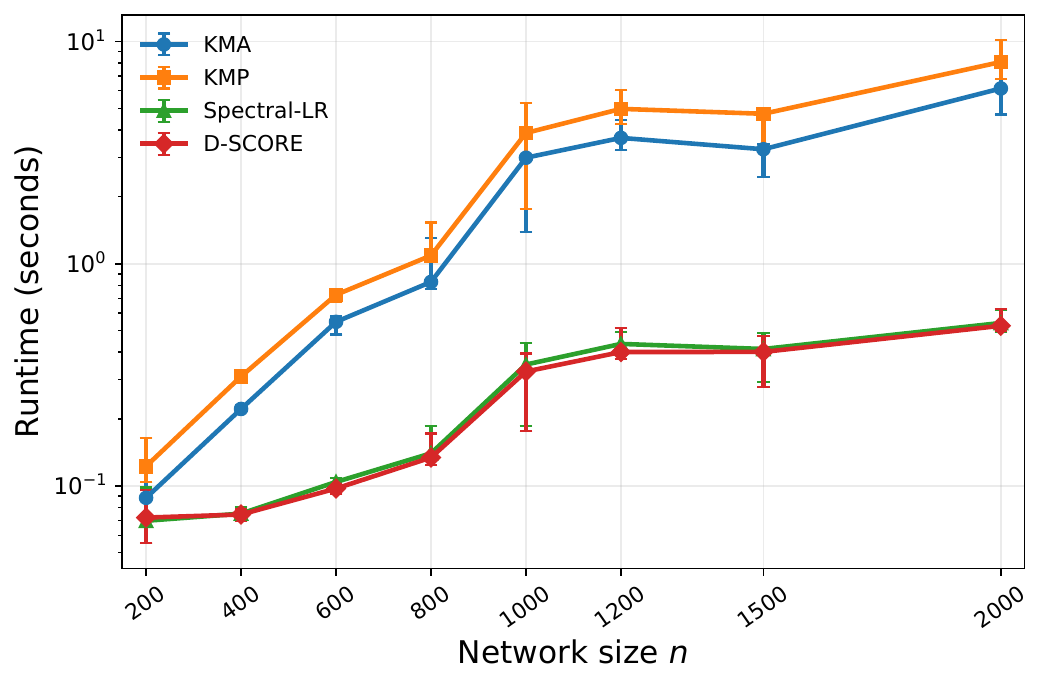}
\caption{Median runtime with interquartile range in weak-layered \(D\),
based on \(MC=100\) paired replicates.}
\label{fig:runtime_weak_D_current}
\end{figure}

Overall, the simulations support KMP as a complementary block-profile
method rather than a universal replacement for spectral clustering. Its clearest benefits occur in dense designs with
separated outgoing profiles and weak community-relevant singular
directions. These benefits concern average partial recovery, weaken
under sparsity or increasing \(K_n\), and come at greater computational
cost than the spectral alternatives. Conversely, informative degree
variation, contaminated neighborhoods, small row separation, or
well-conditioned spectral structure can favor KMA, Spectral-LR, or
D-SCORE.

\FloatBarrier
\section{\textit{C. elegans} connectome application}
\label{sec:real-data}

\subsection{Data and implementation}

We apply the four clustering procedures to the hermaphrodite
\textit{Caenorhabditis elegans} connectome reconstructed by
\citet{cook2019whole}. The edge list and neuron annotations were obtained
through the OpenWorm Connectome Toolbox.\footnote{\url{https://github.com/openworm/ConnectomeToolbox}}
We retain directed chemical synapses and exclude electrical gap
junctions and self-loops. Synapse counts are aggregated over each ordered
neuron pair, and the primary adjacency matrix records whether at least
one chemical synapse is present. Preserving edge orientation is essential because the connection
probabilities from functional type \(a\) to type \(b\) and from type
\(b\) to type \(a\) can differ substantially. Symmetrization would
conflate these directional probabilities and obscure the
sender-receiver structure visible in the empirical block matrix below.

The biological annotations are used to form three broad functional
classes: \(83\) sensory neurons, \(81\) interneurons, and \(116\) motor
neurons. Pharyngeal neurons and neurons without one of these three
labels are excluded, leaving \(n=280\) neurons. The resulting binary
directed graph contains \(2846\) ordered edges and has density \(0.0364\).
The number of clusters is fixed at the metadata value \(K=3\), so the
analysis is an oracle-\(K\) comparison. Apart from specifying \(K\), the
functional labels are not used to construct neighborhoods, form
features, initialize \(K\)-means, or fit any of the four procedures.
They are used only as an external evaluation benchmark and are not
assumed to be the latent partition of a correctly specified directed
SBM.

Because the network is small, KMP uses the exact dissimilarity in
\eqref{eq:distance}, rather than the landmark approximation used in the
simulation section. The primary value \(C_h=0.5\) was fixed before
examining agreement with the biological labels; the post hoc bandwidth
sensitivity analysis is reported in the Supplementary Material. We
apply the empirical-quantile rule in \eqref{eq:neighborhood_i}, exclude
the focal vertex, and retain all ties. The nominal neighborhood size is \(20\), while the mean
realized size is \(56.1\) because many dissimilarities are tied.
All four methods use \(20\) \(K\)-means++ initializations. Spectral-LR
uses the unscaled coordinates
\([\widehat{\mathbf U}_3,\widehat{\mathbf V}_3]\), exactly as in the
simulation study.

\subsection{Directed block-profile structure}

\begin{figure}[t]
\centering
\begin{subfigure}[t]{0.49\textwidth}
\centering
\includegraphics[width=\textwidth]{\detokenize{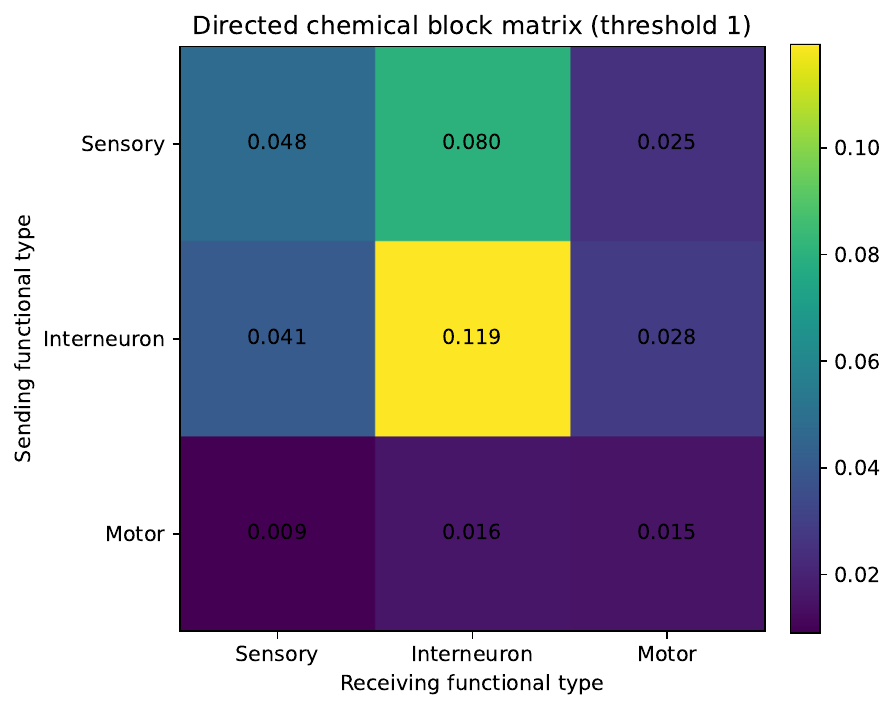}}
\caption{Estimated directed block matrix.}
\end{subfigure}
\hfill
\begin{subfigure}[t]{0.49\textwidth}
\centering
\includegraphics[width=\textwidth]{\detokenize{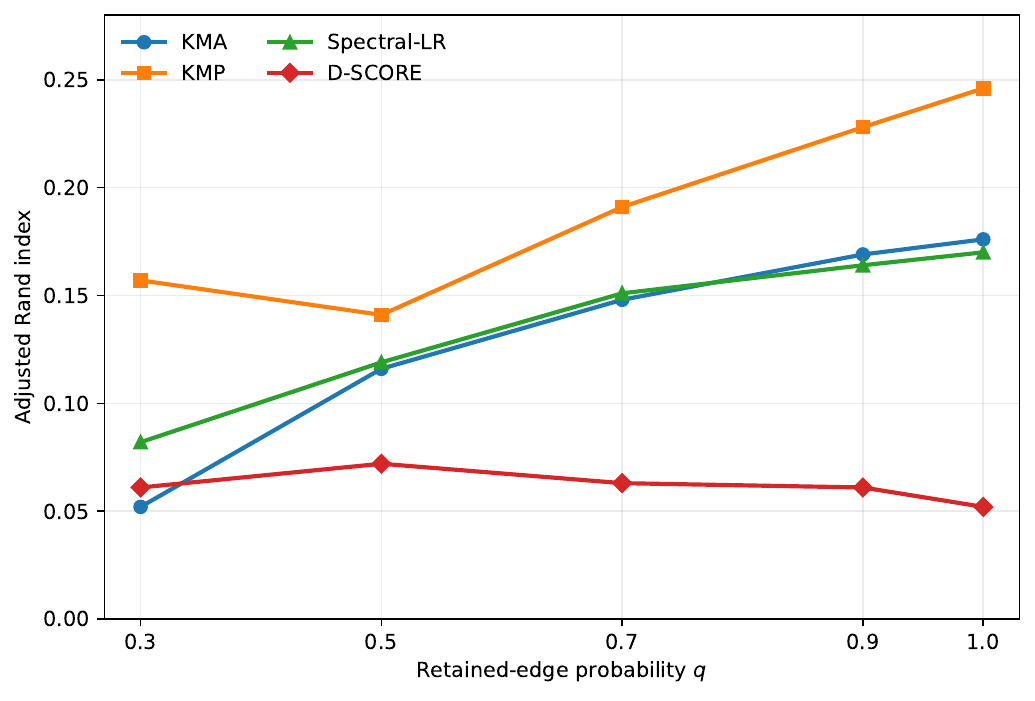}}
\caption{Mean ARI under edge thinning.}
\end{subfigure}
\caption{\textit{C. elegans} chemical-synapse analysis. In panel (b),
each observed edge is retained independently with probability \(q\),
and the lines connect the mean ARI values over \(50\) thinned networks.
At \(q=1\), the graph and algorithmic seed are fixed.}
\label{fig:celegans-main}
\end{figure}

Let \(z_i\in\{1,2,3\}\) denote the functional annotation and let
\(n_a=\sum_i\mathbbm{1}\{z_i=a\}\). We estimate the directed
between-type connection probabilities by
\[
\widehat B_{ab}
=
\frac{
\sum_{i:z_i=a}\sum_{j:z_j=b}A_{ij}
}{
n_an_b-\mathbbm{1}\{a=b\}n_a
}.
\]
With the ordering sensory, interneuron, and motor, the estimated block
matrix is
\[
\widehat{\mathbf B}
=
\begin{pmatrix}
0.0476&0.0799&0.0247\\
0.0409&0.1194&0.0280\\
0.0090&0.0158&0.0150
\end{pmatrix}.
\]
The normalized asymmetry
\(\|\widehat{\mathbf B}-\widehat{\mathbf B}^{\top}\|_F/
\|\widehat{\mathbf B}\|_F\) is \(0.380\), and the minimum Euclidean
distance between distinct outgoing block rows is \(0.0403\).
Figure~\ref{fig:celegans-main}a displays the resulting directional
structure. Sensory neurons send most frequently to interneurons,
interneurons have the largest within-type connection probability, and
motor neurons have a substantially sparser outgoing profile.

\subsection{Clustering agreement and edge-thinning stability}

Table~\ref{tab:celegans-primary} reports the primary comparison.
KMP has the largest ARI, AMI, matched accuracy, balanced accuracy, and
macro-\(F_1\). In particular, its ARI is \(0.246\), compared with
\(0.176\) for KMA, \(0.170\) for Spectral-LR, and \(0.052\) for
D-SCORE. The matched accuracy is \(0.596\), corresponding to \(113\)
misclassified neurons. Spectral-LR has a similar matched accuracy
(\(0.582\)) but a smaller ARI, reflecting a different distribution of
class-specific errors.

\begin{table}[t]
\centering
\caption{Agreement with sensory, interneuron, and motor metadata in the
binary directed chemical-synapse network. The largest value in each
performance column and the smallest error count are shown in bold.
Accuracy, balanced accuracy, macro-\(F_1\), and the error count are
computed after optimal label matching.}
\label{tab:celegans-primary}
\small
\resizebox{\textwidth}{!}{%
\begin{tabular}{@{}lrrrrrr@{}}
\toprule
Method & ARI & AMI & Accuracy & Balanced acc. & Macro-\(F_1\) & Errors\\
\midrule
KMP & \textbf{0.246} & \textbf{0.198} & \textbf{0.596}
    & \textbf{0.557} & \textbf{0.551} & \textbf{113}\\
KMA & 0.176 & 0.150 & 0.536 & 0.489 & 0.475 & 130\\
Spectral-LR & 0.170 & 0.167 & 0.582 & 0.528 & 0.512 & 117\\
D-SCORE & 0.052 & 0.059 & 0.450 & 0.416 & 0.412 & 154\\
\bottomrule
\end{tabular}%
}
\end{table}

We next assess sensitivity to missing connections. For \(q\in\{\)0.3, 0.5, 0.7, 0.9, 1\(\}\), each observed chemical edge is retained
independently with probability \(q\), and every method is refitted with
the same algorithmic seed. KMP has the largest mean ARI at every
retention level; its mean ARI is \(0.157,0.141,0.191,0.228,\) and
\(0.246\), respectively. The corresponding best competing means are
\(0.082,0.119,0.151,0.169,\) and \(0.176\). Thus, the advantage is not
confined to the unthinned graph, although the paired advantage is
smallest at \(q=0.5\). This experiment is a finite-sample sensitivity
analysis conditional on the observed connectome; it is not a consequence
of Theorem~\ref{thm:exact_recovery}, which is proved under the directed
SBM rather than under random deletion from a fixed observed graph.
Full threshold, bandwidth, paired-thinning, confusion-matrix, and degree
diagnostics are reported in the Supplementary Material.

The application also illustrates the limitations of interpreting
biological metadata as an exact-recovery target. Sixty-six neurons have
zero observed out-degree, including \(64\) motor neurons and \(2\)
sensory neurons. Their raw outgoing rows contain no information that
can distinguish the two annotated types, and the motor group has much
greater within-class degree heterogeneity than the other classes.
Accordingly, no method exactly matches all metadata labels. The
connectome results should therefore be interpreted as evidence of
stable partial agreement with broad functional annotations, rather than
as empirical verification of the asymptotic exact-recovery theorem.

\section{Conclusions}
\label{sec:conclusions}

We extend neighborhood smoothing from undirected probability-matrix
estimation to community recovery in directed stochastic block models.
KMP estimates asymmetric outgoing block profiles and clusters their
rows. We prove a uniform two-to-infinity error bound and, equivalently,
normalized row-wise consistency. We then derive exact
recovery when the row-wise estimation error is dominated by the minimum
separation between distinct population profiles. The theory allows a vanishing sparsity factor and a number of
communities that may increase with \(n\).

The fitted block profiles provide both community labels and
interpretable connectivity estimates.

The simulations delineate the regimes in which smoothing is most
effective and the finite-sample limitations created by sparsity,
neighborhood contamination, informative degree variation, and growing
numbers of communities.

The \textit{C. elegans} application provides a complementary
real-network illustration: KMP has the strongest agreement with the
three broad functional labels and favorable mean behavior under edge
thinning, but it does not dominate every biological stability
diagnostic. More generally, the large-\(n\) simulation procedure uses up to
\(128\) landmarks and repeated Lloyd-algorithm fits, whereas the theory
assumes the exact dissimilarity and a global \(K\)-means solution. Establishing explicit landmark-sampling guarantees and closing
this computational gap remain important directions for future work.

Several extensions remain open:
\begin{enumerate}[label=(\roman*)]
\item deriving sharper recovery conditions and transition results for
genuinely growing \(K_n\);
\item developing adaptive bandwidth rules and profile-based procedures
for estimating the number of communities;
\item incorporating degree regularization or degree correction into
probability-profile smoothing;
\item establishing guarantees for landmark, sketching, approximate
nearest-neighbor, and other scalable implementations;
\item studying robustness to model misspecification, including mixed or
overlapping memberships and sender-receiver co-community structure;
and
\item extending the framework to dynamic, weighted, multilayer, and
covariate-assisted directed networks.
\end{enumerate}

\section*{Funding Information}
The authors received no specific financial support for the research, authorship, and/or publication of this article.

\section*{Conflict of Interest}
The authors declare that they have no known competing financial interests or personal relationships that could have appeared to influence the work reported in this paper.
\bibliographystyle{apalike}
\bibliography{Reference}

\begin{center}
\large\bfseries Table of Contents
\end{center}

\begin{enumerate}
\item \textbf{Consolidated notation}
  \item \textbf{Proofs}
    \begin{enumerate}[label*=\arabic*.]
      \item Proof of Lemma~\ref{lem:deterministic_clustering}
      \item Proof of Theorem~\ref{thm:rowwise_bound} (Estimation Error Bound)
        \begin{enumerate}[label*=\arabic*.]
          \item Auxiliary Lemmas for Theorem~\ref{thm:rowwise_bound}
          \item Proof of Theorem~\ref{thm:rowwise_bound}
        \end{enumerate}
      \item Proof of Corollary~\ref{cor:two_to_infinity}
      \item Proof of Corollary~\ref{cor:frobenius_bound}
      \item Proof of Theorem~\ref{thm:exact_recovery}
      (Exact Recovery via Separation)
      \item Proof of Corollary~\ref{cor:estimated-K}
    \end{enumerate}
  \item \textbf{Additional methodological and computational details}
  \item \textbf{Additional simulation results}
  \item \textbf{Additional \textit{C. elegans} connectome results}
\end{enumerate}

\section{Consolidated notation}
\label{sec:supp-notation}

For convenience, Table~\ref{tab:notation} provides a self-contained
reference for the model, estimator, and proof-specific notation used in
the Supplementary Material. It includes the principal symbols from the
main text together with additional quantities required in the proofs.
\begin{table}[H]
\centering
\small
\renewcommand{\arraystretch}{1.15}
\begin{tabularx}{\textwidth}
{@{}>{\raggedright\arraybackslash}p{0.27\textwidth}
>{\raggedright\arraybackslash}X@{}}
\toprule
Symbol & Meaning \\ \midrule
$n$ & Number of nodes. \\[2pt]

$K_n$ & Number of communities (may grow with $n$). \\[2pt]

$\boldsymbol{\rho}=(\rho_1,\ldots,\rho_{K_n})$ &
Community-proportion vector, which may change with \(n\).\\[2pt]

$n_{\min}$ & Size of the smallest realized community.\\[2pt]

$\rho_{\min}=\min_k\rho_k>0$ &
Minimum population community proportion, positive for every \(n\).\\[2pt]

$\pi : \{1,\ldots,n\}\!\to\!\{1,\ldots,K_n\}$ &
Community-label map; $\pi(i)$ is the label of node $i$. \\[2pt]

$C_k=\{\,i:\pi(i)=k\,\}$ & Node set of community $k$;
$n_k=|C_k|$, $n_{\min}=\min_{k}n_k$. \\[2pt]

$\mathbf B\in[0,1]^{K_n\times K_n}$ &
Base block-probability matrix; row \(\mathbf B_{a\cdot}\) is the
unscaled outgoing connection-probability profile of community \(a\).
\\[2pt]

$\gamma_n$ & Sparsity multiplier with $0<\gamma_n\le 1$. \\[2pt]

$\mathbf P^{0}$ &
Node-level expansion of \(\mathbf B\):
\(P^{0}_{ij}=B_{\pi(i)\pi(j)}\) for all \(i,j\in[n]\). Thus,
\(\mathbf P^0_{i\cdot}\) is the unscaled outgoing block profile of
vertex \(i\).\\[2pt]

$\mathbf P=\gamma_n\mathbf P^{0}$ &
Sparsity-scaled block-profile target, including the block-defined
diagonal.\\[2pt]

$\mathbf P^{\mathrm{off}}$ &
Zero-diagonal conditional mean adjacency matrix:
\(
\mathbf P^{\mathrm{off}}
=
\mathbb E(\mathbf A\mid\mathbf Z)
=
\mathbf P-\operatorname{diag}(\mathbf P).
\)\\[2pt]

$d_\mathbf{B}(a,b)$ &
$\bigl\lVert \mathbf B_{a\cdot}-\mathbf B_{b\cdot}\bigr\rVert_{2}$,
row separation in $\mathbf B$. \\[2pt]

$d_\mathbf{B}^{*}$ &
\(\displaystyle\min_{a\neq b}d_\mathbf{B}(a,b)\), minimum row
separation, which may change with \(n\).\\[2pt]

\(d_{\mathbf P}^*\) & minimum Euclidean separation between distinct
population rows of \(\mathbf P\)\\[2pt]

$d_\mathbf{P}(i,j)$ &
\(\bigl\lVert
\mathbf P_{i\cdot}-\mathbf P_{j\cdot}
\bigr\rVert_2\), Euclidean distance between sparsity-scaled block
profiles.\\[2pt]
$h$ &
Neighborhood quantile level.\\[2pt]

$q_i(h)$ &
Lower empirical \(h\)-quantile of
\(\{d(i,j):j\neq i\}\).\\[2pt]

$\widetilde{\mathbf P}$ &
Neighborhood-smoothed estimator:
\(
\widetilde P_{ij}
=
|N_i|^{-1}\sum_{u\in N_i}A_{uj}.
\)\\[2pt]

$\epsilon_\pi,\epsilon_{AP},\epsilon_m$ &
Deviation levels used in the label-count, Gram-matrix, and centered
inner-product concentration bounds, respectively.\\[2pt]

$\mathcal C_{K_n}$ &
Set of matrices in \(\mathbb R^{n\times n}\) having at most
\(K_n\) distinct rows.\\[2pt]
\(C_h,C_m,C_{AP},C_1,\dots\) &
Fixed positive constants independent of \(n\).\\[2pt]

$N_i$ &
Empirical neighborhood of node \(i\):
\(
N_i=\{j\in[n]\setminus\{i\}:d(i,j)\leq q_i(h)\}.
\)\\[2pt]

\(d(i,j)\) &
Normalized dissimilarity
\(
d(i,j)
=
\max_{k\neq i,j}
\left|
n^{-1}
\langle
\mathbf A_{i\cdot}-\mathbf A_{j\cdot},
\mathbf A_{k\cdot}
\rangle
\right|.
\)\\[2pt]
$[m]$ & \(\{1,\ldots,m\}\) for a positive integer \(m\).\\[2pt]
$\lVert\mathbf M\rVert_{F}$ &
Frobenius norm $\bigl(\sum_{i,j}M_{ij}^{2}\bigr)^{1/2}$. \\[2pt]

$\lVert\mathbf M\rVert_{2\to\infty}$ &
Row-wise maximum Euclidean norm
$\displaystyle\max_i\lVert\mathbf M_{i\cdot}\rVert_{2}$. \\
\(\mathcal O_P(a_n)\) &
\(X_n=\mathcal O_P(a_n)\) if, for every \(\varepsilon>0\), there exist
\(M<\infty\) and \(n_0\) such that
\(\sup_{n\geq n_0}
\mathbb P(|X_n|>Ma_n)<\varepsilon\).\\[2pt]
$o(\,\cdot\,)$ &
For sequences $\{a_n\},\{b_n\}$: $a_n=o(b_n)$ means  
$\displaystyle\frac{a_n}{b_n}\to 0$ as $n\to\infty$. \\[2pt]

$\omega(\,\cdot\,)$ &
$a_n=\omega(b_n)$ means $b_n=o(a_n)$, i.e.  
$\displaystyle\frac{a_n}{b_n}\to\infty$ as $n\to\infty$. \\ \bottomrule
\end{tabularx}
\caption{Key model, estimator, and proof-specific notation used in the
main text and Supplementary Material. Bold Roman symbols denote
matrices, while \(\boldsymbol\rho\) denotes the community-proportion
vector.}
\label{tab:notation}
\end{table}

The remaining sections provide proof details, computational remarks, and additional simulation tables.

\section{Proofs}
\label{sec:supp-proofs}

\subsection{Proof of Lemma~\ref{lem:deterministic_clustering}}

\begin{proof}
Let the \(K_n\) distinct rows of \(\mathbf P\) be
\(\mu_1,\ldots,\mu_{K_n}\), and put
\[
\delta=d_{\mathbf P}^*,
\qquad
\beta
=
\|\widehat{\mathbf P}-\mathbf P\|_{2\to\infty}.
\]
The assumed inequality implies
\(
2\beta\sqrt{\frac{n}{n_{\min}}}
<
\frac{\delta}{4}.
\)
Choose a radius \(r\) satisfying
\[
2\beta\sqrt{\frac{n}{n_{\min}}}
<
r
<
\frac{\delta}{4}.
\]
The balls \(B(\mu_a,r)\), \(a\in[K_n]\), are pairwise disjoint.
Because \(\mathbf P\in\mathcal C_{K_n}\), the global optimality of
\(\widehat{\mathbf C}\) yields
\[
\|\widehat{\mathbf C}-\widehat{\mathbf P}\|_F
\leq
\|\mathbf P-\widehat{\mathbf P}\|_F
\leq
\beta\sqrt n.
\]
Suppose that \(B(\mu_a,r)\) contains no fitted centroid. Every row of
\(\widehat{\mathbf C}\) assigned to a vertex in community \(a\) is then
at distance at least \(r\) from \(\mu_a\), and hence
\[
\|\widehat{\mathbf C}-\mathbf P\|_F
\geq
r\sqrt{n_a}
\geq
r\sqrt{n_{\min}}
>
2\beta\sqrt n.
\]
The triangle inequality would give
\[
\|\widehat{\mathbf C}-\widehat{\mathbf P}\|_F
\geq
\|\widehat{\mathbf C}-\mathbf P\|_F
-
\|\widehat{\mathbf P}-\mathbf P\|_F
>
\beta\sqrt n,
\]
contradicting global optimality. Thus every ball contains a fitted centroid. Since there are \(K_n\)
pairwise disjoint balls and \(\widehat{\mathbf C}\) has at most \(K_n\)
distinct rows, it follows that \(\widehat{\mathbf C}\) has exactly
\(K_n\) distinct fitted centroids and that each ball contains exactly
one of them. Denote the centroid in \(B(\mu_a,r)\) by
\(\widehat\mu_a\).

For a vertex \(i\) in community \(a\),
\[
\|\widehat{\mathbf P}_{i\cdot}-\widehat\mu_a\|_2
\leq
\beta+r
<
\frac{3\delta}{8}.
\]
For \(b\neq a\),
\[
\|\widehat{\mathbf P}_{i\cdot}-\widehat\mu_b\|_2
\geq
\|\mu_a-\mu_b\|_2-\beta-r
>
\frac{5\delta}{8}.
\]
Therefore every estimated row is strictly closer to the fitted centroid
associated with its true population row than to any other fitted
centroid. At a global minimizer, every estimated row must be assigned to
a nearest fitted centroid: otherwise, reassigning that row to a closer
centroid would strictly decrease the objective without increasing the
number of distinct rows. Hence all vertices in community \(a\) are
assigned to \(\widehat\mu_a\). Because every true community is nonempty,
all \(K_n\) fitted centroids are used. Hence every global
minimizer induces the true partition up to a permutation of the labels,
and
\[
A(\pi,\widehat\pi)=1.
\]
\end{proof}
\subsection{Proof of Theorem~\ref{thm:rowwise_bound} (Estimation Error Bound)}
\subsubsection{Auxiliary Lemmas for Theorem~\ref{thm:rowwise_bound}}

The label-count lemmas below concern the randomness of
\(\mathbf Z\). For concentration inequalities involving adjacency
entries, we first condition on \(\mathbf Z\), under which the directed
edges are independent. The resulting bounds depend only on
\(n\) and the displayed deviation parameters and are uniform over all
realizations of \(\mathbf Z\); integrating over \(\mathbf Z\) therefore
gives the unconditional inequalities stated in the lemmas.

The concentration events are also uniform over all relevant vertex
pairs. Consequently, they remain valid after the data-dependent
neighborhoods \(N_i\) have been selected from \(\mathbf A\); no
conditioning on the realized neighborhoods or sample splitting is
required.

\begin{lemma}[Proportion of same-community candidates]
\label{lemma:nodeproportion}
Suppose \(n\geq4\) and \(\epsilon_\pi\geq4/n\). Then
\begin{align}
\label{eq:proportion-bound}
\mathbb P\left\{
\max_{i\in[n]}
\left|
\frac{
|\{u\neq i:\pi(u)=\pi(i)\}|
}{n-1}
-
\rho_{\pi(i)}
\right|
<
\epsilon_\pi
\right\}
\geq\\
1-
2K_n\exp\left(-\frac{n\epsilon_\pi^2}{2}\right)
=:
\mathbb P_\pi(\epsilon_\pi).\notag
\end{align}
\end{lemma}

\begin{proof}
For \(a\in[K_n]\), let
\[
n_a
=
\sum_{i=1}^n\mathbbm 1\{Z_i=a\}.
\]
Hoeffding's inequality gives
\[
\mathbb P\left\{
\left|\frac{n_a}{n}-\rho_a\right|
\geq
\frac{\epsilon_\pi}{2}
\right\}
\leq
2\exp\left(-\frac{n\epsilon_\pi^2}{2}\right).
\]
On the complementary event,
\[
\begin{aligned}
\left|
\frac{n_a-1}{n-1}-\rho_a
\right|
&\leq
\frac{
n|n_a/n-\rho_a|+1
}{n-1}\\
&<
\frac{n\epsilon_\pi/2+1}{n-1}
\leq
\epsilon_\pi,
\end{aligned}
\]
For \(n\geq4\) and \(\epsilon_\pi\geq4/n\),
\[
\epsilon_\pi\frac{n-2}{2}
\geq
\frac{2(n-2)}{n}
\geq1,
\]
which is equivalent to
\[
\frac{n\epsilon_\pi/2+1}{n-1}
\leq
\epsilon_\pi.
\]
For every vertex \(i\) in community \(a\), the number of other vertices
in its community is exactly \(n_a-1\). A union bound over
\(a\in[K_n]\) proves \eqref{eq:proportion-bound}.
\end{proof}

Assumption~\ref{aasump:condition_epsilon_pi} implies
\(\epsilon_\pi\geq4/n\) for all sufficiently large \(n\), and it makes
the failure probability in \eqref{eq:proportion-bound} converge to zero.

\begin{lemma}[Neighborhood size]
\label{lem:neighborhood-size}
Let \(N_i\) be the empirical-quantile neighborhood defined in
\eqref{eq:neighborhood_i}. Then, for every \(i\in[n]\),
\[
|N_i|
\geq
\lceil h(n-1)\rceil,
\qquad\text{and hence}\qquad
\frac{|N_i|}{n-1}\geq h.
\]
\end{lemma}

\begin{proof}
For fixed \(i\), the collection
\(\{d(i,j):j\neq i\}\) contains \(n-1\) values. By the definition of the
empirical \(h\)-quantile, at least
\(\lceil h(n-1)\rceil\) of these values do not exceed \(q_i(h)\).
These indices are precisely the elements retained in \(N_i\). Therefore,
\[
|N_i|\geq\lceil h(n-1)\rceil\geq h(n-1),
\]
which proves the result.
\end{proof}

\begin{lemma}[Off-diagonal concentration of the outgoing Gram matrix]
\label{lem:AAT-PPT}
Let \(\mathbf A\) be generated from the directed SBM and let
\(\mathbf P\) be the block-profile target. For every \(n\geq3\) and
\(\epsilon_{AP}>2/n\),
\begin{align}
\label{eq:prob-bound-AAT}
\mathbb P\left\{
\max_{\substack{i,j\in[n]\\i\neq j}}
\left|
\left(\frac1n\mathbf A\mathbf A^\top\right)_{ij}
-
\left(\frac1n\mathbf P\mathbf P^\top\right)_{ij}
\right|
\geq
\epsilon_{AP}
\right\}
\leq\\
2n^2
\exp\left\{
-\frac{n(\epsilon_{AP}-2/n)^2}
{4(1+\epsilon_{AP})}
\right\}.
\end{align}
Denote one minus the right-hand side by
\(\mathbb P_{AP}(\epsilon_{AP})\).
\end{lemma}

The maximum is restricted to off-diagonal entries. Indeed,
\((\mathbf A\mathbf A^\top)_{ii}\) is an observed out-degree and is not
centered at \((\mathbf P\mathbf P^\top)_{ii}\).

\begin{proof}
Fix distinct \(i,j\). Conditional on the labels, for
\(k\notin\{i,j\}\) the variables
\[
X_k
=
A_{ik}A_{jk}-P_{ik}P_{jk}
\]
are independent, mean zero, and bounded in absolute value by one. The
two omitted coordinates \(k=i,j\) contribute at most \(2/n\) in
absolute value. Therefore,
\[
\begin{aligned}
&\mathbb P\left\{
\left|
\left(\frac1n\mathbf A\mathbf A^\top\right)_{ij}
-
\left(\frac1n\mathbf P\mathbf P^\top\right)_{ij}
\right|
\geq\epsilon_{AP}
\,\middle|\,\mathbf Z
\right\}\\
&\qquad\leq
\mathbb P\left\{
\frac1{n-2}
\left|
\sum_{k\notin\{i,j\}}X_k
\right|
\geq
\frac{n}{n-2}
\left(\epsilon_{AP}-\frac2n\right)
\,\middle|\,\mathbf Z
\right\}.
\end{aligned}
\]
Moreover,
\[
\sum_{k\notin\{i,j\}}
\operatorname{Var}(X_k\mid\mathbf Z)
\leq n-2,
\]
because \(|X_k|\leq1\). Put
\[
t
:=
n\left(\epsilon_{AP}-\frac2n\right)
=
n\epsilon_{AP}-2>0.
\]
Bernstein's inequality yields
\[
\mathbb P\left\{
\left|
\sum_{k\notin\{i,j\}}X_k
\right|
\geq t
\,\middle|\,\mathbf Z
\right\}
\leq
2\exp\left\{
-\frac{t^2}
{2\{(n-2)+t/3\}}
\right\}.
\]
Since
\[
2\{(n-2)+t/3\}
\leq
4n(1+\epsilon_{AP}),
\]
we obtain
\[
\frac{t^2}{2\{(n-2)+t/3\}}
\geq
\frac{n(\epsilon_{AP}-2/n)^2}
{4(1+\epsilon_{AP})}.
\]
Consequently,
\[
\mathbb P\left\{
\left|
\left(\frac1n\mathbf A\mathbf A^\top\right)_{ij}
-
\left(\frac1n\mathbf P\mathbf P^\top\right)_{ij}
\right|
\geq\epsilon_{AP}
\,\middle|\,\mathbf Z
\right\}
\leq
2\exp\left\{
-\frac{n(\epsilon_{AP}-2/n)^2}
{4(1+\epsilon_{AP})}
\right\}.
\]
The bound does not depend on the realized labels. Taking a union bound
over the at most \(n(n-1)\) ordered off-diagonal pairs and then
integrating over \(\mathbf Z\) proves \eqref{eq:prob-bound-AAT}.
\end{proof}

\begin{lemma}[Dissimilarity of selected neighbors]
\label{lem:dissimilarity-neighborhood}
Suppose \(h<\rho_{\min}-\epsilon_\pi\). On the intersection of the
events in Lemmas~\ref{lemma:nodeproportion} and~\ref{lem:AAT-PPT},
\[
d(i,i')\leq2\epsilon_{AP}
\qquad
\text{for every }i\in[n]\text{ and every }i'\in N_i.
\]
Consequently, this conclusion holds with probability at least
\[
1-
\{1-\mathbb P_\pi(\epsilon_\pi)\}
-
\{1-\mathbb P_{AP}(\epsilon_{AP})\}.
\]
\end{lemma}

\begin{proof}
Let \(i\) and \(u\neq i\) belong to the same community. For every
\(k\notin\{i,u\}\), the entries \((i,k)\) and \((u,k)\) of the Gram
matrices are off diagonal, and
\[
\left(\mathbf P\mathbf P^\top\right)_{ik}
=
\left(\mathbf P\mathbf P^\top\right)_{uk}
\]
because \(\mathbf P_{i\cdot}=\mathbf P_{u\cdot}\). Hence, on the event
in Lemma~\ref{lem:AAT-PPT},
\[
d(i,u)
=
\max_{k\notin\{i,u\}}
\left|
\left(\frac1n\mathbf A\mathbf A^\top\right)_{ik}
-
\left(\frac1n\mathbf A\mathbf A^\top\right)_{uk}
\right|
\leq
2\epsilon_{AP}.
\]
On the event in Lemma~\ref{lemma:nodeproportion},
\[
\frac{
|\{u\neq i:\pi(u)=\pi(i)\}|
}{n-1}
>
\rho_{\pi(i)}-\epsilon_\pi
\geq
\rho_{\min}-\epsilon_\pi
>
h.
\]
Every such same-community candidate \(u\) satisfies
\(d(i,u)\leq2\epsilon_{AP}\). Hence more than an \(h\)-fraction of the
\(n-1\) candidate dissimilarities are at most \(2\epsilon_{AP}\).
By the definition of the lower empirical quantile,
\[
q_i(h)\leq2\epsilon_{AP}.
\]
Finally, \(i'\in N_i\) implies
\[
d(i,i')
\leq
q_i(h)
\leq
2\epsilon_{AP},
\]
which proves the result simultaneously for every \(i\) and every
\(i'\in N_i\).
\end{proof}

\begin{lemma}[Population-row proximity for selected neighbors]
\label{lemma:Pi-Pi'}
Under the conditions of Lemma~\ref{lem:dissimilarity-neighborhood}, for
all sufficiently large \(n\), the same event implies
\[
\frac1n
\|\mathbf P_{i\cdot}-\mathbf P_{i'\cdot}\|_2^2
\leq
8\epsilon_{AP}
\qquad
\text{for every }i\in[n]\text{ and }i'\in N_i.
\]
\end{lemma}

\begin{proof}
Fix \(i\in[n]\) and \(i'\in N_i\). We first verify that the required
witness vertices exist. On the event in
Lemma~\ref{lemma:nodeproportion}, for any community \(a\) containing at
least one of the vertices under consideration,
\[
\frac{n_a-1}{n-1}
>
\rho_a-\epsilon_\pi
\geq
\rho_{\min}-\epsilon_\pi.
\]
Under Assumption~\ref{aasump:condition_epsilon_pi},
\[
(n-1)(\rho_{\min}-\epsilon_\pi)
\geq
(n-1)(C_\rho-C_\pi)
\sqrt{\frac{\log n}{n}}
\longrightarrow\infty.
\]
Therefore, for all sufficiently large \(n\), the communities containing
\(i\) and \(i'\) each contain at least three vertices. We may choose
\(u,v\notin\{i,i'\}\) such that
\[
\pi(u)=\pi(i),
\qquad
\pi(v)=\pi(i').
\]
If \(i\) and \(i'\) belong to the same community, the same third vertex
may be used for both \(u\) and \(v\).

Define
\[
G^A
:=
\frac1n\mathbf A\mathbf A^\top,
\qquad
G^P
:=
\frac1n\mathbf P\mathbf P^\top.
\]
Since
\[
\mathbf P_{u\cdot}=\mathbf P_{i\cdot},
\qquad
\mathbf P_{v\cdot}=\mathbf P_{i'\cdot},
\]
we have
\[
G^P_{iu}=G^P_{ii},
\qquad
G^P_{i'u}=G^P_{i'i},
\qquad
G^P_{i'v}=G^P_{i'i'},
\qquad
G^P_{iv}=G^P_{ii'}.
\]
Consequently,
\[
\begin{aligned}
\frac1n
\|\mathbf P_{i\cdot}-\mathbf P_{i'\cdot}\|_2^2
&=
G^P_{ii}-G^P_{i'i}
+
G^P_{i'i'}-G^P_{ii'}\\
&=
\left(G^P_{iu}-G^P_{i'u}\right)
+
\left(G^P_{i'v}-G^P_{iv}\right).
\end{aligned}
\]
Because \(u,v\notin\{i,i'\}\), all four Gram-matrix entries on the
last line are off diagonal. On the event in
Lemma~\ref{lem:AAT-PPT},
\[
\begin{aligned}
G^P_{iu}-G^P_{i'u}
&\leq
|G^P_{iu}-G^A_{iu}|
+
|G^A_{iu}-G^A_{i'u}|
+
|G^A_{i'u}-G^P_{i'u}|\\
&\leq
d(i,i')+2\epsilon_{AP},
\end{aligned}
\]
because \(u\) is an admissible witness in the definition of
\(d(i,i')\). Similarly,
\[
G^P_{i'v}-G^P_{iv}
\leq
d(i,i')+2\epsilon_{AP}.
\]
It follows that
\[
\frac1n
\|\mathbf P_{i\cdot}-\mathbf P_{i'\cdot}\|_2^2
\leq
2d(i,i')+4\epsilon_{AP}.
\]
Lemma~\ref{lem:dissimilarity-neighborhood} gives
\[
d(i,i')\leq2\epsilon_{AP},
\]
and hence
\[
\frac1n
\|\mathbf P_{i\cdot}-\mathbf P_{i'\cdot}\|_2^2
\leq
8\epsilon_{AP}.
\]
The argument is uniform in \(i\) and \(i'\in N_i\), proving the lemma.
\end{proof}

\begin{lemma}[Concentration of pairwise centered inner products]
\label{lemma_M}
For distinct \(u,v\), define
\begin{equation}
\label{eq:M-pair}
M(u,v)
=
\frac1{n-2}
\left|
\sum_{j\notin\{u,v\}}
(A_{uj}-P_{uj})(A_{vj}-P_{vj})
\right|.
\end{equation}
Then, for every \(\epsilon_m>0\) and \(n\geq4\),
\begin{equation}
\label{P_cross}
\mathbb P\left\{
\max_{u\neq v}M(u,v)\geq\epsilon_m
\right\}
\leq
n(n-1)
\exp\left\{
-\frac{n\epsilon_m^2}
{4(1+\epsilon_m)}
\right\}.
\end{equation}
Denote one minus the right-hand side by
\(\mathbb P_{\mathrm{cross}}(\epsilon_m)\).
\end{lemma}

\begin{proof}
Conditional on \(\mathbf Z\), fix distinct \(u,v\) and define
\[
Y_j
:=
(A_{uj}-P_{uj})(A_{vj}-P_{vj}),
\qquad
j\notin\{u,v\}.
\]
The variables \(\{Y_j:j\notin\{u,v\}\}\) are independent because they
involve disjoint directed edges. Moreover,
\[
\mathbb E(Y_j\mid\mathbf Z)
=
\mathbb E(A_{uj}-P_{uj}\mid\mathbf Z)
\mathbb E(A_{vj}-P_{vj}\mid\mathbf Z)
=
0,
\]
and
\[
|Y_j|\leq1,
\qquad
\sum_{j\notin\{u,v\}}
\operatorname{Var}(Y_j\mid\mathbf Z)
\leq n-2.
\]
Applying Bernstein's inequality with threshold
\((n-2)\epsilon_m\) gives
\[
\mathbb P\left\{
M(u,v)\geq\epsilon_m
\,\middle|\,\mathbf Z
\right\}
\leq
2\exp\left\{
-\frac{(n-2)\epsilon_m^2}
{2(1+\epsilon_m/3)}
\right\}.
\]
The quantity \(M(u,v)\) is symmetric in \(u,v\). A union bound over the
\(\binom n2\) unordered pairs, followed by the conservative
simplification
\[
\frac{(n-2)\epsilon_m^2}
{2(1+\epsilon_m/3)}
\geq
\frac{n\epsilon_m^2}
{4(1+\epsilon_m)}
\qquad(n\geq4),
\]
proves \eqref{P_cross}. The bound is uniform in the realized labels, so
the same inequality holds unconditionally.
\end{proof}
\subsubsection{Proof of Theorem~\ref{thm:rowwise_bound}}

\begin{proof}
Let \(\widetilde{\mathbf P}\) be defined by
\eqref{P_tilde_Estimate}, and let \(\mathbf P\) be the block-profile
target. We prove the asserted uniform high-probability bound on
\[
\frac1n
\bigl\|
\widetilde{\mathbf P}_{i\cdot}
-
\mathbf P_{i\cdot}
\bigr\|_2^2,
\qquad i\in[n].
\]
Corollary~\ref{cor:two_to_infinity} then specializes the deviation
parameters and converts this bound into a two-to-infinity rate.

Using the definition of \(\widetilde P_{ij}\), write
\begin{align*}
\widetilde P_{ij}-P_{ij}
&=
\frac1{|N_i|}
\sum_{i'\in N_i}
\left(A_{i'j}-P_{ij}\right)\\
&=
\frac1{|N_i|}
\sum_{i'\in N_i}
\left\{
(A_{i'j}-P_{i'j})
+
(P_{i'j}-P_{ij})
\right\}.
\end{align*}
Consequently,
\[
\frac1n
\|\widetilde{\mathbf P}_{i\cdot}-\mathbf P_{i\cdot}\|_2^2
=
\frac1n
\sum_{j=1}^n
\left[
\frac1{|N_i|}
\sum_{i'\in N_i}
\left\{
(A_{i'j}-P_{i'j})
+
(P_{i'j}-P_{ij})
\right\}
\right]^2.
\]

For fixed \(i,j\), define
\[
R_1(i,j)
:=
\left\{
\frac1{|N_i|}
\sum_{i'\in N_i}
(A_{i'j}-P_{i'j})
\right\}^2
\]
and
\[
R_2(i,j)
:=
\left\{
\frac1{|N_i|}
\sum_{i'\in N_i}
(P_{i'j}-P_{ij})
\right\}^2.
\]
The inequality \((a+b)^2\leq2a^2+2b^2\) gives
\[
\left[
\frac1{|N_i|}
\sum_{i'\in N_i}
\left\{
(A_{i'j}-P_{i'j})
+
(P_{i'j}-P_{ij})
\right\}
\right]^2
\leq
2R_1(i,j)+2R_2(i,j).
\]
Consequently,
\begin{equation}
\label{bound 2Js}
\frac1n
\bigl\|
\widetilde{\mathbf P}_{i\cdot}
-
\mathbf P_{i\cdot}
\bigr\|_2^2
\leq
\frac2n\sum_{j=1}^n R_1(i,j)
+
\frac2n\sum_{j=1}^n R_2(i,j).
\end{equation}
It remains to bound the coordinate averages
\[
\frac1n\sum_{j=1}^n R_1(i,j)
\qquad\text{and}\qquad
\frac1n\sum_{j=1}^n R_2(i,j)
\]
uniformly over \(i\). We first consider the empirical-noise component involving \(R_1\).
Recall that \(A_{rr}=0\), whereas the block-profile target may have
\(P_{rr}>0\). Thus, \(A_{rj}-P_{rj}\) is conditionally centered only
when \(j\neq r\). The proof does not use centering at \(j=r\): the
single-row squared terms are controlled deterministically by
\(|A_{rj}-P_{rj}|\leq1\), and in each cross product involving two
distinct rows \(i'\) and \(i''\), the exceptional coordinates
\(j=i'\) and \(j=i''\) are removed and bounded by the explicit
\(2/n\) remainder below.

\begin{align*}
\frac{1}{n} \sum_{j=1}^n &R_1(i,j) 
= \frac{1}{n |N_i|^2} \sum_{j=1}^n 
    \left( \sum_{i' \in N_i} (A_{i'j} - P_{i'j}) \right)^2 \\
&= \frac{1}{n |N_i|^2} \sum_{j=1}^n \left\{ 
    \sum_{i' \in N_i} (A_{i'j} - P_{i'j})^2 
  + \sum_{\substack{i', i'' \in N_i \\ i' \neq i''}} 
    (A_{i'j} - P_{i'j})(A_{i''j} - P_{i''j}) 
\right\}.
\end{align*}

Since \(|A_{i'j}-P_{i'j}|\leq1\),
\[
\frac1n
\sum_{j=1}^n
(A_{i'j}-P_{i'j})^2
=
\frac1n
\|\mathbf A_{i'\cdot}-\mathbf P_{i'\cdot}\|_2^2
\leq1.
\]
Therefore, the diagonal part of the expansion satisfies
\begin{equation}
\label{first term}
\frac{1}{n|N_i|^2}
\sum_{i'\in N_i}
\sum_{j=1}^n
(A_{i'j}-P_{i'j})^2
\leq
\frac1{|N_i|}.
\end{equation}

For the cross term, we have
\begin{align*}
&\frac{1}{n |N_i|^2} 
  \sum_{j=1}^n \sum_{\substack{i', i'' \in N_i \\ i' \neq i''}} 
  (A_{i'j} - P_{i'j})(A_{i''j} - P_{i''j}) \\
&\quad \leq \frac{1}{|N_i|^2} 
  \sum_{\substack{i', i'' \in N_i \\ i' \neq i''}} 
  \left| \frac{1}{n} \sum_{j=1}^n 
  (A_{i'j} - P_{i'j})(A_{i''j} - P_{i''j}) \right| \\
&\quad \leq \frac{1}{|N_i|^2} 
  \sum_{\substack{i', i'' \in N_i \\ i' \neq i''}} 
  \left\{ \frac{n - 2}{n} 
  \cdot \frac{1}{n - 2} \left| \sum_{j \neq i', i''} 
  (A_{i'j} - P_{i'j})(A_{i''j} - P_{i''j}) \right| 
  + \frac{2}{n} \right\} \\
&\quad = \frac{1}{|N_i|^2} 
  \sum_{\substack{i', i'' \in N_i \\ i' \neq i''}} 
  \left( \frac{n - 2}{n} M(i',i'') + \frac{2}{n} \right),
\end{align*}
where \(M(i',i'')\) is defined in \eqref{eq:M-pair}. The two
coordinates \(j=i'\) and \(j=i''\) are excluded from \(M(i',i'')\);
because every factor is bounded in absolute value by one, their combined
contribution is at most \(2/n\).
Therefore, on the event in Lemma~\ref{lemma_M},
\begin{align*}
\frac1n\sum_{j=1}^n R_1(i,j)
&\leq
\frac1{|N_i|}
+
\frac1{|N_i|^2}
\sum_{\substack{i',i''\in N_i\\i'\neq i''}}
\left\{
\frac{n-2}{n}M(i',i'')+\frac2n
\right\}\\
&\leq
\frac1{|N_i|}
+
\frac{|N_i|-1}{|N_i|}
\left\{
\frac{n-2}{n}\epsilon_m+\frac2n
\right\}.
\end{align*}
The factor \(|N_i|(|N_i|-1)\) appears because the cross-term expansion
contains ordered pairs \((i',i'')\).

For the profile-bias component, the Cauchy-Schwarz inequality gives
\begin{align*}
\frac1n\sum_{j=1}^n R_2(i,j)
&=
\frac1n
\sum_{j=1}^n
\left\{
\frac1{|N_i|}
\sum_{i'\in N_i}
(P_{i'j}-P_{ij})
\right\}^2\\
&\leq
\frac1{|N_i|}
\sum_{i'\in N_i}
\left\{
\frac1n
\sum_{j=1}^n
(P_{i'j}-P_{ij})^2
\right\}\\
&=
\frac1{|N_i|}
\sum_{i'\in N_i}
\frac1n
\|\mathbf P_{i'\cdot}-\mathbf P_{i\cdot}\|_2^2.
\end{align*}
On the event in Lemma~\ref{lemma:Pi-Pi'}, every \(i'\in N_i\)
satisfies
\[
\frac1n
\|\mathbf P_{i'\cdot}-\mathbf P_{i\cdot}\|_2^2
\leq
8\epsilon_{AP}.
\]
Therefore,
\begin{equation}
\label{j2bound}
\frac1n\sum_{j=1}^n R_2(i,j)
\leq
8\epsilon_{AP}.
\end{equation}
This conclusion holds on the intersection of the events controlled by
Lemmas~\ref{lemma:nodeproportion} and~\ref{lem:AAT-PPT}.
Combining \eqref{bound 2Js} with the bounds for \(R_1\) and
\(R_2\), we obtain, simultaneously for every \(i\in[n]\),
\begin{align}
\frac1n
\|\widetilde{\mathbf P}_{i\cdot}-\mathbf P_{i\cdot}\|_2^2
&\leq
\frac2{|N_i|}
+
2\frac{|N_i|-1}{|N_i|}
\left\{
\frac{n-2}{n}\epsilon_m+\frac2n
\right\}
+
16\epsilon_{AP}
\notag\\
&\label{simplified bound}
\leq
\frac2{|N_i|}
+
2\epsilon_m
+
\frac4n
+
16\epsilon_{AP}.
\end{align}
The success event is the intersection of the three events controlled by
Lemmas~\ref{lemma:nodeproportion}, \ref{lem:AAT-PPT}, and
\ref{lemma_M}. A union bound on their complements yields probability at
least
\begin{equation}
\label{simplified}
\begin{aligned}
1
&-
2K_n
\exp\left\{
-\frac{n\epsilon_\pi^2}{2}
\right\}\\
&-
2n^2
\exp\left\{
-\frac{n(\epsilon_{AP}-2/n)^2}
{4(1+\epsilon_{AP})}
\right\}\\
&-
n(n-1)
\exp\left\{
-\frac{n\epsilon_m^2}
{4(1+\epsilon_m)}
\right\}.
\end{aligned}
\end{equation}
This is the probability bound in Theorem~\ref{thm:rowwise_bound}.
Assumption~\ref{aasump:condition_epsilon_pi} and
conditions~\eqref{assump:condition_epsilon_AP} and
\eqref{assump:condition_epsilon_m} imply that each failure probability
converges to zero.
\end{proof}

\subsection{Proof of Corollary~\ref{cor:two_to_infinity}}

\begin{proof}
Set
\[
\epsilon_{AP}
=
C_{AP}\sqrt{\frac{\log n}{n}},
\qquad
\epsilon_m
=
C_m\sqrt{\frac{\log n}{n}},
\]
where \(C_{AP}>\sqrt8\) and \(C_m>\sqrt8\) are fixed. Indeed,
\[
\frac{
n(\epsilon_{AP}-2/n)^2
}{
4(1+\epsilon_{AP})
}
-
2\log n
=
\left(\frac{C_{AP}^2}{4}-2+o(1)\right)\log n
\longrightarrow\infty,
\]
and
\[
\frac{n\epsilon_m^2}{4(1+\epsilon_m)}
-
\log\!\bigl(n(n-1)\bigr)
=
\left(\frac{C_m^2}{4}-2+o(1)\right)\log n
\longrightarrow\infty.
\] Assumption
\ref{aasump:condition_epsilon_pi} controls the first failure probability
in \eqref{simplified}, and the two displayed conditions control the
remaining two. Hence the total failure probability in
\eqref{simplified} is \(o(1)\).

Now let
\[
h
=
C_h\sqrt{\frac{\log n}{n}},
\qquad
0<C_h<C_\rho-C_\pi.
\]
Assumption~\ref{aasump:condition_epsilon_pi} gives
\(h<\rho_{\min}-\epsilon_\pi\) for all sufficiently large \(n\).
Lemma~\ref{lem:neighborhood-size} yields
\[
|N_i|
\geq
(n-1)C_h\sqrt{\frac{\log n}{n}},
\]
and therefore
\[
\frac1{|N_i|}
\leq
\frac{1}{
C_h(n-1)\sqrt{\log n/n}
}
=
\frac{\sqrt n}{
C_h(n-1)\sqrt{\log n}
}.
\]
For all sufficiently large \(n\),
\[
\frac{\sqrt n}{
(n-1)\sqrt{\log n}
}
\leq
\sqrt{\frac{\log n}{n}},
\]
and hence
\[
\frac1{|N_i|}
\leq
C_h^{-1}\sqrt{\frac{\log n}{n}}.
\]
Using \eqref{simplified bound},
\[
\frac1n
\|\widetilde{\mathbf P}_{i\cdot}-\mathbf P_{i\cdot}\|_2^2
\leq
C_1\sqrt{\frac{\log n}{n}}
\]
simultaneously for all \(i\), where one may take
\[
C_1
=
2C_h^{-1}
+
2C_m
+
4
+
16C_{AP}.
\]
Multiplying by \(n\), taking square roots, and maximizing over \(i\)
gives
\[
\|\widetilde{\mathbf P}-\mathbf P\|_{2\to\infty}
\leq
C_1^{1/2}(n\log n)^{1/4}
\]
on an event whose probability tends to one. This proves
Corollary~\ref{cor:two_to_infinity}.
\end{proof}

\subsection{Proof of Corollary~\ref{cor:frobenius_bound}}

\begin{proof}
For every matrix \(\mathbf M\in\mathbb R^{n\times n}\),
\[
\|\mathbf M\|_F^2
=
\sum_{i=1}^n
\|\mathbf M_{i\cdot}\|_2^2
\leq
n\|\mathbf M\|_{2\to\infty}^2.
\]
Therefore,
\[
\|\mathbf M\|_F
\leq
\sqrt n\,\|\mathbf M\|_{2\to\infty}.
\]
Applying this inequality to
\(\mathbf M=\widetilde{\mathbf P}-\mathbf P\) gives the event
containment
\[
\left\{
\|\widetilde{\mathbf P}-\mathbf P\|_F
\geq
C_1^{1/2}n^{3/4}(\log n)^{1/4}
\right\}
\subseteq
\left\{
\|\widetilde{\mathbf P}-\mathbf P\|_{2\to\infty}
\geq
C_1^{1/2}(n\log n)^{1/4}
\right\}.
\]
The probability of the event on the right converges to zero by
Corollary~\ref{cor:two_to_infinity}, proving the result.
\end{proof}

\subsection{Proof of Theorem~\ref{thm:exact_recovery}
(Exact Recovery via Separation)}
\label{sec:supp-proofs-exact}

We first record the minimum-community-size event needed for the
separation argument.

\begin{lemma}[Minimum realized community size]
\label{lemma:n_min}
Under Assumption~\ref{aasump:condition_epsilon_pi},
\[
\mathbb P\left\{
n_{\min}\geq\frac{n\rho_{\min}}2
\right\}
\geq
1-
K_n\exp\left(-\frac{n\rho_{\min}}8\right)
=
1-o(1).
\]
\end{lemma}

\begin{proof}
For each \(a\in[K_n]\),
\[
n_a
\sim
\operatorname{Binomial}(n,\rho_a).
\]
The multiplicative Chernoff inequality gives
\[
\mathbb P\left\{
n_a\leq\frac{n\rho_a}{2}
\right\}
\leq
\exp\left(-\frac{n\rho_a}{8}\right)
\leq
\exp\left(-\frac{n\rho_{\min}}8\right).
\]
Since
\[
\left\{
n_a<\frac{n\rho_{\min}}2
\right\}
\subseteq
\left\{
n_a<\frac{n\rho_a}{2}
\right\},
\]
a union bound over the \(K_n\) communities proves the displayed
probability bound.

It remains to verify that the failure probability tends to zero.
Because the proportions sum to one,
\(K_n\rho_{\min}\leq1\), and hence \(K_n\leq1/\rho_{\min}\).
Assumption~\ref{aasump:condition_epsilon_pi} gives
\(n\rho_{\min}\geq C_\rho\sqrt{n\log n}\), so
\[
K_n\exp\left(-\frac{n\rho_{\min}}8\right)
\leq
\frac1{\rho_{\min}}
\exp\left(-\frac{n\rho_{\min}}8\right)
\longrightarrow0.
\]
\end{proof}

\begin{lemma}[Minimum between-community row separation]
\label{lemma:diff in P rows}
On the event
\[
n_{\min}\geq\frac{n\rho_{\min}}2,
\]
the block-profile target satisfies
\[
d_{\mathbf P}^*
\geq
\gamma_n
\sqrt{\frac{n\rho_{\min}}2}\,
d_{\mathbf B}^*.
\]
Consequently, this bound holds with probability tending to one under
Assumption~\ref{aasump:condition_epsilon_pi}.
\end{lemma}

\begin{proof}
Equation~\eqref{P_p_nul_B relation} in the main text gives, for every
pair \(i,j\) with \(\pi(i)\neq\pi(j)\),
\[
d_{\mathbf P}(i,j)
\geq
\gamma_n\sqrt{n_{\min}}\,d_{\mathbf B}^*.
\]
On the stated event, the right-hand side is at least
\[
\gamma_n
\sqrt{\frac{n\rho_{\min}}2}\,
d_{\mathbf B}^*.
\]
Taking the minimum over all between-community pairs proves the result.
The same minimum-size event controls all pairs simultaneously, so no
additional union bound over community pairs is needed.
\end{proof}

\begin{proof}[Proof of Theorem~\ref{thm:exact_recovery}]
Let
\[
\beta_n
=
\|\widetilde{\mathbf P}-\mathbf P\|_{2\to\infty}.
\]
By Corollary~\ref{cor:two_to_infinity},
\[
\beta_n
\leq
C_1^{1/2}(n\log n)^{1/4}
\]
with probability tending to one. By Lemmas~\ref{lemma:n_min}
and~\ref{lemma:diff in P rows}, with probability tending to one,
\[
n_{\min}\geq\frac{n\rho_{\min}}2
\]
and
\[
d_{\mathbf P}^*
\geq
\gamma_n
\sqrt{\frac{n\rho_{\min}}2}\,
d_{\mathbf B}^*.
\]
On this intersection of events,
\[
\begin{aligned}
\frac{d_{\mathbf P}^*}{8}
\sqrt{\frac{n_{\min}}n}
&\geq
\frac{\gamma_n d_{\mathbf B}^*}{8}
\sqrt{\frac{n\rho_{\min}}2}
\sqrt{\frac{\rho_{\min}}2}\\
&=
\frac{\gamma_n d_{\mathbf B}^*\rho_{\min}\sqrt n}{16}.
\end{aligned}
\]
Assumption~\ref{assump:gammaBrho} and
\(C_{\mathrm{sep}}>16C_1^{1/2}\) imply that, for all sufficiently large
\(n\),
\[
\frac{\gamma_n d_{\mathbf B}^*\rho_{\min}\sqrt n}{16}
>
C_1^{1/2}(n\log n)^{1/4}
\geq
\beta_n.
\]
Therefore,
\[
\beta_n
<
\frac{d_{\mathbf P}^*}{8}
\sqrt{\frac{n_{\min}}n}
\]
with probability tending to one. Lemma~\ref{lem:deterministic_clustering}
then gives
\[
A(\pi,\widehat\pi)=1.
\]
This proves the theorem.
\end{proof}

\subsection{Proof of Corollary~\ref{cor:estimated-K}}

\begin{proof}
Let
\[
\mathcal E_n
:=
\{\widehat K_n=K_n\}.
\]
On \(\mathcal E_n\), the plug-in procedure uses the true number of
communities and therefore coincides with the oracle-\(K_n\) procedure
in Theorem~\ref{thm:exact_recovery}. Consequently,
\begin{align*}
&\mathbb P\left\{
\substack{
\widehat K_n=K_n
\ \text{and there exists }\sigma\in\mathcal S_{K_n}\\
\pi(i)
=
\sigma\bigl(\widehat\pi_{\widehat K_n}(i)\bigr)
\text{ for every }i\in[n]
}
\right\}\\
&\qquad\geq
1-\mathbb P(\widehat K_n\neq K_n)\\
&\qquad\quad
-\mathbb P\left\{
A\bigl(\pi,\widehat\pi_{K_n}\bigr)<1
\right\}.
\end{align*}
The first probability on the right converges to zero by consistency of
\(\widehat K_n\), and the second converges to zero by
Theorem~\ref{thm:exact_recovery}. The claimed result follows.
\end{proof}

\section{Additional methodological and computational details}
\label{sec:supp-method-computation}

\subsection{Relation between spectral comparison methods and block profiles}
\label{sec:supp-spectral-profile-relation}

D-SCORE and related directed spectral procedures construct their
coordinates from the leading left and right singular vectors of
\(\mathbf A\), equivalently from the leading eigenspaces of
\(\mathbf A\mathbf A^\top\) and
\(\mathbf A^\top\mathbf A\). Recall that
\(\mathbf P=\gamma_n\mathbf P^0\) is the complete block-profile target,
including its block-defined diagonal, whereas
\[
\mathbf P^{\mathrm{off}}
=
\mathbb E(\mathbf A\mid\mathbf Z)
=
\mathbf P-\operatorname{diag}(\mathbf P)
\]
is the zero-diagonal conditional mean adjacency matrix.

Conditional independence of the directed edges gives
\[
\mathbb E\!\left(
\mathbf A\mathbf A^\top
\mid\mathbf Z
\right)
=
\mathbf P^{\mathrm{off}}
(\mathbf P^{\mathrm{off}})^\top
+
\mathbf D_{\mathrm{out}},
\]
where
\[
\mathbf D_{\mathrm{out}}
=
\operatorname{diag}\!\left\{
\sum_{k=1}^n
P^{\mathrm{off}}_{ik}
\bigl(1-P^{\mathrm{off}}_{ik}\bigr)
\right\}_{i=1}^n.
\]
Similarly,
\[
\mathbb E\!\left(
\mathbf A^\top\mathbf A
\mid\mathbf Z
\right)
=
(\mathbf P^{\mathrm{off}})^\top
\mathbf P^{\mathrm{off}}
+
\mathbf D_{\mathrm{in}},
\]
where
\[
\mathbf D_{\mathrm{in}}
=
\operatorname{diag}\!\left\{
\sum_{k=1}^n
P^{\mathrm{off}}_{ki}
\bigl(1-P^{\mathrm{off}}_{ki}\bigr)
\right\}_{i=1}^n.
\]
In particular, for \(i\neq j\),
\[
\mathbb E\!\left[
(\mathbf A\mathbf A^\top)_{ij}
\mid\mathbf Z
\right]
=
\left\langle
\mathbf P^{\mathrm{off}}_{i\cdot},
\mathbf P^{\mathrm{off}}_{j\cdot}
\right\rangle.
\]

Thus, \(\mathbf A\mathbf A^\top\) contains empirical similarities
between outgoing adjacency rows, whereas
\(\mathbf A^\top\mathbf A\) contains empirical similarities between
incoming adjacency columns. At the population level,
\[
\mathbf P^{\mathrm{off}}
(\mathbf P^{\mathrm{off}})^\top
\]
encodes pairwise Euclidean distances between the zero-diagonal outgoing
edge-block profiles, while
\[
(\mathbf P^{\mathrm{off}})^\top
\mathbf P^{\mathrm{off}}
\]
encodes the corresponding incoming-profile geometry.

These Gram matrices do not directly estimate the individual rows of the
full block-profile target \(\mathbf P\). Moreover, spectral truncation
replaces the complete profiles by low-dimensional coordinates. KMP
instead retains estimated outgoing block profiles as the features
supplied to \(K\)-means. Incoming profiles enter this discussion only
to describe the additional information used by the left-right spectral
benchmarks.

\subsection{Computational details for exact and landmark KMP}
\label{sec:supp-computation-details}

The principal computational cost is the construction of the pairwise
dissimilarities \(d(i,j)\). Define
\[
\mathbf G=\frac{1}{n}\mathbf A\mathbf A^\top.
\]
Then the normalized dissimilarity in \eqref{eq:distance} can be written as
\[
d(i,j)
=
\max_{k\neq i,j}
|G_{ik}-G_{jk}|.
\]
A direct dense implementation first forms \(\mathbf G\) and then
evaluates the maximum over \(k\) for every pair \((i,j)\). Both stages
have worst-case computational cost \(O(n^3)\). The dissimilarities may
be processed in blocks, so the working memory need not exceed
\(O(n^2)\).

Let
\[
s_i:=|N_i|,
\qquad
S_N:=\sum_{i=1}^n s_i.
\]
Direct dense averaging of the selected adjacency rows costs
\(
O(nS_N),
\)
because the \(s_i\) selected rows for node \(i\) each have \(n\)
coordinates. If the realized neighborhood sizes satisfy
\(s_i=O(nh)\) uniformly and
\(
h=C_h\sqrt{\frac{\log n}{n}},
\)
then
\[
S_N=O(n^2h)
\]
and the averaging step costs
\(
O(n^3h)
=
O\!\left(n^{5/2}\sqrt{\log n}\right).
\)
This is a typical-size calculation rather than a worst-case bound.
Because all ties at the empirical quantile are retained, the realized
neighborhoods can be substantially larger than \(nh\); in the worst
case \(S_N=O(n^2)\), and dense row averaging costs \(O(n^3)\).
Sparse multiplication can be considerably cheaper when
\(\mathbf A\) and the neighborhood-incidence matrix are sparse.

If Lloyd's algorithm uses \(T\) iterations, clustering \(n\) profiles
of dimension \(n\) into \(K_n\) groups costs
\(O(TK_n n^2)\).

For comparison, a truncated rank-\(K_n\) singular-value decomposition
has approximate dense cost \(O(n^2K_n)\) when the number of numerical
iterations is fixed, and sparse implementations may scale with
\(\operatorname{nnz}(\mathbf A)K_n\). Thus, for fixed or slowly growing
\(K_n\), exact KMP is computationally more expensive than the spectral
alternatives.

The \(O(n^3)\) cost of the all-witness dissimilarity stage arises from
evaluating the maximum over all \(n\) witness nodes for all
\(O(n^2)\) vertex pairs. A landmark approximation selects a subset
\(\mathcal A\subset[n]\), with \(|\mathcal A|=m\ll n\), and computes
\[
d_m(i,j)
=
\max_{k\in\mathcal A\setminus\{i,j\}}
\left|
\frac{1}{n}
\left\langle
\mathbf A_{i\cdot}-\mathbf A_{j\cdot},
\mathbf A_{k\cdot}
\right\rangle
\right|.
\]

\begin{algorithm}[ht]
\caption{Landmark neighborhood smoothing and clustering}
\label{alg:landmark-kmp}
\KwIn{Adjacency matrix \(\mathbf A\in\{0,1\}^{n\times n}\);
number of clusters \(K\); bandwidth \(h\in(0,1)\);
landmark count \(3\leq m\leq n\).}
\KwOut{Landmark-smoothed profiles
\(\widetilde{\mathbf P}^{(m)}\) and labels \(\widehat\pi^{(m)}\).}

Sample a landmark set
\(\mathcal A=\{a_1,\ldots,a_m\}\subset[n]\), with
\(|\mathcal A|=m\), uniformly without replacement\;

Compute
\[
H_{i\ell}
=
\frac1n
\left\langle
\mathbf A_{i\cdot},
\mathbf A_{a_\ell\cdot}
\right\rangle,
\qquad
i\in[n],\ \ell\in[m];
\]

\For{\(i\gets1\) \KwTo \(n\)}{
  \ForEach{\(j\in[n]\setminus\{i\}\)}{
    Set
    \[
    d_m(i,j)
    =
    \max_{\substack{\ell\in[m]\\a_\ell\notin\{i,j\}}}
    |H_{i\ell}-H_{j\ell}|;
    \]
  }

  Let \(q_i^{(m)}(h)\) be the empirical \(h\)-quantile of
  \(\{d_m(i,j):j\neq i\}\)\;

  Set
  \[
  N_i^{(m)}
  =
  \left\{
  j\neq i:
  d_m(i,j)\leq q_i^{(m)}(h)
  \right\};
  \]
  all ties at the quantile are retained\;

  Set
  \[
  \widetilde{\mathbf P}^{(m)}_{i\cdot}
  =
  \frac1{|N_i^{(m)}|}
  \sum_{u\in N_i^{(m)}}\mathbf A_{u\cdot};
  \]
}

Apply \(K\)-means with \(K\) clusters to the rows of
\(\widetilde{\mathbf P}^{(m)}\), producing
\(\widehat\pi^{(m)}\)\;
\end{algorithm}

The implementation uses the same endpoint-exclusion convention as
\eqref{eq:distance}: when \(i\) or \(j\) belongs to the landmark set,
its corresponding landmark column is omitted from the maximum. The only
approximation is therefore the restriction from all admissible witnesses
to the sampled landmark set.

Forming
\[
\mathbf H
=
n^{-1}\mathbf A\mathbf A_{\mathcal A}^{\top}
\]
and evaluating all landmark dissimilarities require \(O(n^2m)\) dense
operations. Storing \(\mathbf H\) costs \(O(nm)\) memory. Retaining the
full pairwise dissimilarity matrix adds \(O(n^2)\) memory, whereas
row-wise or blocked processing avoids that additional storage. These
costs describe the dissimilarity stage; neighborhood averaging and
\(K\)-means incur the additional costs described above. Sparse matrix
multiplication or approximate search may provide further reductions.

The main exact-recovery theorem concerns the all-witness dissimilarity
in \eqref{eq:distance}, whereas the simulation study uses \(d_{m_n}\)
with
\[
m_n=\min\{128,n\}
\]
to make the full numerical design computationally feasible. The following remark identifies a sufficient uniform
approximation condition under which the row-wise and exact-recovery
arguments extend to landmark KMP. Deriving an explicit sampling bound
that verifies this condition as a function of \(m\), \(n\), and the
block geometry remains open.

\begin{remark}[Conditional stability of the landmark approximation]
\label{rem:landmark-stability}

Because
\(\mathcal A\setminus\{i,j\}\subset[n]\setminus\{i,j\}\),
\(
d_m(i,j)\leq d(i,j).
\)
Define the uniform landmark deficit by
\[
\Delta_m
:=
\max_{i\neq j}
\{d(i,j)-d_m(i,j)\}.
\]
Suppose that there exists a deterministic sequence
\(\eta_{m,n}\geq0\) such that
\[
\mathbb P\{\Delta_m\leq\eta_{m,n}\}
\longrightarrow1.
\]
On the concentration event in
Lemma~\ref{lem:dissimilarity-neighborhood}, every same-community
candidate \(u\) satisfies
\(
d_m(i,u)
\leq
d(i,u)
\leq
2\epsilon_{AP}.
\)
Because more than an \(h\)-fraction of the candidate vertices belong to
the same community as \(i\), the landmark quantile obeys
\[
q_i^{(m)}(h)\leq2\epsilon_{AP}.
\]
Consequently, every landmark-selected neighbor
\(i'\in N_i^{(m)}\) satisfies
\[
d(i,i')
\leq
d_m(i,i')+\eta_{m,n}
\leq
2\epsilon_{AP}+\eta_{m,n}.
\]
Repeating the argument of Lemma~\ref{lemma:Pi-Pi'} gives
\(
\frac1n
\|\mathbf P_{i\cdot}-\mathbf P_{i'\cdot}\|_2^2
\leq
8\epsilon_{AP}+2\eta_{m,n}.
\)
Hence, the row-wise bound for the landmark-smoothed estimator
\(\widetilde{\mathbf P}^{(m)}\) becomes
\[
\frac1n
\|\widetilde{\mathbf P}^{(m)}_{i\cdot}
-\mathbf P_{i\cdot}\|_2^2
\leq
\frac2{|N_i^{(m)}|}
+
2\epsilon_m
+
\frac4n
+
16\epsilon_{AP}
+
4\eta_{m,n}.
\]
Therefore, if
\[
\eta_{m,n}
=
O\!\left(
\sqrt{\frac{\log n}{n}}
\right),
\]
the landmark estimator retains the normalized row-wise rate in
Corollary~\ref{cor:two_to_infinity}, up to a change in the constant.
The exact-recovery argument then remains valid under the corresponding
strengthened separation constant. Deriving an explicit probabilistic
bound on \(\Delta_m\) as a function of \(m\), \(n\), and the block
geometry is beyond the scope of this paper.
\end{remark}

\section{Additional simulation results}
\label{sec:supp-simulation-results}

The main text reports selected figures and compact endpoint summaries.
This section gives the complete \(MC=100\) numerical trajectories.
All methods are evaluated on the same simulated network within each
replicate.

\subsection{Matched-separation block matrix}
\label{sec:supp-matched-block}

The matched-separation asymmetric control is obtained by a column
permutation of the cyclic-banded matrix:
\[
\mathbf B_{\mathrm{match}}=
\begin{pmatrix}
0.42&0.28&0.18&0.55&0.50\\
0.50&0.42&0.28&0.20&0.55\\
0.55&0.50&0.42&0.28&0.20\\
0.20&0.55&0.50&0.42&0.28\\
0.28&0.20&0.55&0.50&0.42
\end{pmatrix}.
\]
As discussed in the main text, this column permutation preserves the
row distances and singular values while increasing normalized
asymmetry.

\subsection{Full independent-size ARI table}
\label{sec:supp-full-ari}
Tables~\ref{tab:supp-primary-ari-weak} and
\ref{tab:supp-primary-ari-controls} list the complete independent-size
ARI trajectories for all nine block matrices. The first table collects
the three weak-spectrum designs together with the weak-irregular and
strong-asymmetric contrasts; the second collects the hub designs and
well-conditioned controls.
\begin{table}[ht]
\centering
\caption{Mean ARI trajectories for the three weak-spectrum designs and
the weak-irregular and strong-asymmetric contrasts, based on \(MC=100\)
paired replicates. Within each network size, boldface identifies the
largest unrounded mean.}
\label{tab:supp-primary-ari-weak}
\scriptsize
\setlength{\tabcolsep}{4pt}
\renewcommand{\arraystretch}{0.82}
\begin{tabular}{@{}lccccc@{}}
\toprule
Design & \(n\) & KMA & KMP & Spectral-LR & D-SCORE\\
\midrule
Weak-layered \(D\)
& \makecell{200\\400\\600\\800\\1000\\1200\\1500\\2000}
& \makecell{0.217\\0.329\\0.438\\0.556\\0.631\\0.696\\0.746\\0.734}
& \makecell{0.190\\0.356\\\textbf{0.560}\\\textbf{0.665}\\\textbf{0.742}\\\textbf{0.794}\\\textbf{0.842}\\\textbf{0.889}}
& \makecell{0.250\\0.467\\0.491\\0.498\\0.499\\0.503\\0.502\\0.508}
& \makecell{\textbf{0.334}\\\textbf{0.478}\\0.495\\0.499\\0.500\\0.502\\0.503\\0.508}\\
\addlinespace[2pt]

Weak-profile \(C\)
& \makecell{200\\400\\600\\800\\1000\\1200\\1500\\2000}
& \makecell{0.188\\\textbf{0.374}\\\textbf{0.499}\\0.577\\0.656\\0.716\\0.770\\0.819}
& \makecell{0.161\\0.315\\0.484\\\textbf{0.633}\\\textbf{0.781}\\\textbf{0.838}\\\textbf{0.884}\\\textbf{0.923}}
& \makecell{\textbf{0.225}\\0.267\\0.279\\0.286\\0.296\\0.299\\0.303\\0.309}
& \makecell{0.217\\0.267\\0.396\\0.450\\0.470\\0.475\\0.479\\0.484}\\
\addlinespace[2pt]

Weak-cross \(B\)
& \makecell{200\\400\\600\\800\\1000\\1200\\1500\\2000}
& \makecell{0.198\\0.446\\\textbf{0.534}\\0.587\\0.649\\0.695\\0.739\\0.762}
& \makecell{0.178\\0.340\\0.489\\\textbf{0.609}\\\textbf{0.715}\\\textbf{0.778}\\\textbf{0.837}\\\textbf{0.895}}
& \makecell{0.229\\0.429\\0.475\\0.485\\0.492\\0.494\\0.495\\0.502}
& \makecell{\textbf{0.340}\\\textbf{0.462}\\0.487\\0.496\\0.499\\0.500\\0.500\\0.504}\\
\addlinespace[2pt]

Weak-irregular \(A\)
& \makecell{200\\400\\600\\800\\1000\\1200\\1500\\2000}
& \makecell{0.169\\\textbf{0.476}\\\textbf{0.630}\\\textbf{0.704}\\\textbf{0.746}\\\textbf{0.784}\\0.775\\0.789}
& \makecell{0.182\\0.330\\0.465\\0.552\\0.680\\0.752\\\textbf{0.812}\\\textbf{0.866}}
& \makecell{0.201\\0.352\\0.455\\0.472\\0.490\\0.492\\0.497\\0.500}
& \makecell{\textbf{0.276}\\0.425\\0.484\\0.559\\0.621\\0.675\\0.727\\0.740}\\
\addlinespace[2pt]

Strong asymmetric
& \makecell{200\\400\\600\\800\\1000\\1200\\1500\\2000}
& \makecell{\textbf{0.312}\\\textbf{0.466}\\\textbf{0.587}\\\textbf{0.670}\\\textbf{0.707}\\0.739\\0.739\\0.767}
& \makecell{0.211\\0.359\\0.483\\0.558\\0.622\\0.733\\0.802\\0.861}
& \makecell{0.246\\0.332\\0.425\\0.539\\0.599\\0.627\\0.646\\0.669}
& \makecell{0.256\\0.316\\0.381\\0.541\\0.668\\\textbf{0.796}\\\textbf{0.939}\\\textbf{0.985}}\\
\bottomrule
\end{tabular}
\end{table}

\begin{table}[p]
\centering
\caption{Mean ARI trajectories for the hub and well-conditioned
independent-size designs, based on \(MC=100\) paired replicates.
Within each network size, boldface identifies the largest unrounded
mean.}
\label{tab:supp-primary-ari-controls}
\scriptsize
\setlength{\tabcolsep}{4pt}
\renewcommand{\arraystretch}{0.90}
\begin{tabular}{@{}lccccc@{}}
\toprule
Design & \(n\) & KMA & KMP & Spectral-LR & D-SCORE\\
\midrule
Five-community hub
& \makecell{200\\400\\600\\800\\1000\\1200\\1500\\2000}
& \makecell{0.156\\0.171\\0.477\\0.571\\0.617\\0.572\\0.577\\0.607}
& \makecell{\textbf{0.197}\\\textbf{0.303}\\0.320\\0.333\\0.355\\0.375\\0.419\\0.529}
& \makecell{0.064\\0.102\\0.356\\0.642\\0.736\\0.878\\0.993\\0.999}
& \makecell{0.033\\0.113\\\textbf{0.534}\\\textbf{0.878}\\\textbf{0.958}\\\textbf{0.994}\\\textbf{0.999}\\\textbf{1.000}}\\
\addlinespace[2pt]

Eight-community hub
& \makecell{200\\400\\600\\800\\1000\\1200\\1500\\2000}
& \makecell{0.049\\0.066\\0.170\\0.305\\0.396\\0.463\\0.528\\0.604}
& \makecell{\textbf{0.076}\\\textbf{0.090}\\0.126\\0.135\\0.116\\0.128\\0.146\\0.190}
& \makecell{0.025\\0.050\\0.179\\0.366\\0.466\\0.527\\0.608\\0.712}
& \makecell{0.019\\0.061\\\textbf{0.213}\\\textbf{0.474}\\\textbf{0.733}\\\textbf{0.841}\\\textbf{0.922}\\\textbf{0.977}}\\
\addlinespace[2pt]

Cyclic-banded
& \makecell{200\\400\\600\\800\\1000\\1200\\1500\\2000}
& \makecell{0.827\\0.999\\\textbf{1.000}\\\textbf{1.000}\\\textbf{1.000}\\\textbf{1.000}\\\textbf{1.000}\\\textbf{1.000}}
& \makecell{0.454\\0.969\\0.996\\1.000\\1.000\\\textbf{1.000}\\\textbf{1.000}\\\textbf{1.000}}
& \makecell{0.956\\\textbf{1.000}\\\textbf{1.000}\\\textbf{1.000}\\\textbf{1.000}\\\textbf{1.000}\\\textbf{1.000}\\\textbf{1.000}}
& \makecell{\textbf{0.956}\\\textbf{1.000}\\\textbf{1.000}\\\textbf{1.000}\\\textbf{1.000}\\\textbf{1.000}\\\textbf{1.000}\\\textbf{1.000}}\\
\addlinespace[2pt]

Matched-separation asymmetric
& \makecell{200\\400\\600\\800\\1000\\1200\\1500\\2000}
& \makecell{0.867\\0.999\\\textbf{1.000}\\\textbf{1.000}\\\textbf{1.000}\\\textbf{1.000}\\\textbf{1.000}\\\textbf{1.000}}
& \makecell{0.486\\0.967\\0.996\\0.999\\1.000\\1.000\\\textbf{1.000}\\\textbf{1.000}}
& \makecell{\textbf{0.970}\\1.000\\\textbf{1.000}\\\textbf{1.000}\\\textbf{1.000}\\\textbf{1.000}\\\textbf{1.000}\\\textbf{1.000}}
& \makecell{0.969\\\textbf{1.000}\\\textbf{1.000}\\\textbf{1.000}\\\textbf{1.000}\\\textbf{1.000}\\\textbf{1.000}\\\textbf{1.000}}\\
\bottomrule
\end{tabular}
\end{table}
\FloatBarrier

\subsection{Full empirical exact-recovery table}
\label{sec:supp-full-exact}

Table~\ref{tab:supp-exact-recovery-extended} reports the designs
having at least one nonzero empirical exact-recovery frequency. Because
each rate is based on \(100\) paired replicates, the numerical
resolution is \(0.01\).

\begin{table}[ht]
\centering
\caption{Empirical global exact-recovery rates based on \(MC=100\)
paired replicates. Within each row, the largest positive rate is shown
in bold.}
\label{tab:supp-exact-recovery-extended}
\scriptsize
\begin{tabular}{@{}llcccc@{}}
\toprule
Design & \(n\) & KMA & KMP & Spectral-LR & D-SCORE\\
\midrule
\multirow{8}{*}{Cyclic-banded}&200&0.00&0.00&\textbf{0.07}&0.06\\
&400&0.84&0.01&\textbf{0.96}&\textbf{0.96}\\
&600&\textbf{1.00}&0.40&\textbf{1.00}&\textbf{1.00}\\
&800&\textbf{1.00}&0.89&\textbf{1.00}&\textbf{1.00}\\
&1000&\textbf{1.00}&0.94&\textbf{1.00}&\textbf{1.00}\\
&1200&\textbf{1.00}&\textbf{1.00}&\textbf{1.00}&\textbf{1.00}\\
&1500&\textbf{1.00}&\textbf{1.00}&\textbf{1.00}&\textbf{1.00}\\
&2000&\textbf{1.00}&\textbf{1.00}&\textbf{1.00}&\textbf{1.00}\\
\midrule
\multirow{8}{*}{Matched-separation asymmetric}&200&0.00&0.00&\textbf{0.12}&\textbf{0.12}\\
&400&0.83&0.00&0.95&\textbf{0.96}\\
&600&\textbf{1.00}&0.35&\textbf{1.00}&\textbf{1.00}\\
&800&\textbf{1.00}&0.85&\textbf{1.00}&\textbf{1.00}\\
&1000&\textbf{1.00}&0.98&\textbf{1.00}&\textbf{1.00}\\
&1200&\textbf{1.00}&0.99&\textbf{1.00}&\textbf{1.00}\\
&1500&\textbf{1.00}&\textbf{1.00}&\textbf{1.00}&\textbf{1.00}\\
&2000&\textbf{1.00}&\textbf{1.00}&\textbf{1.00}&\textbf{1.00}\\
\midrule
\multirow{8}{*}{Five-community hub}&200&0.00&0.00&0.00&0.00\\
&400&0.00&0.00&0.00&0.00\\
&600&0.00&0.00&0.00&0.00\\
&800&0.00&0.00&0.00&0.00\\
&1000&\textbf{0.01}&0.00&0.00&0.00\\
&1200&0.01&0.00&0.00&\textbf{0.08}\\
&1500&0.02&0.00&0.01&\textbf{0.55}\\
&2000&0.01&0.00&0.61&\textbf{0.92}\\
\midrule
\multirow{8}{*}{Weak-irregular $A$}&200&0.00&0.00&0.00&0.00\\
&400&0.00&0.00&0.00&0.00\\
&600&0.00&0.00&0.00&0.00\\
&800&0.00&0.00&0.00&0.00\\
&1000&0.00&0.00&0.00&0.00\\
&1200&0.00&0.00&0.00&0.00\\
&1500&0.00&0.00&0.00&0.00\\
&2000&\textbf{0.08}&0.00&0.00&0.00\\
\midrule
\multirow{8}{*}{Weak-cross $B$}&200&0.00&0.00&0.00&0.00\\
&400&0.00&0.00&0.00&0.00\\
&600&0.00&0.00&0.00&0.00\\
&800&0.00&0.00&0.00&0.00\\
&1000&0.00&0.00&0.00&0.00\\
&1200&0.00&0.00&0.00&0.00\\
&1500&0.00&0.00&0.00&0.00\\
&2000&\textbf{0.01}&0.00&0.00&0.00\\
\bottomrule
\end{tabular}
\end{table}
\FloatBarrier

The remaining primary designs have zero exact-recovery frequency for
all four methods over the displayed range. In particular, KMP's
advantage in weak-layered \(D\), weak-profile \(C\), and weak-cross
\(B\) concerns partial recovery rather than an earlier empirical
exact-recovery transition.

\subsection{Full sparsity trajectories}
\label{sec:supp-sparse-trajectories}

The summaries below are computed directly from the replicate-level file
using the stable grouping keys
\((\text{design},\text{regime},n,K,\text{method})\). Floating-point
diagnostic quantities are averaged within each cell rather than used as
grouping variables.

\begin{table}[H]
\centering
\caption{Mean ARI in the Dense regime, based on \(MC=100\)
paired replicates.}
\label{tab:supp-sparse-ari-dense}
\small
\begin{tabular}{@{}rcccc@{}}
\toprule
\(n\) & KMA & KMP & Spectral-LR & D-SCORE\\
\midrule
400 & 0.319 & 0.359 & 0.466 & \textbf{0.476}\\
600 & 0.429 & \textbf{0.561} & 0.489 & 0.495\\
800 & 0.537 & \textbf{0.668} & 0.497 & 0.500\\
1000 & 0.634 & \textbf{0.741} & 0.499 & 0.501\\
1200 & 0.689 & \textbf{0.789} & 0.501 & 0.502\\
1500 & 0.706 & \textbf{0.841} & 0.503 & 0.502\\
2000 & 0.723 & \textbf{0.889} & 0.508 & 0.506\\
\bottomrule
\end{tabular}
\end{table}

\begin{table}[H]
\centering
\caption{Mean ARI in the Quarter-power regime, based on \(MC=100\)
paired replicates.}
\label{tab:supp-sparse-ari-quarter-power}
\small
\begin{tabular}{@{}rcccc@{}}
\toprule
\(n\) & KMA & KMP & Spectral-LR & D-SCORE\\
\midrule
400 & 0.072 & 0.059 & 0.177 & \textbf{0.253}\\
600 & 0.209 & 0.094 & 0.277 & \textbf{0.372}\\
800 & 0.255 & 0.135 & 0.395 & \textbf{0.425}\\
1000 & 0.276 & 0.161 & 0.443 & \textbf{0.450}\\
1200 & 0.290 & 0.169 & 0.462 & \textbf{0.467}\\
1500 & 0.307 & 0.185 & 0.475 & \textbf{0.481}\\
2000 & 0.335 & 0.260 & 0.486 & \textbf{0.490}\\
\bottomrule
\end{tabular}
\end{table}

\begin{table}[H]
\centering
\caption{Mean ARI in the Square-root degree regime, based on \(MC=100\)
paired replicates.}
\label{tab:supp-sparse-ari-sqrt-degree}
\small
\begin{tabular}{@{}rcccc@{}}
\toprule
\(n\) & KMA & KMP & Spectral-LR & D-SCORE\\
\midrule
400 & 0.009 & 0.019 & 0.065 & \textbf{0.070}\\
600 & 0.008 & 0.021 & 0.110 & \textbf{0.131}\\
800 & 0.008 & 0.023 & 0.133 & \textbf{0.179}\\
1000 & 0.009 & 0.023 & 0.160 & \textbf{0.230}\\
1200 & 0.008 & 0.023 & 0.195 & \textbf{0.278}\\
1500 & 0.008 & 0.024 & 0.234 & \textbf{0.320}\\
2000 & 0.011 & 0.025 & 0.310 & \textbf{0.364}\\
\bottomrule
\end{tabular}
\end{table}

\begin{table}[H]
\centering
\caption{Mean ARI in the Log-degree regime, based on \(MC=100\)
paired replicates.}
\label{tab:supp-sparse-ari-log-degree}
\small
\begin{tabular}{@{}rcccc@{}}
\toprule
\(n\) & KMA & KMP & Spectral-LR & D-SCORE\\
\midrule
400 & 0.004 & 0.019 & 0.024 & \textbf{0.025}\\
600 & 0.003 & 0.017 & 0.022 & \textbf{0.023}\\
800 & 0.002 & 0.013 & 0.023 & \textbf{0.026}\\
1000 & 0.001 & 0.009 & 0.027 & \textbf{0.030}\\
1200 & 0.001 & 0.006 & 0.025 & \textbf{0.029}\\
1500 & 0.001 & 0.003 & 0.028 & \textbf{0.029}\\
2000 & 0.000 & 0.001 & 0.025 & \textbf{0.028}\\
\bottomrule
\end{tabular}
\end{table}

\FloatBarrier

At \(n=2000\), the KMP neighborhood-purity and tie-inflation pairs are
\[
(0.920,1.23),\quad
(0.350,3.48),\quad
(0.228,9.24),\quad
(0.200,13.99)
\]
for the dense, quarter-power, square-root-degree, and logarithmic-degree
regimes, respectively. These diagnostics explain why the dense KMP
advantage disappears as the graph becomes sparse.

\subsection{Nested growing-network details}
\label{sec:supp-nested-details}

Table~\ref{tab:supp-kmp-cohort-trajectories} gives the complete KMP
cohort-error trajectories.

\begin{table}[t]
\centering
\caption{Mean number of KMP errors among the initial \(100\) vertices
in the nested growing-network experiments, based on \(MC=100\) paired
paths.}
\label{tab:supp-kmp-cohort-trajectories}
\small
\begin{tabular}{@{}rcccc@{}}
\toprule
\(n\)
& Weak-layered \(D\)
& Weak-profile \(C\)
& Weak-irregular \(A\)
& Cyclic-banded\\
\midrule
100 & 52.24 & 56.60 & 54.36 & 47.83\\
200 & 51.53 & 56.31 & 50.62 & 29.99\\
400 & 38.59 & 45.96 & 45.42 & 1.24\\
600 & 21.30 & 34.24 & 35.83 & 0.21\\
800 & 15.09 & 22.52 & 31.07 & 0.02\\
1200 & 8.93 & 6.68 & 10.54 & 0.00\\
1500 & 6.60 & 5.24 & 7.36 & 0.00\\
\bottomrule
\end{tabular}
\end{table}

\begin{table}[t]
\centering
\caption{Global and fixed-cohort exact-recovery rates in the nested
cyclic-banded experiment, based on \(MC=100\) paired growing-network
paths.}
\label{tab:supp-nested-cyclic-exact}
\small
\begin{tabular}{@{}rlcccc@{}}
\toprule
\(n\) & Criterion & KMA & KMP & Spectral-LR & D-SCORE\\
\midrule
100 & Global & 0.00 & 0.00 & 0.00 & 0.00\\
 & Cohort & 0.00 & 0.00 & 0.00 & 0.00\\
200 & Global & 0.00 & 0.00 & 0.05 & 0.07\\
 & Cohort & 0.00 & 0.00 & 0.15 & 0.17\\
400 & Global & 0.77 & 0.01 & 0.97 & 0.96\\
 & Cohort & 0.94 & 0.23 & 0.99 & 0.99\\
600 & Global & 0.98 & 0.41 & 1.00 & 1.00\\
 & Cohort & 0.99 & 0.81 & 1.00 & 1.00\\
800 & Global & 1.00 & 0.80 & 1.00 & 1.00\\
 & Cohort & 1.00 & 0.98 & 1.00 & 1.00\\
1200 & Global & 1.00 & 0.99 & 1.00 & 1.00\\
 & Cohort & 1.00 & 1.00 & 1.00 & 1.00\\
1500 & Global & 1.00 & 1.00 & 1.00 & 1.00\\
 & Cohort & 1.00 & 1.00 & 1.00 & 1.00\\
\bottomrule
\end{tabular}
\end{table}
\FloatBarrier

The nested and independent-size experiments target different
probability laws. The former use dependent principal subgraphs from one
growing path within each replicate, whereas the latter generate a new
network at every \(n\). Their exact-recovery rates therefore need not
coincide.

\subsection{Controlled growth in the number of communities}
\label{sec:supp-growing-k}

The growing-\(K_n\) construction has mean block probability \(0.11\)
for every \(K\), with cyclic increments chosen to keep minimum row
separation approximately constant. Table~\ref{tab:supp-growing-k}
reports the mean ARI trajectories and the KMP exact-recovery rate.

\begin{table}[ht]
\centering
\caption{Controlled fixed- and growing-\(K_n\) experiment based on
\(MC=100\) paired replicates. The four middle columns give mean ARI;
the final column gives the KMP exact-recovery rate.}
\label{tab:supp-growing-k}
\scriptsize
\setlength{\tabcolsep}{4pt}
\begin{tabular}{@{}llrrrrrr@{}}
\toprule
Setting & \(n\) & \(K\) & KMA & KMP & Spectral-LR & D-SCORE
& KMP exact\\
\midrule
\multirow{8}{*}{Fixed $K=5$}&400&5&0.944&0.879&0.982&0.982&0.00\\
&600&5&0.994&0.968&0.999&0.999&0.00\\
&800&5&1.000&0.991&1.000&1.000&0.04\\
&1000&5&1.000&0.996&1.000&1.000&0.25\\
&1200&5&1.000&0.999&1.000&1.000&0.51\\
&1500&5&1.000&1.000&1.000&1.000&0.84\\
&2000&5&1.000&1.000&1.000&1.000&0.95\\
&3000&5&1.000&1.000&1.000&1.000&1.00\\
\midrule
\multirow{8}{*}{$K_n=\lceil\log n\rceil$}&400&6&0.690&0.810&0.914&0.913&0.00\\
&600&7&0.798&0.870&0.934&0.937&0.00\\
&800&7&0.937&0.935&0.989&0.989&0.00\\
&1000&7&0.979&0.971&0.997&0.997&0.00\\
&1200&8&0.946&0.959&0.998&0.998&0.00\\
&1500&8&0.975&0.982&1.000&1.000&0.00\\
&2000&8&0.989&0.995&1.000&1.000&0.01\\
&3000&9&0.977&0.997&1.000&1.000&0.01\\
\bottomrule
\end{tabular}
\end{table}
\FloatBarrier

For fixed \(K=5\), all methods reach exact-recovery rate \(1.00\) at
\(n=3000\). Under \(K_n=\lceil\log n\rceil\), KMP has mean ARI
\(0.997\) and exact-recovery rate \(0.01\) at \(n=3000\). At the same
endpoint, KMA has exact-recovery rate \(0.82\), while Spectral-LR and
D-SCORE both have exact-recovery rate \(1.00\).

\section{Additional \textit{C. elegans} connectome results}
\label{sec:supp-celegans}

\subsection{Data construction and implementation}

Section~\ref{sec:real-data} of the main text presents the primary
connectome analysis. Here we record the data-construction and
implementation details used for the supplementary diagnostics. The
analysis uses the hermaphrodite connectome of \citet{cook2019whole},
distributed through the OpenWorm Connectome
Toolbox.\footnote{\url{https://github.com/openworm/ConnectomeToolbox}}
Chemical-synapse counts are aggregated over each ordered neuron pair.
Electrical gap junctions, self-loops, pharyngeal neurons, and neurons
without a sensory, interneuron, or motor annotation are excluded. The
primary graph places an edge whenever the aggregated chemical-synapse
count is at least one.

Unlike the large-\(n\) simulations, the connectome analysis uses the
exact dissimilarity in equation~\eqref{eq:distance}. For each pair
\((i,j)\), the maximization excludes \(i\) and \(j\), the neighborhood
excludes the focal vertex, and every tie at the empirical quantile is
retained. Spectral-LR uses the unscaled concatenated singular-vector
coordinates \([\widehat{\mathbf U}_3,\widehat{\mathbf V}_3]\), and all
methods use \(20\) \(K\)-means++ initializations. The edge-thinning
experiment holds the algorithmic seed fixed across repetitions, so its
variation is due to edge deletion rather than initialization or
randomized linear algebra.

Table~\ref{tab:supp-celegans-degree} summarizes the pronounced degree
heterogeneity. The \(64\) zero-outdegree motor neurons account for most
of the \(66\) zero rows in the primary adjacency matrix. This feature
helps distinguish motor neurons as a group, but it also means that the
observed outgoing rows alone cannot separate the two zero-outdegree
sensory neurons from zero-outdegree motor neurons.
\begin{table}[t]
\centering
\caption{Functional-class sizes and outgoing-degree diagnostics in the
primary \textit{C. elegans} graph.}
\label{tab:supp-celegans-degree}
\small
\begin{tabular}{@{}lrrrr@{}}
\toprule
Functional type & \(n_a\) & Mean out-degree & Median out-degree
& Zero out-degree\\
\midrule
Sensory & 83 & 13.24 & 12 & 2\\
Interneuron & 81 & 16.20 & 17 & 0\\
Motor & 116 & 3.75 & 0 & 64\\
\bottomrule
\end{tabular}
\end{table}

\subsection{Synapse-count and bandwidth sensitivity}

To assess whether the primary result is driven only by weak
connections, we define
\[
A_{ij}^{(t)}
=
\mathbbm{1}\{W_{ij}^{\mathrm{chem}}\geq t\},
\qquad
t\in\{1,2,3,5\}.
\]
Table~\ref{tab:supp-celegans-threshold} reports ARI with matched
accuracy in parentheses. KMP is strongest at thresholds \(1,2,\) and
\(3\); at threshold \(5\), KMP and KMA have nearly identical ARI.
Performance decreases as increasingly many weak and moderate chemical
connections are removed.

\begin{table}[t]
\centering
\caption{Synapse-count sensitivity. Each entry gives ARI with
label-matched accuracy in parentheses. Boldface identifies the largest
unrounded ARI in each row.}
\label{tab:supp-celegans-threshold}
\small
\begin{tabular}{@{}rcccc@{}}
\toprule
Threshold \(t\) & KMP & KMA & Spectral-LR & D-SCORE\\
\midrule
1 & \textbf{0.246} (0.596) & 0.176 (0.536) & 0.170 (0.582) & 0.052 (0.450)\\
2 & \textbf{0.221} (0.571) & 0.145 (0.529) & 0.166 (0.561) & 0.048 (0.425)\\
3 & \textbf{0.165} (0.546) & 0.144 (0.529) & 0.130 (0.532) & 0.061 (0.446)\\
5 & \textbf{0.092} (0.521) & 0.092 (0.518) & 0.000 (0.407) & 0.084 (0.486)\\
\bottomrule
\end{tabular}
\end{table}

Table~\ref{tab:supp-celegans-bandwidth} gives a post hoc bandwidth
sensitivity analysis. Each clustering fit is computed without using the
metadata labels; the labels are used only afterward to report ARI and
matched accuracy. Although \(C_h=0.25\) produces the largest ARI, the
prespecified primary value \(C_h=0.5\) produces the largest matched
accuracy and is retained to avoid selecting a bandwidth from the
metadata labels. Ties substantially inflate the realized
neighborhood sizes throughout the grid.

\begin{table}[t]
\centering
\caption{KMP bandwidth sensitivity in the primary connectome.}
\label{tab:supp-celegans-bandwidth}
\small
\begin{tabular}{@{}rrrrrr@{}}
\toprule
\(C_h\) & Nominal size & Mean realized size & Mean purity & ARI & Accuracy\\
\midrule
0.25 & 10 & 40.98 & 0.603 & \textbf{0.257} & 0.593\\
0.50 & 20 & 56.08 & 0.577 & 0.246 & \textbf{0.596}\\
0.75 & 30 & 69.80 & 0.566 & 0.239 & 0.589\\
1.00 & 40 & 81.98 & 0.555 & 0.234 & 0.586\\
1.50 & 60 & 98.19 & 0.543 & 0.215 & 0.564\\
\bottomrule
\end{tabular}
\end{table}

\subsection{Edge thinning and class-specific errors}

Conditional on the observed connectome, Table~\ref{tab:supp-celegans-thinning}
reports the mean ARI over \(50\) independent edge-retention masks at each
value of \(q\). KMP has the largest mean at every retention probability.
These mean values are plotted in panel (b) of
Figure~\ref{fig:celegans-main} in the main text. The
experiment measures sensitivity to randomly missing observed
connections and should not be interpreted as a finite-sample
verification of the exact-recovery theorem.

\begin{table}[t]
\centering
\caption{Mean ARI under independent edge thinning.}
\label{tab:supp-celegans-thinning}
\small
\begin{tabular}{@{}rcccc@{}}
\toprule
Retained probability \(q\) & KMP & KMA & Spectral-LR & D-SCORE\\
\midrule
0.3 & \textbf{0.157} & 0.052 & 0.082 & 0.061\\
0.5 & \textbf{0.141} & 0.116 & 0.119 & 0.072\\
0.7 & \textbf{0.191} & 0.148 & 0.151 & 0.063\\
0.9 & \textbf{0.228} & 0.169 & 0.164 & 0.061\\
1.0 & \textbf{0.246} & 0.176 & 0.170 & 0.052\\
\bottomrule
\end{tabular}
\end{table}

The paired ARI comparisons in
Table~\ref{tab:supp-celegans-paired} qualify the mean differences.
KMP wins most replicates at \(q=0.3,0.7,\) and \(0.9\), but its
advantage over KMA and Spectral-LR is much less uniform at \(q=0.5\).
The \(q=1\) endpoint is omitted from the paired table because all \(50\)
rows are identical copies of the same unthinned graph under the fixed
algorithmic seed.

\begin{table}[t]
\centering
\caption{Number of the \(50\) thinned graphs on which KMP has strictly
larger ARI than each competitor.}
\label{tab:supp-celegans-paired}
\small
\begin{tabular}{@{}rccc@{}}
\toprule
\(q\) & versus KMA & versus Spectral-LR & versus D-SCORE\\
\midrule
0.3 & 45 & 41 & 42\\
0.5 & 29 & 26 & 49\\
0.7 & 46 & 49 & 50\\
0.9 & 49 & 50 & 50\\
\bottomrule
\end{tabular}
\end{table}

\begin{figure}[ht]
\centering
\begin{subfigure}[t]{0.49\textwidth}
\centering
\vspace{0pt}
\includegraphics[width=\textwidth]{\detokenize{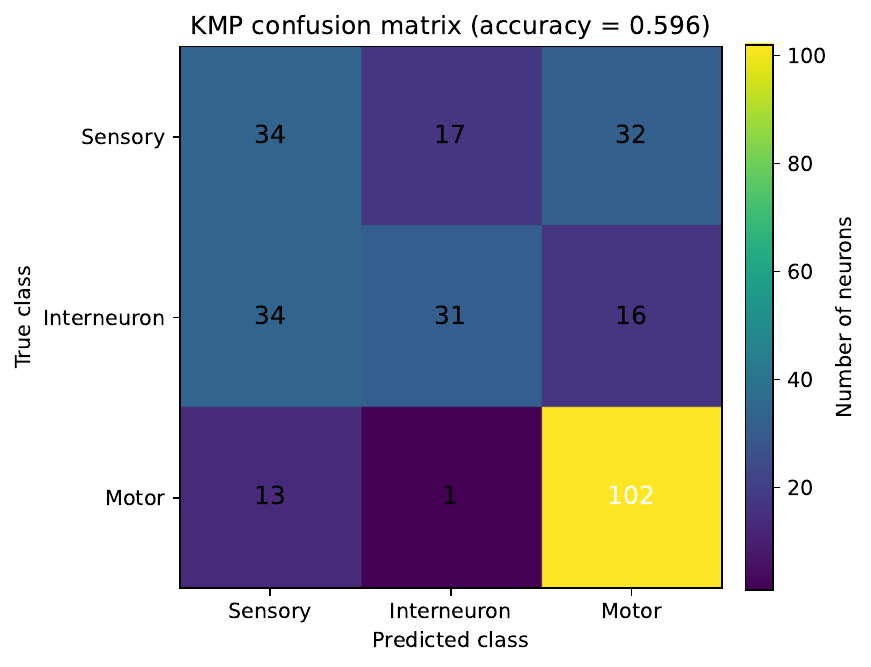}}
\caption{KMP.}
\end{subfigure}
\hfill
\begin{subfigure}[t]{0.49\textwidth}
\centering
\vspace{0pt}
\includegraphics[width=\textwidth]{\detokenize{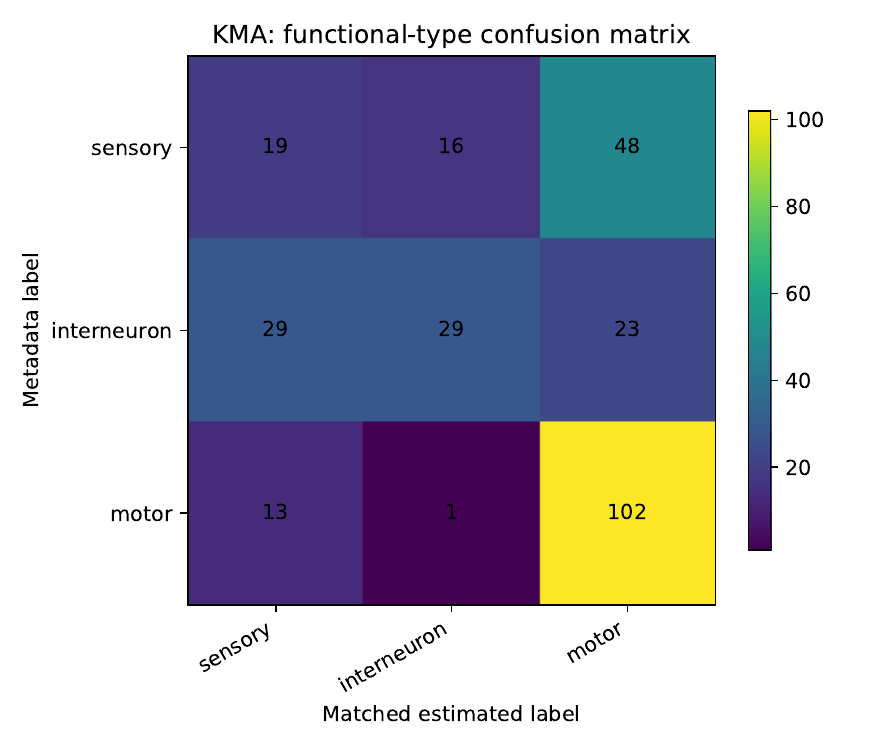}}
\caption{KMA.}
\end{subfigure}

\begin{subfigure}[t]{0.49\textwidth}
\centering
\vspace{0pt}
\includegraphics[width=\textwidth]{\detokenize{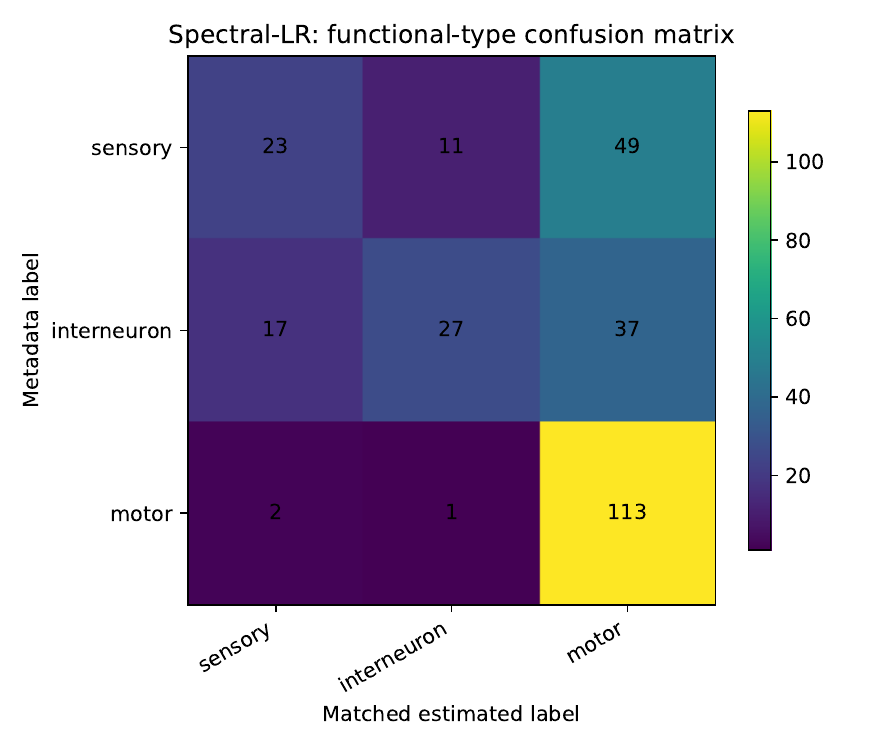}}
\caption{Spectral-LR.}
\end{subfigure}
\hfill
\begin{subfigure}[t]{0.49\textwidth}
\centering
\vspace{0pt}
\includegraphics[width=\textwidth]{\detokenize{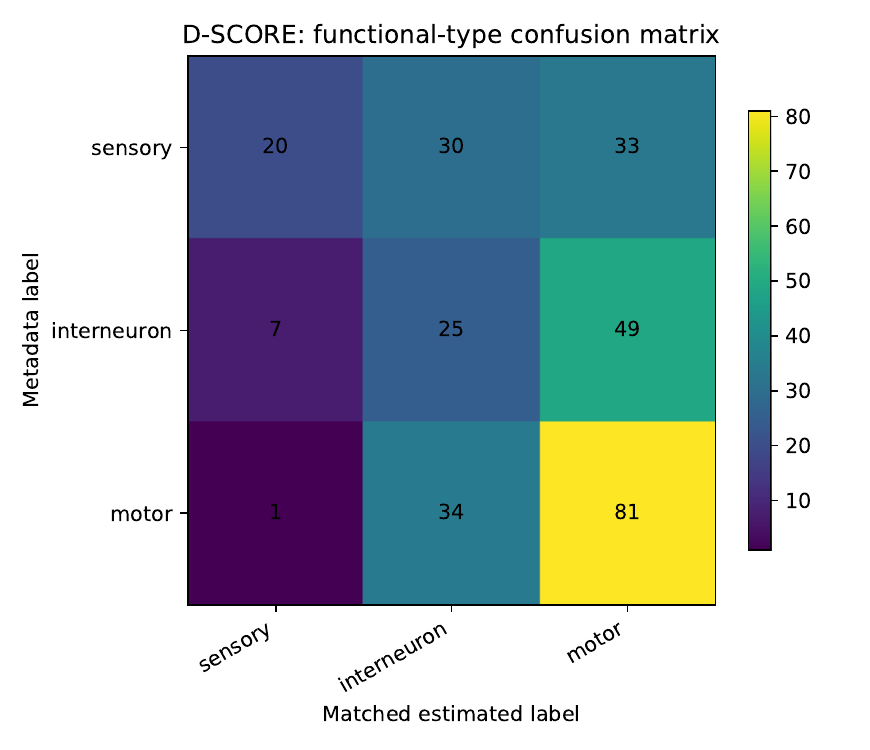}}
\caption{D-SCORE.}
\end{subfigure}
\caption{Confusion matrices after optimal label matching in the primary
\textit{C. elegans} graph. Rows are metadata classes and columns are
estimated classes.}
\label{fig:supp-celegans-confusion}
\end{figure}

Figure~\ref{fig:supp-celegans-confusion} gives the matched confusion
matrices. KMP correctly assigns \(102\) of the \(116\) motor neurons,
but sensory and interneuron errors remain substantial. This agrees with
the empirical outgoing block geometry: the sensory-interneuron row
distance is the smallest of the three pairwise separations.

\FloatBarrier

As an additional biological stability diagnostic, we identify \(92\)
left-right neuron pairs from their canonical names. The fractions
assigned to the same estimated cluster are \(0.826\) for KMP, \(0.859\)
for KMA, \(0.924\) for Spectral-LR, and \(0.891\) for D-SCORE. This
criterion therefore favors the spectral methods and provides a useful
counterpoint to the metadata ARI comparison. Together with the degree
heterogeneity and zero-row diagnostics, it supports a cautious
interpretation: KMP has the strongest agreement with the three broad
functional labels and the most favorable mean thinning behavior, but it
does not dominate every biologically motivated stability measure.
\setcounter{equation}{0}

\end{document}